\documentclass[nonacm,sigplan]{acmart}
\usepackage{multirow}
\usepackage{listings}
\usepackage{xcolor}

\usepackage{comment}

\usepackage{float}
\usepackage{caption}

\usepackage{array}
\usepackage{fancyhdr}
\usepackage{amsthm}

\usepackage{framed}
\usepackage[libertine]{newtxmath}
\usepackage[tt=false]{libertine}

\usepackage[utf8]{inputenc} 
\usepackage[T1]{fontenc}    
\usepackage{hyperref}       
\usepackage{url}            
\usepackage{booktabs}       
\usepackage{amsfonts}       
\usepackage{nicefrac}       
\usepackage{microtype}      
\microtypecontext{spacing=nonfrench}
\usepackage{xspace}

\makeatletter
\def\url@leostyle{%
  \@ifundefined{selectfont}{\def\UrlFont{\sf}}{\def\UrlFont{\small\ttfamily}}}
\makeatother
\usepackage{proof}
\usepackage{cleveref}
\usepackage{graphicx}
\usepackage{wrapfig}
\usepackage{subcaption}
\usepackage{mathpartir}

\usepackage[frozencache,cachedir=.]{minted}
\usepackage{minted}

\AtEndPreamble{%
}

\definecolor{codegreen}{rgb}{0,0.6,0}
\definecolor{codegray}{rgb}{0.5,0.5,0.5}
\definecolor{codepurple}{rgb}{0.58,0,0.82}
\definecolor{backcolour}{rgb}{0.95,0.95,0.92}

\lstdefinestyle{mystyle}{
    backgroundcolor=\color{white},   
    commentstyle=\color{blue},
    keywordstyle=[1]\color{blue},
    keywordstyle=[2]\color{magenta},
    keywordstyle=[3]\color{red},
    keywordstyle=[4]\color{codegray},
    numberstyle=\tiny\color{codegray},
    stringstyle=\color{codegreen},
    basicstyle=\ttfamily\footnotesize,
    breakatwhitespace=false,         
    breaklines=true,                 
    captionpos=b,                    
    keepspaces=true,                 
    numbers=left,                    
    numbersep=5pt,                  
    showspaces=false,                
    showstringspaces=false,
    showtabs=false,                  
    tabsize=2,
    texcl=true,
    escapeinside={(*@}{@*)}
}
\lstdefinelanguage{mlir}{
    alsoletter={\%,\#,!,.,_},
    morekeywords={\%},
    keywords=[1]{func, return, attributes},
    keywords=[2]{loop, \#tile, \#sum, slice, yield, atomic, spmd, range},
    keywords=[4]{tensor, f32},
    showstringspaces=false,
	breaklines=true,
    breakatwhitespace=true,
    morestring=[b]",
    morecomment=[l]{//},
}
\newcommand{\inlinecode}[1]{{\lstinline[basicstyle=\ttfamily]!#1!}}
\newcommand{\inlc}[1]{\inlinecode{#1}}

\newcommand{\coreloop}{{{\small\tt loop}}\xspace}
\newcommand{\ploop}[3]{\ensuremath{\coreloop_{#1}~#2~(\lambda r_{#1}. #3)}\xspace}   
\newcommand{\coretile}{{{\small\tt \#tile}}\xspace}
\newcommand{\shorttileaction}[1]{\ensuremath{\coretile\langle{\tt #1}\rangle}\xspace} 
\newcommand{\tileaction}[2]{\ensuremath{\coretile\langle{\tt #1, #2}\rangle}\xspace} 
\newcommand{\coresum}{{\small\tt \#sum}\xspace}
\newcommand{\sumaction}[1]{\ensuremath{\coresum{\langle{\tt #1}\rangle}}\xspace}     
\newcommand{\shortsumaction}{\ensuremath{\coresum}\xspace} 
\newcommand{\coretensor}{{\small\tt tensor}\xspace}
\newcommand{\tensor}[1]{\ensuremath{\coretensor\langle#1\rangle}\xspace}             
\newcommand{\coreslice}{{{\small\tt slice}}\xspace}
\newcommand{\slice}[3]{\ensuremath{\coreslice~#1~#2[#3]}\xspace}                     
\newcommand{\plet}{\xspace{\tt let}~}                                                
\newcommand{\pin}{~{\tt in}~}                                                        
\newcommand{\coreyield}{{\tt yield}\xspace}
\newcommand{\yield}[1]{\ensuremath{\coreyield(#1)}\xspace}                           
\newcommand{\corerange}{{\inlc{range}}\xspace}

\newcommand{\as}{\overline{a}}
\newcommand{\bs}{\overline{b}}
\newcommand{\cs}{\overline{c}}
\newcommand{\xs}{\overline{x}}
\newcommand{\ys}{\overline{y}}
\newcommand{\zs}{\overline{z}}

\newcommand{\rhos}{\overline{\rho}}
\newcommand{\sigmas}{\overline{\sigma}}

\newcommand{\ol}[1]{\overline{#1}}
\newcommand{\rs}{\overline{r}}

\newcommand{\ddim}[2]{\{#1\}#2}
\newcommand{\disttensor}[2]{\ensuremath{{\small\tt dtensor\langle}#1, #2{\tt\rangle}}\xspace}
\newcommand{\spmdexecute}{{\small\tt spmd.execute}\xspace}
\newcommand{\execute}[5]{\spmdexecute~#1~#2~(\lambda #3.\lambda #4.#5)}

\newcommand{\spmdredist}{{\small\tt spmd.redistribute}\xspace}
\newcommand{\redist}[2]{\ensuremath{\spmdredist~#1\blacktriangleright#2}\xspace}
\newcommand{\spmdtst}{{\small\tt spmd.tile\_reduce}\xspace}
\newcommand{\tst}[2]{\ensuremath{\spmdtst~#1~#2}\xspace}

\newcommand{\spmdallgather}{{\small\tt spmd.all\_gather}\xspace}
\newcommand{\allgather}[2]{\ensuremath{\spmdallgather~#1~#2}\xspace}
\newcommand{\spmdallsum}{{\small\tt spmd.all\_reduce}\xspace}
\newcommand{\allsum}[2]{\ensuremath{\spmdallsum~#1~#2}\xspace}
\newcommand{\spmdallslice}{{\small\tt spmd.all\_slice}\xspace}
\newcommand{\allslice}[2]{\ensuremath{\spmdallslice~#1~#2}\xspace}

\newcommand\localtype[1]{\mathcal{L}[#1]}
\newcommand\globaltype[1]{\mathcal{G}[#1]}
\newcommand\localdim[1]{\mathcal{L}[#1]}
\newcommand\globaldim[1]{\mathcal{G}[#1]}

\newcommand\tilered[2]{[\!\![#1]\!\!](#2)}

\newcommand{\core}{{\tt \scriptsize{core}}}
\newcommand{\spmd}{{\tt \scriptsize{spmd}}}

\newcommand{\tcore}{\vdash^{\hspace{-3pt}{\core}}}
\newcommand{\tspmd}{\vdash^{\hspace{-3pt}{\spmd}}}

\newcommand{\pctx}[1]{{\tt {#1}}\xspace}
\newcommand{\psctx}[1]{{\tt {#1}}^{\spmd}\xspace}

\newcommand{\FV}[1]{{\it free}(#1)\xspace}

\newcommand{\defs}[1]{{\it defs}(#1)\xspace}
\newcommand{\axes}[1]{{\it axes}(#1)\xspace}

\newcommand{\opmatmul}{{\tt matmul}\xspace}
\newcommand{\opadd}{{\tt add}\xspace}

\newif\ifcomments
\commentstrue

\ifcomments
\newcommand{\authorcomment}[3]{\textcolor{#2}{#1: {\emph{#3}}}}
\else
\newcommand{\authorcomment}[3]{}
\fi

\newcommand*{\defeq}{\stackrel{\text{def}}{=}}

\newcommand\mapping[3]{(#1\hookrightarrow #2)}

\newcommand{\sfixup}{\vdash^{\hspace{-3pt}{\scriptsize \tt act}}}

\newcommand\live{\ell}

\newcommand{\Infer}[3]{\inferrule*[right={#1}]{#2}{#3}}

\restylefloat{figure}

\newcommand{\reducedstrut}{\vrule width 0pt height .9\ht\strutbox depth .9\dp\strutbox\relax}
\newcommand{\codehl}[2]{%
  \begingroup
  \setlength{\fboxsep}{0pt}%
  \colorbox{#1}{\reducedstrut#2\/}%
  \endgroup
}

\newcommand\tmr{\textrm{TMR}}
\newcommand\emptystacked{\{\}}

\newif\ifextended
\extendedtrue

\crefname{lstlisting}{listing}{listings}
\Crefname{lstlisting}{Listing}{Listings}
\AtBeginDocument{%
  }
 
\newcommand{\partirjit}{\inlc{partir.jit}}
\newcommand{\partir}{PartIR\xspace}
\hypersetup{
  colorlinks=true,  
  linkcolor=black,  
}

\title{
\partir{}: Composing SPMD Partitioning Strategies for Machine Learning
}

\author{Sami Alabed}
\authornote{Equal contribution, authors in alphabetical order. Correspondence to: \url{dvytin@google.com}}
\affiliation{\institution{Google DeepMind} \city{London} \country{UK}}
\orcid{0000-0001-8716-526X}

\author{Daniel Belov}
\authornotemark[1]
\affiliation{\institution{Google DeepMind} \city{London} \country{UK}}

\author{Bart Chrzaszcz}
\authornotemark[1]
\affiliation{\institution{Google DeepMind} \city{London} \country{UK}}

\author{Juliana Franco}
\authornotemark[1]
\affiliation{\institution{Google DeepMind} \city{London} \country{UK}}
\orcid{0009-0005-1031-2598}

\author{Dominik Grewe}
\authornotemark[1]
\affiliation{\institution{Google DeepMind} \city{London} \country{UK}}

\author{Dougal Maclaurin}
\authornotemark[1]
\affiliation{\institution{Google DeepMind} \city{Cambridge} \country{US}}

\author{James Molloy}
\authornotemark[1]
\affiliation{\institution{Google DeepMind} \city{London} \country{UK}}

\author{Tom Natan}
\authornotemark[1]
\affiliation{\institution{Google DeepMind} \city{London} \country{UK}}

\author{Tamara Norman}
\authornotemark[1]
\affiliation{\institution{Google DeepMind} \city{London} \country{UK}}

\author{Xiaoyue Pan}
\authornotemark[1]
\affiliation{\institution{Google DeepMind} \city{London} \country{UK}}

\author{Adam Paszke}
\authornotemark[1]
\affiliation{\institution{Google DeepMind} \city{Warsaw} \country{Poland}}
\orcid{0000-0002-7665-4559}

\author{Norman A. Rink}
\authornotemark[1]
\affiliation{\institution{Google DeepMind} \city{London} \country{UK}}
\orcid{0009-0000-8591-5215}

\author{Michael Schaarschmidt}
\affiliation{\institution{Isomorphic Labs} \city{London} \country{UK}}
\authornote{Work done while at Google DeepMind}
\authornotemark[1]

\author{Timur Sitdikov}
\authornotemark[1]
\affiliation{\institution{Google DeepMind} \city{London} \country{UK}}

\author{Agnieszka Swietlik}
\authornotemark[1]
\affiliation{\institution{Google DeepMind} \city{London} \country{UK}}
\orcid{0009-0003-7276-9038}

\author{Dimitrios Vytiniotis}
\authornotemark[1]
\affiliation{\institution{Google DeepMind} \city{London} \country{UK}}
\orcid{0009-0007-2079-1996}

\author{Joel Wee}
\authornotemark[1]
\affiliation{\institution{Google DeepMind} \city{London} \country{UK}}
\orcid{0009-0005-3518-3526}

\begin{document}

\begin{abstract}

Training modern large neural networks (NNs) requires a combination of parallelization strategies, including data, model, or optimizer sharding. 
To address the growing complexity of these strategies, we introduce \partir{}, a hardware-and-runtime agnostic NN partitioning system. 
\partir{} is: 1) Expressive: It allows for the composition of multiple sharding strategies, whether user-defined or automatically derived; 2) Decoupled: the strategies are separate from the ML  implementation; and 3) Predictable: It follows a set of well-defined general rules to partition the NN. 
\partir{} utilizes a schedule-like API that incrementally rewrites the ML program intermediate representation (IR) after each strategy, allowing simulators and users to verify the strategy's performance.
\partir{} has been successfully used both for training large models and across diverse model architectures, demonstrating its predictability, expressiveness, and performance.
\end{abstract}

\maketitle

\section{Introduction}\label{sec:intro}
The recent growth of NN training requirements has significantly outpaced the increase in accelerator memory and FLOPS.
Google TPU~\cite{DBLP:journals/corr/JouppiYPPABBBBB17, jouppi+:lessons} v2 reports 46 TFLOPS / 16 GB of high-bandwidth memory (HBM) per chip, while v4 reports 275 TFLOPS / 32 GB\cite{tpu_cloud} over a period where model parameters and FLOPS requirements increased by $10^4$~\cite{flopsGrowth}. Due to the need to scale out, today's large NNs \cite{gpt3_2020, gopher2021, palm2022, chinchilla2022, galactica22, llama23} are trained on many accelerators through a mixture of parallelism strategies.
User-driven partitioners~\cite{gshard, gspmd2021, shazeer2018mesh} facilitate expressing a wide range of parallelism strategies in high-level ML frameworks~\cite{jax2018github, pytorch2019, tensorflow_osdi_2016}, without requiring ML engineers to write low-level distributed communication primitives; an error-prone and non-portable practice.
Despite their success, these tools regularly require users to provide sharding annotations inside their ML code. 
This practice raises significant maintainability concerns as it makes the code non-portable to different
topologies (e.g. when a pre-trained model on system X needs to be fine-tuned on system Y, to be deployed on system Z).
Furthermore, it makes it impossible to verify each parallelism strategy independently, forcing users to perform time-consuming profiling to debug the partitioner's result.
The complexity of partitioning for end-users has motivated research on automated partitioning tools~\cite{flexflow2019, wang+:automap, automap, automap-reloaded, unity2022, alpa2022}. 
These automatic partitioning tools provide flexibility, but can be unpredictable and slow~\cite{alpa2022, automap-reloaded}. Users want to benefit from the flexibility of automatic partitioning, while leveraging known and easy-to-apply strategies (such as batch parallelism) to speedup these tools and achieve performance guarantees.

\partir{} makes partitioning large models easier: it fully separates the model implementation from its partitioning, allowing ML engineers to focus on building models that {\em outlive the partitioning strategies} without concerns about how they need to be partitioned.
The details of partitioning are provided separately using sequences of {\em tactics}  that conceptually capture a ``parallelism strategy.''
These tactics invoke manual or automatic partitioning APIs and compose in any order; their composition forms a \partir \textit{schedule}. 
Each tactic is independent and can never undo sharding decisions introduced earlier in the same schedule.
The tactics are translated through \partir's compiler stack to device-local code that makes communication collectives explicit and verifiable. 

\partir{} is a set of passes in a compiler pipeline, invoked after NN tracing and before device-local optimization and code generation.
It can be invoked by any frontend (TensorFlow, PyTorch, JAX) to execute on any backend (XLA~\cite{xla}, OpenXLA~\cite{openxla}), achieved by designing \partir{} as an MLIR-based compiler~\cite{mlir2020}.
\partir{} is composed of several IRs, known as dialects, each of which is
tested independently, along with optimization and lowering transformations.
\partir:Core (\Cref{sec:core}) is our foundational dialect that introduces functional tiling and reduction loops on top of an array IR (e.g., StableHLO~\cite{stablehlo}). 
These loops abstract away execution semantics, enabling both sequential semantics for testing and parallel semantics for SPMD execution.
The \partir:HLO (\Cref{sec:mhlo}) dialect introduces and optimizes SPMD collective  communication operations. 
Having the collectives in the IR allows users to verify their strategies and analytically estimate the performance of the partitioned NNs after every tactic.
The critical transformation from \partir:Core to \partir:HLO is formally verified and shown in \Cref{app:sec:translation-correctness}.

\partir{} propagates sharding decisions across the module without requiring per-tensor annotations.
Our propagation avoids cost-based heuristics, relying instead on rewrite rules derived from the algebraic semantics of each operation. 
Ordering the tactics in the schedule resolves sharding conflicts (e.g., when sharding a tensor on multiple logical axes along the same dimension), resulting in a predictable and easier to control propagation.
Other tools, such as GSPMD~\cite{gspmd2021}, in contrast, require the user to add annotations in the right places by trial-and-error to resolve these conflicts.


\subsection{Contributions}
\partir has been actively used for scaling many ML models~\cite{bugliarello-etal-2023-measuring, bugliarello-etal-2023-weakly, da2023, doersch2023tapir}. The contributions of this work are:
\begin{itemize}
    \item A partitioner that decouples parallelism strategies from the NN implementation using a schedule API to compose manual or automated strategies (\Cref{ssec:schedule}).

    \item A compartmentalized MLIR compiler (\Cref{sec:architecture}), organized in independently tested dialects.
    The dialects abstract reasoning about array IRs, partitioning across device meshes, and optimizing SPMD collectives.

    \item A compiler pass that propagates sharding decisions throughout the module without relying on heuristics or cost models (\Cref{ssec:propagation}).

    \item The interplay between expressive tactics and powerful propagation enables partitioning of very large ML models. 
    While schedule-implementation separation has been influential in modern kernel-level optimization~\cite{Halide:PLDI:2013}, our work is the first to adapt and make the idea practical for the
    partitioning of full programs.

\end{itemize} %

We incorporated the learnings and improvements from \partir and GSPMD~\cite{gspmd2021} into Shardy~\cite{shardy}, a joint open-source project. See \Cref{sec:actual-related-work} for a brief comparison.

\section{Background}\label{sec:related}

\subsection{SPMD for ML workloads}
JAX~\cite{jax2018github} leverages the single program, multiple data (SPMD) paradigm for multi-device parallelism.
In this model different devices operate on different data slices using the same computation (e.g., different data batches in the forward pass of a NN). 
The process of placing different chunks of a tensor on different devices is known as sharding in the literature, and partitioning refers to the overall strategy of sharding the model.
In many cases communication is required across groups of devices -- for example a data-parallel NN needs the gradients from all devices to update its parameters. The SPMD model is hence employing MPI-style~\cite{MPI:1995} primitives for collective communication:
\begin{itemize}
    \item AllReduce(AR): combines a sharded tensor across devices to produce one replicated value using a reduction operation (e.g., sum).
    \item AllGather(AG): collects a sharded tensors from all devices into a full tensor.
    \item ReduceScatter(RS): applies a reduction (e.g., sum), then distributes partial results to each device.
    \item All2All(A2A): exchanges unique chunks of a distributed tensor between participating devices.
\end{itemize}

\subsection{Device meshes}
Distributed execution of NNs leverages the concept of a {\em mesh}, exposed in ML frameworks (e.g., \inlc{jax.mesh}~\cite{jax2018github}).
A mesh is an n-dimensional array with named axes that offers a logical view of the available devices.
A system of 16 devices may be viewed as a 2D mesh $\{a{:}2, b{:}8\}$; as a 3D mesh $\{a{:}2, b{:}2, c{:}4\}$; or as a 1D mesh $\{a{:}16\}$, among many others. 
In practice, the mesh structure reflects the system's communication topology, allowing for reasoning about performance and utilizing the fastest networks. 
For example, consider 4 servers connected by an Ethernet connection; each server has 8 GPUs connected by a fast interconnect~\cite{nvidia-nvlink}. 
The mesh $\{eth{:}4, ic{:}8\}$ makes it explicit which of the two networks is used for communication between devices.
This flexibility allows meshes to model a wide range of connectivity setups between devices. 
As a logical abstraction, meshes can model a physical tree topology (e.g.,  [N×M] represents a cluster of N hosts, each with M GPUs), or a physical mesh topology (like TPU pods), making it a general-purpose abstraction for SPMD and suitable for \partir{} abstractions.

\subsection{Parallelism strategies}\label{background:parallelism}

\begin{figure}
    \centering
\includegraphics[scale=0.5]{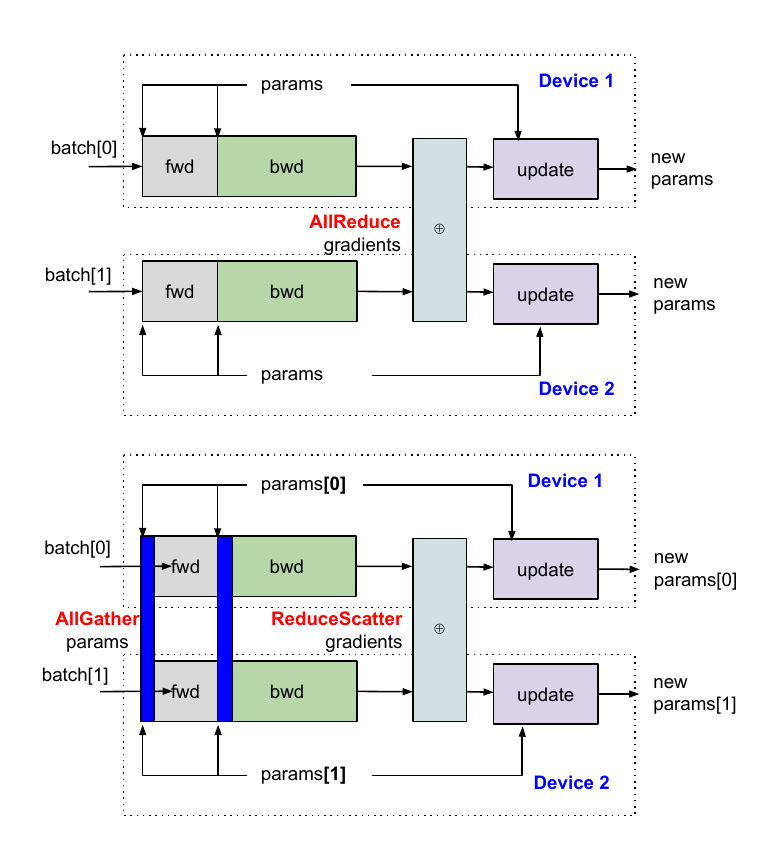}
\caption{
\textbf{Top}: batch parallelism, the gradients are AllReduced to update the parameters.
\textbf{Bottom}: Z3/FSDP, note the parameters are sharded and only AllGathered before their use, highlighted in thick blue bars in the figure. The gradients are ReducedScattered before updating the parameters.}
\label{fig:batch-vs-fsdp}
\end{figure}


\paragraph{Batch parallelism (BP)} The input batch is sharded across the devices while the model 
parameters and optimizer state are replicated. We expect to see one AllReduce operation per parameter in the backward pass, cf. \Cref{fig:batch-vs-fsdp}.
    
\paragraph{Model parallelism (MP)} Model parameters are sharded across the devices -- for example, in Megatron sharding for Transformer models~\cite{megatron2019, large_megatron_21, seq_paralleism_nvidia}, 2 AllReduce operations (on activation tensors) are introduced per Transformer layer (and 2 more for its backward pass).
  
\paragraph{Optimizer sharding} The optimizer state is sharded to reduce peak memory. 
Two variants exist: ZeRO2 (Z2) that additionally partitions the gradients, and ZeRO3 (Z3) that further partitions the parameters~\cite{zero2019, zero_offload2021, weight-update-sharding}. 
This paper uses an SPMD variant that shards all parameters, known as fully-sharded data parallelism (FSDP)~\cite{fsdp}.
Z3/FSDP (illustrated in \Cref{fig:batch-vs-fsdp}) reduce peak memory usage as the optimizer uses gradient shards and parameter/optimizer shards to update the parameter shards. These shards are distributed and gathered back when they are needed. 
From the perspective of SPMD collectives, every parameter is gathered once in the forward and once in the backward pass, using AllGathers. 
The gradients are gathered during the backward pass only using ReduceScatter -- a cheaper operation than AllReduce.
    
\begin{figure}
\centering
  \includegraphics[scale=0.3]{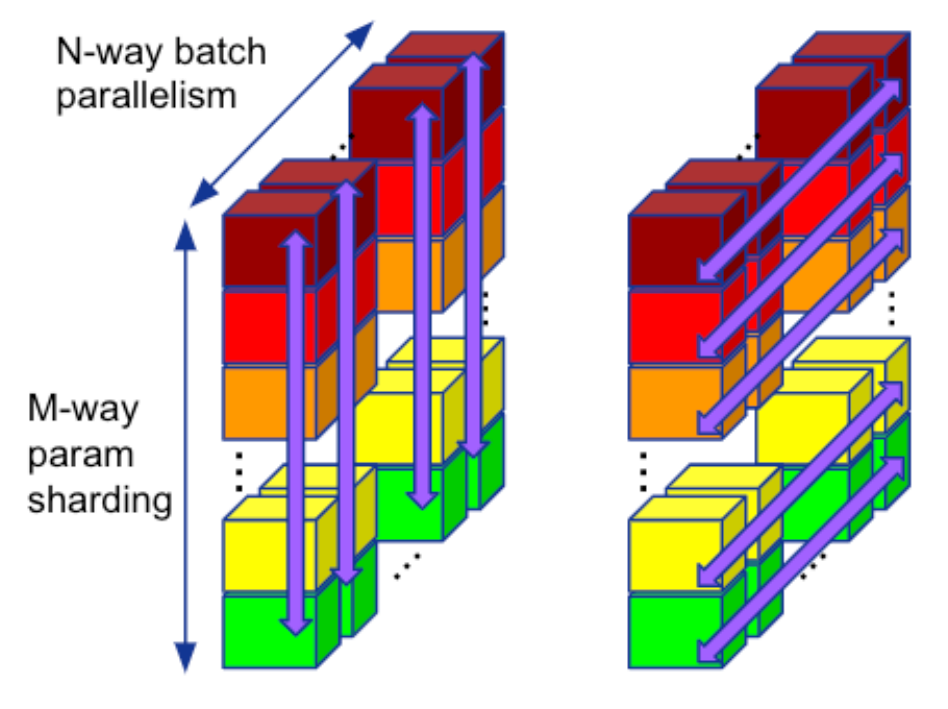}
    \caption{
    \label{fig:logical-partitioning}
    \small Batch (N) and model (M) parallelism. 
    On the left, the communication along the M axis (e.g., Megatron's~\cite{megatron2019} activation reductions).
    On the right, the communication along the N axis (e.g., gradient reductions). 
    Each device parameter shard is color-coded; all devices along N hold the same shard.
    }
\end{figure}

\paragraph{Combining strategies} 
Training large models efficiently requires combining multiple partitioning strategies across several axes, e.g., {\em BP+MP+Z3}~\cite{gopher2021, large_megatron_21}.
For example, \Cref{fig:logical-partitioning} shows a model partitioned over a 2D mesh with BP on one axis and MP over the other. 
There exist many other strategies, such as activation sharding after MP~\cite{collective-matmul, seq_paralleism_nvidia}, or Transformer's multi-query sharding~\cite{inference_transformer}. Many of these combinations were implemented and verified in \partir{}.

\subsection{Strategies as program transforms}\label{background:trace}
To demonstrate these partitioning strategies, consider a \inlc{JAX} program composed of two matrix multiplications\footnote{For simplicity,
we do not show a full training step with back-propagation, just a feed-forward function. We even skip the elementwise non-linear operators, as they are trivial to partition.}:
\begin{lstlisting}[language=python, 
% caption={A matmul chain in JAX.},
label={lst:jax_matmul}]
def f(x, w1, w2):
  return (x @ w1) @ w2
y = jax.jit(f)(input, w1, w2) # Trace, compile, run.
\end{lstlisting}
The \inlinecode{jax.jit()} function traces the Python function
into StableHLO~\cite{mlir2020} MLIR, the array IR encoding of XLA HLO~\cite{xla}, before compiling the function for a specific backend.
The StableHLO for the program above looks as follows:\footnote{StableHLO uses \texttt{dot\_general} to represent matrix multiplication. We use \texttt{matmul} as syntactic sugar in the paper for easier presentation.}
\begin{lstlisting}[language=mlir, 
caption={
An unpartitioned \texttt{matmul} chain, where each value is annotated with a type, e.g., \%x1 is an array of shape \inlinecode{256x8} and a \inlinecode{float32} element type.
},
label={lst:unpartitioned-chain}]
func @main(%x: tensor<256x8xf32>,
           %w1: tensor<8x16xf32>,
           %w2: tensor<16x8xf32>) {
  %x1 = matmul(%x,  %w1) : tensor<256x16xf32>
  %x2 = matmul(%x1, %w2) : tensor<256x8xf32>
  return %x2 : tensor<256x8xf32> 
}
\end{lstlisting}

Assuming this program executes on a 2D mesh $\{B{:}4,M{:}2\}$,
what kind of parallelization strategy should we use?
\paragraph{Batch (data) parallelism} One strategy is to partition the first (256-sized) dimension of input \inlinecode{\%x} across axis $B$ - resulting in a pure ``map'' over that dimension:
\begin{lstlisting}[language=mlir, caption={Data-parallel \texttt{matmul} chain.}, label={lst:data-parallel-chain}]
func @main(%x: tensor<(*@\codehl{pink}{64}@*)x8xf32>,
           %w1: tensor<8x16xf32>,
           %w2: tensor<16x8xf32>)
     attributes {mesh={"B":4, "M":2}} {
  %x1 = matmul(%x, %w1) : tensor<64x16xf32>
  %x2 = matmul(%x1, %w2) : tensor<64x8xf32>
  return %x2 : tensor<64x8xf32> 
}
\end{lstlisting}
The resulting device-local program above takes a first argument of smaller shape \inlinecode{64x8} since each device acts in parallel on a slice of \inlinecode{\%x} determined by the devices along axis $B$.
At the same time, the shape of parameters \ \inlinecode{\%w1} and \inlinecode{\%w2} remains constant across all devices. 

\paragraph{Adding model parallelism}
By further partitioning the parameter \inlinecode{\%w1} on  $dim=1$ and \inlinecode{\%w2} on  $dim=0$ along axis $M$, each device along axis $B$ may perform a smaller multiplication with a different parameter shard:
\begin{lstlisting}[language=mlir, caption={Data-parallel and sharded \texttt{matmul} chain.},
label={lst:partitioned-chain}]
func @main(%x: tensor<64x8xf32>,
           %w1: tensor<8x(*@\codehl{green}{8}@*)xf32>, 
           %w2: tensor<(*@\codehl{green}{8}@*)x8xf32>) attributes ... {
  %x1 = matmul(%x, %w1) : tensor<64x8xf32>
  %x2 = matmul(%x1, %w2) : tensor<64x8xf32>
  %x3 = all_reduce <"M"> %x2 : tensor<64x8xf32>
  return %x3 : tensor<64x8xf32> }
\end{lstlisting}
The first \texttt{matmul} is a map over the second dimension of \inlinecode{\%w1}, so no special care is necessary.
We only need to remember that \inlinecode{\%x1} is partitioned along its second dimension. For the second \texttt{matmul}, observe that both of its operands are partitioned along the contracting dimensions. Hence, an  \inlinecode{all_reduce} operation across axis $M$ recovers the original program semantics. 
This sharding of pairs of matrix multiplications is the essence of the Megatron sharding in Transformers~\cite{megatron2019}.

\paragraph{Adding fully sharded parameters}
Notice that the parameters are only sharded on axis $M$ (but not $B$) in \Cref{lst:partitioned-chain}. 
To further shard the parameters on dimensions 0 and 1, respectively, we would need to insert two \inlinecode{all_gather} operations before they are needed in the multiplications:
\begin{lstlisting}[language=mlir, caption={Data-parallel and fully sharded \texttt{matmul} chain.},
label={lst:fsdp-chain}]
func @main(%x: tensor<64x8xf32>,
           %w1: tensor<(*@\codehl{yellow}{2}@*)x8xf32>,
           %w2: tensor<8x(*@\codehl{yellow}{2}@*)xf32>)
     attributes {mesh = {"B":4, "M":2}} {
  %w1g = all_gather [{"B"},{}] %w1 : tensor<8x8xf32>
  %x1 = matmul(%x, %w1g) : tensor<64x8xf32>
  %w2g = all_gather [{}, {"B"}] %w2 : tensor<8x8xf32>
  %x2 = matmul(%x1, %w2g) : tensor<64x8xf32>
  %x3 = all_reduce <"M"> %x2 : tensor<64x8xf32>
  return %x3 : tensor<64x8xf32> }
\end{lstlisting}
These operations gather the shards on the corresponding dimensions.
This sharding of parameters,
after batch parallelism, is the essence of FSDP~\cite{zero2019, fsdp}. 
More complex shardings are possible on top of this schedule.
For example, sharding the input and output activation ({\tt \%x} and the return value) on the model axis $M$ will convert the \inlinecode{all_reduce} operation to a \inlinecode{reduce_scatter} and introduce an \inlinecode{all_gather}\xspace on the input {\tt \%x} before using it in the first \texttt{matmul}.
We omit the code but stress that this is the ES strategy in \Cref{sec:eval}.

The sequence of examples above highlights that (i) sharding strategies compose; (ii) by following algebraic reasoning, a model can be partitioned just by sharding its inputs and parameters, and sometimes internal operations, see \Cref{sec:tags};
(iii) it suffices to utilize information about parallel and contracting dimensions. In the rest of the paper we show
how to express sharding strategies through semantics-preserving rewrite actions to go from an unpartitioned program (as in \Cref{lst:unpartitioned-chain}) to a device-local program (as in \Cref{lst:data-parallel-chain,lst:partitioned-chain,lst:fsdp-chain}). 

\paragraph{A note on scale} 
A large ML model may contain +100k of tensors and operations on them (e.g., dot-products, convolutions, scatter ops, control-flow operations, and more). 
A good API should not burden the user with a decision per tensor or operation.

\section{A schedule is all you need}\label{ssec:schedule}
Users express their partitioning strategies in \partir{} using {\em tactics}. Conceptually, a tactic mirrors a strategy: it defines which values must be sharded and how (e.g., BP is achieved by sharding the data arrays on the batch dimension). Additionally, it defines propagation barriers that ensure conflict-free shardings (\Cref{sec:conflicts}).
The user can manually define these tactics or invoke automatic tools~\cite{automap, automap-reloaded, alpa2022}. 
A sequence of these tactics forms a schedule in an API inspired by work in kernel-generating DSLs~\cite{Taco2017,distal:pldi,Halide:PLDI:2013,tvm2018}.
The API itself is fairly minimal, and shown in \Cref{table:api}, yet it is powerful enough to express most sharding strategies.
For example, the following schedule achieves the FSDP sharding of~\Cref{lst:fsdp-chain}:
\begin{lstlisting}[caption={\small Partitioning strategies as a series of tactics.},captionpos=b,
escapeinside={*@}{@*},%
label={lst:schedule-simple}, language=python]
  # 1. Arrange devices in a BxM mesh.
  mesh = maps.mesh(device_array, ("B", "M"))
  # 2. Define sharding strategies as series of tactics.
  BP = ManualPartition({"x":0}, axis="B")
  MP = ManualPartition({"w1":1}, axis="M")
  Z3 = ManualPartition({"w1":0, "w2":1}, axis="B")
  schedule = [BP, MP, Z3]
  # 3. Partition and get distributed function \& metadata.
  dist_fn, metadata = *@{\partir}@*.jit(f, mesh, schedule)
\end{lstlisting}
The first tactic {\tt BP} partitions the first argument {\tt "x"} on dimension (DIM) 0 and across axis "B";
yielding batch parallelism (\Cref{lst:data-parallel-chain}). 
The second tactic {\tt MP} partitions input {\tt w1} on DIM $1$. The compiler will identify that this action shards the contracting DIM of a \texttt{matmul} and will shard {\tt w2} on DIM 0 through a process we call {\em propagation} invoked at the end of every tactic (\Cref{lst:partitioned-chain}).
The final tactic {\tt Z3} shards the parameters on the remaining available DIMs and axis $B$ (\Cref{lst:fsdp-chain}).
The function to partition, device mesh, and schedule are then finally passed to \partirjit{}, which works like \inlc{jax.jit} but goes through the \partir{} partitioning stack (\Cref{sec:architecture}) before compilation. 
\partirjit{} returns a partitioned module exposed as a Python callable, ready to be called with JAX-sharded arrays and executed on the devices.
Notably, \partir{}'s incrementality makes debugging easier.
\partir returns metadata containing debug information, sharding specifications of the function inputs and outputs produced by \partir{}, and every tactic's cost model estimates (e.g., SPMD collectives breakdown by type and simulation results).
Furthermore, users can inspect the rewritten module after every tactic, a natural consequence of the \partir{} incrementality, unlike the situation in other partitioners that rewrite the whole module at once~\cite{gspmd2021}.

\begin{table*}
\centering
\begin{tabular}{lll}
\toprule
\textbf{API} & \textbf{Inputs} & \textbf{Description} \\
\midrule
ManualPartition (tactic) & 
\begin{tabular}[t]{@{}l@{}}
  \texttt{inputs}: Dictionary \\
   - keys: function input names \\
   - values: dimension to shard \\
  \texttt{axis}: String (e.g., "batch", "model") \\
\end{tabular} & 
\begin{tabular}[t]{@{}l@{}}
Specifies the sharding dimension for each input \\
along the given axis. For example: \\
\texttt{(\{"D": 0, "w0": 1\}, axis="x")} \\
 Shards \texttt{D}'s 0th and \texttt{w0} 1st's dimensions on the "x" axis.
\end{tabular} \\

\midrule
AutomaticPartition (tactic) & 
\begin{tabular}[t]{@{}l@{}}
  \texttt{axes}: List of Strings (e.g., ["x", "y"]) \\
  \texttt{options}: Dictionary (passed to AUTO) \\
\end{tabular} & 
\begin{tabular}[t]{@{}l@{}}
Automatically determines the program sharding \\ over the specified axes.
\end{tabular} \\

\midrule
\partir.jit & 
\begin{tabular}[t]{@{}l@{}}
  \texttt{func}: The tensor program to partition \\ 
  \texttt{schedule}: List of \partir's tactics \\
  \texttt{kwargs}:  Dictionary \\
  - Passed to \inlc{jax.jit}\cite{jax2018github}.
\end{tabular} & 
\begin{tabular}[t]{@{}l@{}}
The entry point to \partir{}. It returns: \\ 
- The partitioned program. \\
- Analytical performance estimates after each tactic.\\
- Inserted collectives after each tactic.
\end{tabular} \\

\bottomrule
\end{tabular}
\caption{
\label{table:api}\small 
\partir{} Python API.
}
\end{table*}

\paragraph{Mixing automatic and manual tactics} 
\partir{} exposes an {\tt AutomaticPartition} tactic that operates on one or more mesh axes and composes with other tactics. 
For example, users could use a manual tactic to introduce batch parallelism manually and rely on an automatic tactic to discover partitioning along the second axis "M": 
\begin{lstlisting}[
caption={\small Composing manual and automatic tactics.},
captionpos=b, 
label={lst:schedule}, language=python,escapeinside={*@}{@*}]
  BP = ManualPartition({"x":0}, axis="B")
  AutoMP = AutomaticPartition(axis="M")
  part_fn, _ = *@{\partir}@*.jit(fn, mesh, [BP, AutoMP])
\end{lstlisting}
{\tt AutomaticPartition} is an interface for any optimization algorithm.
We implemented a Monte Carlo tree search for discovering partitioning strategies~\cite{automap, automap-reloaded}, using a cost model that seeks runtime improvement.


What is essential for composability is that both manual and automatic tactics issue sequences of (the same) lower-level \partir compiler {\em actions} that either (i) shard a value dimension along an axis or (ii) explicitly keep a value replicated across a mesh axis, or (iii) propagate sharding information in a module. 
For example, the schedule from \Cref{lst:schedule-simple} generates a sequence of 7 \partir{} actions:
\begin{lstlisting}[language=mlir]
tile<%x,0,"B">; propagate // BP tactic 
tile<%w1,1,"M">; propagate // MP tactic
tile<%w1,0,"B">; tile<%w2,1,"B">; propagate // Z3 
\end{lstlisting}

Next, we present the \partir system architecture, the implementation of actions as program rewrites, and how \partir eventually generates device-local SPMD code.

\section{System architecture}\label{sec:architecture}
\begin{figure}[H]
  \centering
  \includegraphics[width=0.4\textwidth]{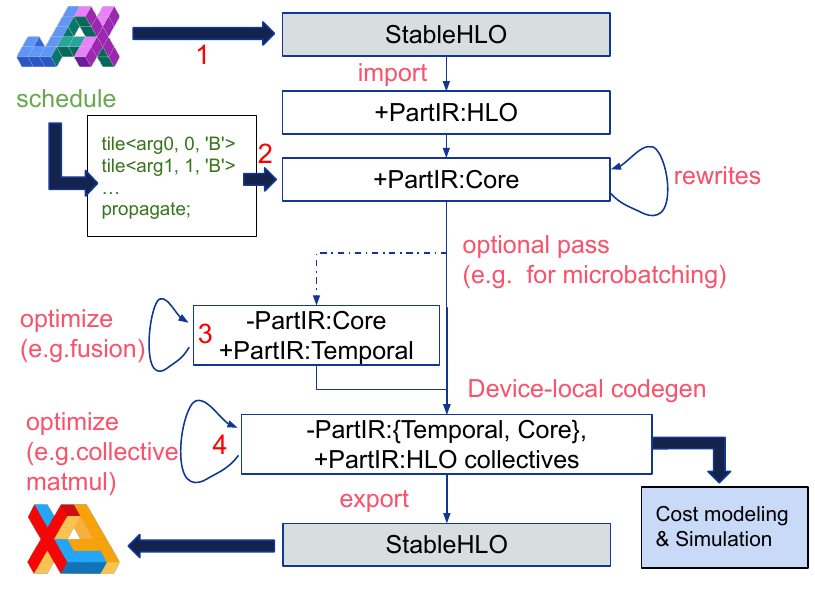}
    \caption{
    \label{fig:part_stack}
    \small \partir{} partitioning stack, built using MLIR, supporting layering of new operators on top of existing ones -- hence, we use "+" to signify the introduction of new operators, and "-" to signify that operators have now become illegal.
    }
\end{figure}

\partir{} partitioning is done at the MLIR level through MLIR-based rewriting passes illustrated in \Cref{fig:part_stack}, following the highlighted number on the figure:

\begin{enumerate}

\item Programs are generated from tracing functions (\Cref{background:trace}) into the StableHLO~\cite{stablehlo} dialect.
 
\item Manual or automatic tactics invoke sequences of compiler
actions that introduce and propagate {\em functional loops} and
specialized {\em slicing} ops, that belong in the \partir:Core dialect. 

\item An optional pass to lower to the \partir:Temporal dialect that is used for niche applications like automatic micro-batching by referencing the semantics of \partir:Core.
We omit further details to focus on SPMD.

\item \partir:Core ops are lowered to the \partir:HLO dialect generating device-local collective communication ops. 
The collectives in this dialect refer to mesh axes that make their IR encoding independent of the total number of devices in the mesh (as opposed to collectives in StableHLO and XLA:HLO that reference groups of logical device IDs) and make it easy to reason about and fuse (\Cref{sec:mhlo}).
Simulators are also implemented at this level. 
To export, we lower any custom high-level \partir:HLO ops to StableHLO computations and hand the module to XLA for compilation.
\end{enumerate}

Our architecture allows us to implement the right rewrites at the right abstraction level and independently test various internal dialects.
For example, \partir:Core is unaware of SPMD execution, and its rewrite axioms remain very simple; this is a separate step from SPMD lowering
(a pass that we have additionally proven correct), or the optimization of SPMD primitives.
Additionally, the input and output of our system is the StableHLO dialect, making \partir a frontend- and backend-agnostic tool.%
\footnote{We only present APIs for JAX since it is our users' frontend tool of choice.
Typical choices for backends are XLA, CUDA or OpenXLA.}

We have formally proven the correctness of the most complex part of the \partir stack, the lowering to \partir:SPMD, and described the process in detail in  \Cref{app:sec:translation-correctness}.
The rest of the stack is compartmentalized and tested extensively.

\section{\partir{}:Core}\label{sec:core}

\partir:Core introduces two operations on top of StableHLO: 1) a \coreloop op that expresses pure (parallel) loops performing tiling or reductions, and 2) a \coreslice op that extracts a tensor slice based on a \coreloop index. 
We will present these constructs and their semantics 
as part of describing the \partir:Core compiler {\em actions}. 



\subsection{Value tiling action}\label{ssec:tiling}
A tiling action \inlc{tile<\%value, dim, axis>} creates a loop that in each iteration yields a slice of 
\inlc{\%value} along dimension \inlc{dim}. For example, value tiling
\inlc{\%x} along dimension 0 and axis "B" from \Cref{lst:unpartitioned-chain} - \inlc{tile<\%x, 0, B>} - produces:
\begin{lstlisting}[language=mlir, label={lst:dumb-tiled-argument}]
func @f(%x: tensor<256x8xf32>,
        %w1: tensor<8x16xf32>,
        %w2: tensor<16x8xf32>) {
  %xt = loop "B" [#tile<0>] (%rB: range<4>) {
    yield (slice 0 %x[%rB]) : tensor<64x8xf32>
  } : tensor<256x8xf32>
  %x1 = matmul(%xt, %w1)
  %x2 = matmul(%x1, %w2)
  return %x2 : tensor<256x8xf32> }
\end{lstlisting}

The \coreloop operation contains two static attributes: (i) a mesh axis (\inlinecode{"B"}) and (ii) an {\em action} attribute 
(\inlinecode{#tile<0>}).
It also accepts a single-argument closure (a {\em region} in the MLIR jargon) that represents the loop body: \inlinecode{(\%rB: range<4>) \{ ... \}}.
The closure takes as an argument a {\em range value} (\inlc{\%rB}) and performs a tensor computation returning a value of type \inlinecode{tensor<64x8xf32>}. The range argument \inlc{\%rB} plays the
role of a loop index that has a \partir-specific \corerange type.

The \coreslice ops consume these loop indices.
The meaning of \inlinecode{slice 0 \%x[\%rB]} is that it extracts the \inlinecode{\%rB}-th chunk of the tensor {\tt \%x} along dimension 0.
The tiling here perfectly partitions the tensor dimension into $4$ equally-sized, contiguous chunks since axis \inlinecode{"B"} has size $4$.
Hence, the result of \coreslice has shape \inlinecode{64x8}.
Furthermore, the value tiling action has replaced \inlinecode{\%x} of
type \inlinecode{tensor<256x8xf32>} with value \inlinecode{\%xt} of
the same type.
Value tiling is a type-preserving, and also semantics-preserving local rewrite.



\subsection{Propagation action}
Value tiling actions create fairly trivial loops, 
making them not particularly interesting.
However, they help bootstrap a powerful {\em propagation}
pass that subsequently creates loops around {\em operations consuming or producing these values} and further slices other function arguments. 
This propagation is justified by program equivalences.

\subsubsection{Program equivalences} \label{ssec:equivalences}
\partir propagation is built around program equivalences that involve \coreloop and \coreslice instructions.
\Cref{fig:matmul-examples} presents three admissible program equivalences for a matrix multiplication. The first two rewrite a matrix multiplication as
a tiling loop with a smaller multiplication inside. The last one
introduces the \inlc{#sum} action attribute accompanying the loop.
This signifies that the results of each
iteration of the loop should be reduced, as
the operands are sliced on their contracting dimension.\footnote{\partir{} supports custom reductions for any associative reduction function.} 

To allow us to implement the rewriting code once for all
operators, we use a {\em tile-mapping registry} (TMR).
The TMR contains, for every tensor operation with $n$ inputs, a set of specifications of the form
\[t_1^\bot,\ldots,t_n^\bot \hookrightarrow \sigma_1,\ldots,\sigma_k \]
where $t^\bot$ stands for an optional tiling action, while $\sigma$ stands for an arbitrary action (including \inlc{#sum}). StableHLO ops and our loops may return multiple results, hence the generalized form $\sigma_1,\ldots,\sigma_k$.
Each such specification asserts that a given operation can be rewritten as a loop with action(s) $\sigma_1,\ldots,\sigma_k$ if we slice its operands according to $t_1^\bot,\ldots,t_n^\bot$ (a missing action implies no slicing).
For example, these are the entries for \inlc{matmul} (corresponding to the equivalences from Figure~\ref{fig:matmul-examples}) and for an element-wise \opadd operation that asserts that tiling its result requires tiling its operands in the same way:
\[\small\begin{array}{lcl}
\tmr(\opmatmul) & = & \{ (\shorttileaction{0},\bot) \hookrightarrow \shorttileaction{0} \} \\
               & \cup & \{ (\bot, \shorttileaction{1}) \hookrightarrow \shorttileaction{1} \} \\
               & \cup & \{ (\shorttileaction{1},\shorttileaction{0}) \hookrightarrow \shortsumaction\} \\
\tmr(\opadd) & = & \{ (\shorttileaction{d},\shorttileaction{d})  \hookrightarrow \shorttileaction{d} \} 
\end{array}\]
It turns out that this abstraction (similar to {\em split annotations}~\cite{palkar+:splitannotations}) is sufficient to capture a wide variety of equivalences, and for substantially more complex ops, e.g. convolutions, scatter and gather, reshapes, and others.

\begin{figure}
\begin{subfigure}{1\textwidth}
\begin{lstlisting}[language=mlir]
%t = loop "B" [#tile<(*@\codehl{yellow}{0}@*)>] (%rB:range<4>) {
  %xs = slice (*@\codehl{green}{0}@*) %x[%rB]
  %z = matmul(%xs, y)
  yield %z : tensor<8x8xf32> 
}
\end{lstlisting}
\end{subfigure}
\hfill
\begin{subfigure}{1\textwidth}
\begin{lstlisting}[language=mlir]
%t = loop "B" [#tile<(*@\codehl{yellow}{1}@*)>] (%rB:range<4>) {
  %ys = slice (*@\codehl{green}{1}@*) %y[%rB]
  %z = matmul(%x, %ys)
  yield %z : tensor<32x2xf32> 
}
\end{lstlisting}
\end{subfigure}
\hfill
\begin{subfigure}{1\textwidth}
\begin{lstlisting}[language=mlir]
%t = loop "B" [#sum] (%rB:range<4>) {
  %xs = slice (*@\codehl{green}{1}@*) %x[%rB]
  %ys = slice (*@\codehl{green}{0}@*) %y[%rB]
  %z = matmul(%xs, %ys)
  yield %z : tensor<32x8xf32> 
}
\end{lstlisting}
\end{subfigure}
\caption{
Programs equivalent to \inlc{\%t = matmul(\%x, \%y)},
where \inlc{"B"} is a mesh axis of size 4, 
assuming \inlc{\%x : tensor<32x16xf32>} and
\inlc{\%y : tensor<16x8xf32>}.
}\label{fig:matmul-examples}
\end{figure}

\subsubsection{Propagation pass}\label{ssec:propagation}
Propagation is a pass that greedily propagates {\em known} and {\em partially known} information and introduces more loops, based on the TMR specifications.


\paragraph{Propagation of known information}
{\em Forward} propagation searches for an entry that matches the actions of \coreloop{s} that produce operands of an operation, whereas {\em backward} propagation searches for an entry that matches the way the operation result is sliced downstream.
For example, assume that {\tt \%x1} in \Cref{lst:unpartitioned-chain} has been tiled:
\begin{lstlisting}[language=mlir]
func @main(%x: tensor<256x8xf32>,
           %w1: tensor<8x16xf32>,
           %w2: tensor<16x8xf32>) {
  %x1 = matmul(%x,  %w1)
  // value tiling
  %x1t = loop "B" [#tile<0>] (%rB: range<4>) {
    yield (slice 0 %x1[%rB]) 
  }
  %x2 = matmul(%x1t, %w2)
  return %x2 : tensor<256x8xf32> 
}
\end{lstlisting}%
To propagate tiling forward observe that the \opmatmul defining \inlc{\%x2} takes an operand whose first dimension is tiled, matching the \opmatmul's TMR entry $(\shorttileaction{0},\bot) \hookrightarrow \shorttileaction{0}$ operand context.
Propagating backward, \inlc{\%x1} is produced by a \opmatmul and is then {\em sliced} along dimension 0, which matches the result of that same TMR entry.
Therefore, both \texttt{matmul} operations can be rewritten:
\begin{lstlisting}[language=mlir]
func @main(%x: ..., %w1: ..., %w2: ...) {
  // result of backward propagation
  %x1 = loop "B" [#tile<0>] (%rB: range<4>) { 
    yield (matmul(slice 0 %x[%rB], %w1)) 
  }
  // value tiling
  %x1t = loop "B" [#tile<0>] (%rB: range<4>) {
    yield (slice 0 %x1[%rB]) 
  }
  // result of forward propagation
  %x2 = loop "B" [#tile<0>] (%rB: range<4>) { 
    yield (matmul(slice 0 %x1t[%rB], %w2)) 
  }
  return %x2 : tensor<256x8xf32> 
}
\end{lstlisting}
Through propagation, we have arrived at a program where every operation is within a loop context. These loops may be interpreted sequentially in \partir:Temporal or lowered to SPMD in \partir:HLO.

Here is what the fused program looks like:
\begin{lstlisting}[
    language=mlir,
    caption={Chained \texttt{matmul} with a tiling loop.},    
    label={lst:Core-chain-matmul-tiled-on-batch}
]
func @main(%x: ..., %w1: ..., %w2: ...) {
  %r = loop "B" [#tile<0>] (%rB : range<4>) {
    %xs  = slice 0 %x[%rB] : tensor<64x8xf32>
    %x1s = matmul(%xs,  %w1)
    %x2s = matmul(%x1s, %w2)
    yield %x2s : tensor<64x8xf32>
  }
  return %r : tensor<256x8xf32>
}
\end{lstlisting}

\paragraph{Inference from partially known information}
Inference is the process of deducing missing operand value tiling based on a {\em partial} match against a TMR entry.
Continuing from Listing~\ref{lst:Core-chain-matmul-tiled-on-batch}, consider value-tiling \inlc{\%w2}:
\begin{lstlisting}[language=mlir]
func @main(%x: ..., %w1: ..., %w2: ...) -> ... {
  %w2t = loop "M" [#tile<0>] (%rb: range<2>) {
    yield (slice 0 %x2[%rb]) 
  }
  %x2 = loop "B" [#tile<0>] (%rB: range<4>) {
    %xs = slice 0 %x[%rB]
    %x1s = matmul(%xs,  %w1)
    %x2s = matmul(%x1s, %w2t)
    yield %x2s
  }
  return %x2 : tensor<256x8xf32> 
}
\end{lstlisting}
The TMR entry ${\shorttileaction{1},\shorttileaction{0}) \hookrightarrow \shortsumaction}$ is a partial match on the operands of the second \opmatmul, since the second operand (\inlc{\%w2t}) is already tiled.
We can extend it into a full match by value tiling the first operand (\inlc{\%x1s}) and then continuing with propagation to eventually
(after some simplification) arrive at: 
\begin{lstlisting}[
    language=mlir,
    label={lst:Core-tiled-chain-matmul-propagation}
]
func @main(%x: ..., %w1: ..., %w2: ...) -> ... {
  %r = loop "B" [#tile<0>] (%rB: range<4>) {
    %x2s = loop "M" [#sum] (%rM: range<2>) {
      %xs  = slice 0 %x[%rB]  : tensor<64x8xf32>
      %w1s = slice 1 %w1[%rM] : tensor<8x8xf32>
      %x1ss = matmul(%xs, %w1s) : tensor<64x8xf32>
      %w2s  = slice 0 %w2[%rM] : tensor<8x8xf32>
      %x2ss = matmul(%x1ss, %w2s)
        : tensor<64x8xf32>
      yield %x2ss : tensor<64x8xf32>
    }
    yield %x2s : tensor<64x8xf32>
  }
  return %r : tensor<256x8xf32>
}
\end{lstlisting}
Notice how in that final program, both \inlc{\%w1} and \inlc{\%w2} end up only used {\em sliced} across axis \inlc{"M"}, even though only \inlc{\%w2} was explicitly value-tiled. 

Inference is very important in ML programs. For example, it can identify that parameters and optimizer states are partitioned similarly, as they participate in element-wise operations during the parameter update.

\subsubsection{Conflicts during propagation}\label{sec:conflicts}

In some situations, it is impossible to propagate the tiling. 
For example, when a loop over an axis needs to be inserted in the {\em scope} of an existing loop over the same axis, which we forbid because nested loops along the same axis cannot be mapped to meshes, or when multiple (partial) TMR matches are found, a situation that we refer to as a {\em conflict}. Consider:%
\begin{lstlisting}[language=mlir]
func @main(%x: ..., %w1: ...) {
  %xt = loop "B" [#tile<0>] (%rB: range<4>) {
    yield (slice 0 %x[%rB]) 
  }
  %w1t = loop "B" [#tile<1>] (%rB: range<4>) {
    yield (slice 1 %w1[%rB]) 
  }
  %x1 = matmul(%xt, %w1t)
  ...
\end{lstlisting}%
Here, the operands of the \opmatmul defining \inlc{\%x1} are tiled in a way that matches two TMR entries: 
$(\shorttileaction{0},\bot) \hookrightarrow \shorttileaction{0}$ and 
$(\bot,\shorttileaction{1}) \hookrightarrow \shorttileaction{1}$.

\partir will not attempt to resolve conflicts automatically. 
Instead, the canonical solution is to perform the rewriting \emph{incrementally}.
For example, performing value tiling on \inlc{\%x} and propagating that choice \emph{before} value tiling \inlc{\%w1} would yield the following program:

\begin{lstlisting}[
    language=mlir]
func @main(%x:..., %w1: ...) {
  %w1t = loop "B" [#tile<1>] (%rB: range<4>) {
    yield (slice 1 %w1[%rB]) }
  // below is result of propagation after tiling \%x
  %r = loop "B" [#tile<0>] (%rB : range<4>) { 
    %xs  = slice 0 %x[%rB] : tensor<64x8xf32>
    %x1s = matmul(%xs,  %w1t)
    ...
\end{lstlisting}
At this point the TMR entry $(\bot,\shorttileaction{1}) \hookrightarrow \shorttileaction{1}$ matches the definition of \inlc{\%x1s} again. 
Alas, the operation in hand is {\em already nested} inside a loop over axis \inlc{"B"} and no further propagation is possible --
creating a doubly-nested loop over \inlc{"B"} is invalid.
This prioritization of BP over subsequent parameter sharding is exactly what is needed for the ZeRO~\cite{zero2019} sharding strategies. 

The prioritization of rewrites, happening naturally at the boundaries of
\partir manual tactics, makes conflicts fairly rare, and as a result dramatically reduces the need for many internal sharding decisions, but does not entirely remove their need; see \Cref{sec:tags}.
Finally, our tiling and propagation actions naturally extend to loop nests over multiple axes, see \Cref{appendix:core}. 



\section{SPMD code generation}\label{sec:mhlo}
\begin{figure*}[!t]
\begin{lstlisting}[caption={\small \partir:HLO collectives by example, with relevant dimensions and attributes (e.g. slicing/gathering axes) highlighted.}, language=mlir, label={lst:mhlo-collectives}, numbers=left]
  // Below, assume mesh: \{x1 : 2, x2 : 4, x3 : 8\}
  (*@\label{line:all_reduce}@*)%rst = all_reduce<@red_fn> <"x1", "x2"> %operand : tensor<16x5x40xf32> -> tensor<16x5x40xf32>
  (*@\label{line:all_slice1}@*)%rst = all_slice [{(*@\codehl{pink}{"x1"}@*)}, {}, {(*@\codehl{yellow}{"x2"}@*)}] %operand : tensor<(*@\codehl{pink}{16}@*)x5x(*@\codehl{yellow}{40}@*)xf32> -> tensor<(*@\codehl{pink}{8}@*)x5x(*@\codehl{yellow}{10}@*)xf32>
  (*@\label{line:all_slice2}@*)%rst = all_slice [{(*@\codehl{pink}{{"x1, "x3"}}@*)}, {}, {(*@\codehl{yellow}{"x2"}@*)}] %operand : tensor<(*@\codehl{pink}{16}@*)x5x(*@\codehl{yellow}{40}@*)xf32> -> tensor<(*@\codehl{pink}{1}@*)x5x(*@\codehl{yellow}{10}@*)xf32>
  (*@\label{line:all_gather}@*)%rst = all_gather [{(*@\codehl{pink}{{"x1"}}@*)}, {(*@\codehl{yellow}{{"x3"}}@*)}, {}] %operand : tensor<(*@\codehl{pink}{8}@*)x(*@\codehl{yellow}{10}@*)x16xf32> -> tensor<(*@\codehl{pink}{16}@*)x(*@\codehl{yellow}{80}@*)x16xf32>
  (*@\label{line:reduce_scatter}@*)%rst = reduce_scatter<@red_fn> [{(*@\codehl{pink}{"x1"}@*)}, {}, {(*@\codehl{yellow}{"x2"}@*)}] %operand : tensor<(*@\codehl{pink}{16}@*)x5x(*@\codehl{yellow}{40}@*)xf32> -> tensor<(*@\codehl{pink}{8}@*)x5x(*@\codehl{yellow}{10}@*)xf32>
  (*@\label{line:all_to_all}@*)%rst = all_to_all {(*@\codehl{pink}{0}@*) -> (*@\codehl{yellow}{1}@*)} <"x1","x2"> %operand : tensor<(*@\codehl{pink}{16}@*)x(*@\codehl{yellow}{32}@*)xf32> ->s tensor<(*@\codehl{pink}{128}@*)x(*@\codehl{yellow}{4}@*)xf32>
\end{lstlisting}
\end{figure*}

\partir:HLO extends StableHLO with ops that express per-device SPMD computation using specialized collective ops for communication. Unlike their low-level HLO counterparts, \partir's ops operate on {\em mesh axes}, i.e. communication spans across device groups defined by coordinates along mesh axes.
We define these collectives by example in \Cref{lst:mhlo-collectives}: 

\begin{figure}
    \centering
    \includegraphics[scale=0.45]{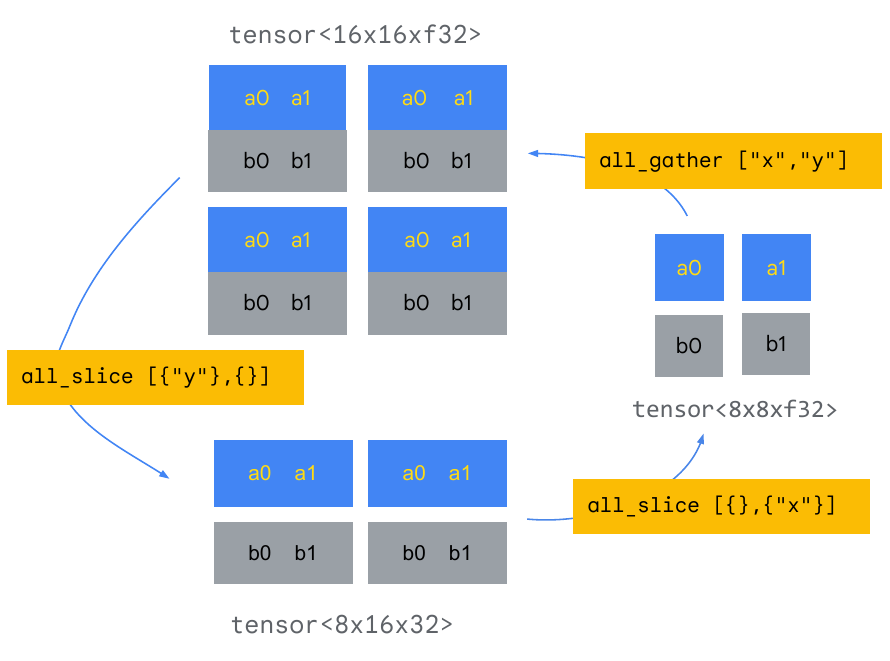}
    \caption{
    \small
    Demonstration of sequences of \inlc{all\_slice}
    and \inlc{all\_gather} collectives on a mesh of four devices \inlc{\{x:2, y:2\}}, each device is represented as a box in the figure. 
    Top: all devices hold the same 2D array;
    bottom: data is sliced row-wise along axis "y";
    right: data is further sliced column-wise along axis "x".
    In each case we give the device-local tensor types.
    }
    \label{fig:collectives}
\end{figure}

\begin{itemize}
    \item \inlc{all\_reduce} (\Cref{line:all_reduce}) reduces a tensor along one or more mesh axes (using reduction function \inlc{@red_fn}), then replicates it to all devices.
    \item \inlc{all\_slice} (Lines \ref{line:all_slice1},\ref{line:all_slice2}) takes an array of {\em axes per dimension} -- the axes in which each dimension is sliced
    In \cref{line:all_slice1}, the first dimension is sliced along "x1", the second dimension is not sliced, and the third dimension is sliced along "x2". 
    In the result array, each dimension size is divided by the size of the slicing axes in this dimension -- each device holds a slice of the original array.
    \item \inlc{all\_gather} (\Cref{line:all_gather}) gathers its operand along the array of axes in each dimension, dual to \inlc{all\_slice}. 
    Each dimension size of the result is multiplied by the gathering axes in this dimension. 
\end{itemize}
The (dual) semantics of \inlc{all\_slice} and \inlc{all\_gather} are demonstrated in \Cref{fig:collectives}.

Other collectives are produced by fusion passes:
\begin{itemize}
    \item A \inlc{reduce_scatter} (\Cref{line:reduce_scatter}) is produced by fusing \inlc{all\_reduce} (\Cref{line:all_reduce}) with \inlc{all\_slice} (\Cref{line:all_slice1}).
    \item An \inlc{all_to_all} (\Cref{line:all_to_all}) results by fusing  \inlc{all\_gather} along a dimension (0) with \inlc{all\_slice} over the same sequence of axes in another dimension (1).
\end{itemize}

\subsection{From \partir:Core to \partir:HLO}
Lowering of \partir:Core to \partir:HLO is a type-preserving transformation and formally proven correct (cf.~\Cref{app:sec:translation-correctness}).
It flattens \partir:Core \inlc{loop} instructions by replacing
(i) \inlc{slice}s with
\inlc{all\_slice} collective operations, and
(ii) inserting \inlc{all\_gather}/\inlc{all\_reduce} collectives on the results of \inlc{loop}s.
Thus,
\begin{lstlisting}[
    language=mlir,
    label={lst:core-lowering}
]
func @f(%x: tensor<256x8xf32>,
        %w1: tensor<8x16xf32>,
        %w2: tensor<16x8xf32>) {
  %r = loop "B" [#tile<0>] (%rB: range<4>) {
    %x2s = loop "M" [#sum] (%rM: range<2>) {
      %xs = slice 0 %x[%rB]  : tensor<64x8xf32>
      %w1s = slice 1 %w1[%rM] : tensor<8x8xf32>
      %x1ss = matmul(%xs, %w1s) : tensor<64x8xf32>
      %w2s = slice 0 %w2[%rM] : tensor<8x8xf32>
      %x2ss = matmul(%x1ss, %w2s) :
        tensor<64x8xf32>
      yield %x2ss : tensor<64x8xf32> }
    yield %x2s : tensor<64x8xf32> }
  %s = loop "B" [#tile<0>] (%rB: range<4>) {
     %rs = slice 0 %r[%rB]  : tensor<64x8xf32>
     %xs = slice 0 %s[%rB] : tensor<64x8xf32>
     %w = yield(add %rs %xs) : tensor<64x8xf32> }
  return %s : tensor<256x8xf32> 
}
\end{lstlisting}
is lowered to
\begin{lstlisting}[
    language=mlir,
    label={lst:core-lowering-mhlo}
]
func @f(%x: tensor<256x8xf32>,
        %w1: tensor<8x16xf32>,
        %w2: tensor<16x8xf32>) {
  // First loop nest.
  %xs1 = all_slice [{"B"},{}] %x : tensor<64x8xf32>
  %w1s = all_slice [{},{"M"}] %w1 : tensor<8x8xf32>
  %x1ss = matmul(%xs1, %w1s) : tensor<64x8xf32>
  %w2s = all_slice [{"M"},{}] %w2 : tensor<8x8xf32>
  %x2ss = matmul(%x1ss, %w2s) : tensor<64x8xf32>
  %x2s = all_reduce <"M"> %x2ss : tensor<64x8xf32>
  %r = all_gather [{"B"},{}] %x2s : tensor<256x8xf32>
  // Second loop nest.
  %rs = all_slice [{"B"}, {}] %r : tensor<64x8xf32>
  %xs2 = all_slice [{"B"}, {}] %x : tensor<64x8xf32>
  %w = add %rs %xs2 : tensor<64x8xf32>
  %s = all_gather [{"B"}, {}] %w : tensor<256x8xf32>
  return %s : tensor<256x8xf32>
}
\end{lstlisting}
The \inlc{all\_slice(all\_gather)(\%xs2)} in this example is fused away. 
Additionally, function arguments used by \inlc{all\_slice} operations and results produced by \inlc{all\_gather}s are converted to device-local arrays:
\begin{lstlisting}[
    language=mlir,
    label={lst:core-lowering-mhlo-optimized}
]
func @f(%xs: tensor<64x8xf32>,
        %w1s: tensor<8x8xf32>,
        %w2s: tensor<8x8xf32>) {
  // First loop nest.
  %x1ss = matmul(%xs, %w1s) : tensor<64x8xf32>
  %x2ss = matmul(%x1ss, %w2s) : tensor<64x8xf32>
  %x2s = all_reduce <"M"> %x2ss : tensor<64x8xf32>
  // Second loop nest.
  %w = add %x2s %xs : tensor<64x8xf32>
  return %w : tensor<64x8xf32> 
}
\end{lstlisting}%
After these transformations, \partir applies collective ops optimizations and additional rewrites to enable compute-communication overlap~\cite{collective-matmul}. 

\paragraph{Lowering proof sketch}
While we provide a detailed proof in \Cref{app:sec:translation-correctness}, here we give only a high-level overview of the proof.
In \partir{}:Core, the source language, tensors are regular multi-dimensional arrays.
Apart from tensor-valued variables, the source language also has range variables that act as (tiling) loop indices.
In \partir{}:SPMD, the target language of our lowering, tensors are maps from the device mesh to regular arrays.
A point in the device mesh is identified by an index tuple. There are only tensor-valued variables in the target language.
Expressions in the source language are evaluated in an environment that maps tensor and range variables to values.
To relate evaluation of expressions to the target language, one abstracts over the range variables.
This turns the source level expressions into maps from the device mesh to arrays, the same as target language expressions.
To show that a range-variable-abstracted source language expression agrees with the corresponding target language expression, one proceeds by an induction argument that follows the structure of our lowering function.

\section{Evaluation}\label{sec:eval}



We evaluate \partir{} on: 1) whether its partitioning of models achieves SOTA performance across different systems~(\Cref{eval:sota}); 2) whether it is predictable~(\Cref{eval:debug}); 3) whether it composes tactics, manual and automated~(\Cref{eval:auto}); 4) whether its propagation reduces the number of sharding decisions for users, and resolves conflicts~(\Cref{eval:conflict}); 5) whether it has small overhead~(\Cref{eval:overhead}).
    
\subsection{Benchmarking setup}\label{eval:setup}
The benchmark uses a range of JAX models for training and inference, the training models use Adam optimizer~\cite{adam}.

\begin{description}
\item[U-Net] A model variant used in the reverse process of a diffusion model~\cite{unet_diffusion}.
It uses 9 residual blocks for the down-sampling convolutions and 12 for up-sampling, and between them, there are two residual blocks and one attention layer with 16 heads.
\item[GNS] A Graph Network Simulator~\cite{sanchez2020learning} model used in molecular property prediction~\cite{godwin2022simple}, configured with 5 MLP layers of hidden size 1024. The network is configured with 24 message-passing steps and a latent size of 512. Each graph contains 2048 nodes and varying number of edges between 8192 and 65536.
\item[T32] A 32-layer Transformer based on Chinchilla~\cite{chinchilla2022},
with an additional normalization layer,
configured with input batch size 48, 32 heads and $d_{model}=4096$.
Vocabulary size for the embedding is the standard 32k.

\item[T48] A variant of T32 scaled up to 48 layers and configured with input batch size 64, 64 heads and $d_{model}=8192$.
Vocabulary size is again the standard 32k.
\end{description}

We measured the performance on Nvidia A100 and TPUv3:
\begin{itemize}
    \item \textbf{Nvidia A100}~\cite{a100} using the 40GB memory HBM2 version; it performs 156 TFLOPS on float32 and 312 TFLOPS for bfloat16. 
    The A100s are connected using Nvidia's NVLink, capable of 600GB/s data transfer. 
\item \textbf{TPUv3}~\cite{tpu_cloud} each TPU chip comes with two tensor cores. Each core has 16GiB HBM2 memory capacity, each capable of 61.5 TFLOPs on float32 and 123 TFLOPs on float16.
The chips are connected over four links, each capable of doing 70GB/s data transfer~\cite{tpuv4}.

\end{itemize}

In the figures, we report the average (training or inference) step-time from collecting runtime measurements. First we perform a warm-up step, followed by $10$ restarts of $100$ steps each, the process repeated three times. During measurement, we switch off XLA rematerialization to avoid its added noise.


\subsection{Partitioned models match SOTA} \label{eval:sota}
We validated \partir{}'s partitioned models performance comparable to that of GSPMD~\cite{gspmd2021} in terms of Model FLOPS Utilization (MFU) \cite{palm2022} and High-Bandwidth Memory (HBM) usage, by training the T32 and T48 models on different hardware configurations.
We partitioned the T32 model on two configurations: 1) 32 TPUs, and 2) 16 GPUs, while the T48 partitioned over 128 TPUs.
Both models used a \partir{} schedule of four tactics (\inlinecode{BP+MP+Z3+EMB}, see: \Cref{appendix:schedules}), and relied on equivalent sharding annotations for GSPMD. 
The results in \Cref{table:mfu} show that the performance of \partir{} is on par with that of GSPMD, with negligible differences between configurations ($\pm 1\%$). The HBM usage is closely comparable.
This is a positive results for \partir{}:
(i) our system has a simpler API,
(ii) has a less complex design because it includes no heuristics for resolving incompatible shardings, which avoids bugs that lead to degraded performance (\Cref{sec:discussion}),
(iii) provides detailed user feedback in each step, which GSPMD does not, as we discuss next.


\begin{table}
\centering
{\small
\begin{tabular}{ll|cc|cc|}
\cline{3-6}
 & & \multicolumn{2}{c|}{\textbf{MFU (\%)}} 
   & \multicolumn{2}{c|}{\textbf{HBM (GB)}} 
 \\ \cline{1-6} 
 \textbf{Mesh} & \textbf{Size} & \textbf{\partir{}} & \textbf{GSPMD}
   & \textbf{\partir{}} & \textbf{GSPMD}
 \\ \hline
 16x2 TPU & \textbf{5B} & \textbf{58.5} & 58.3 & 14.38 & 14.38 \\ 
 32x4 TPU & \textbf{32B} & \textbf{52.3} & 52.2 & 14.48 & 14.48 \\ 
 8x2 GPU  &  \textbf{5B} & 42.2 & \textbf{42.9} & 27.02 & 26.73  \\
 \hline 
 
\end{tabular}}
\caption{
\label{table:mfu}\small 
The MFU (higher is better) and HBM usage (lower is better) on GPUs and TPUs using \partir{} and GSPMD.
}
\end{table}

\subsection{Composability and predictability}\label{eval:debug}


We demonstrate that \partir{} users compose strategies using manual tactics and verify the model achieves the expected communication collectives from the respective papers.
\Cref{table:collectives_manual} shows the four different models partitioned with different schedules and reports the resulting number of collectives.

\begin{table}[!t]
\centering
{
\footnotesize
\begin{tabular}{l|l|rrrrrr}
\toprule
\textbf{Model} & \textbf{Schedule} & \textbf{AG} & \textbf{AR} & \textbf{RS} & \textbf{A2A} \\
\midrule
\multirow{7}{*}{T32} & BP &    0 &    290 &    0 &      0 \\
    & BP+MP &    0 &    418 &    0 &      0 \\
    & BP+MP+Z2 &  129 &    289 &  129 &      0 \\
    & BP+MP+Z3 &  259 &    289 &  129 &      0 \\
    & BP+MP+Z3+EMB &  515 &    354 &  257 &      0 \\
    & MP &    0 &    128 &    0 &      0 \\
    & EMB &  256 &    193 &  128 &      0 \\
\cline{1-6}
\multirow{4}{*}{IT32} & BP &    0 &      0 &    0 &      0 \\
    & BP+MP &    0 &  98304 &    0 &      0 \\
    & BP+MP+MQ &   64 &  98304 &    0 &  98240 \\
    & MP &    0 &  98304 &    0 &      0 \\
\cline{1-6}
\multirow{3}{*}{UNet} & BP &    0 &    503 &    0 &      0 \\
    & BP+Z2 &  517 &      2 &  501 &      0 \\
    & BP+Z3 &  799 &      2 &  501 &      0 \\
\cline{1-6}
GNS & ES &    0 &    423 &    0 &      0 \\
\bottomrule
\end{tabular}
}
\caption{\label{table:collectives_manual}
\small
Collectives introduced in the MLIR by different schedules.
\textbf{AG}: AllGather, \textbf{AR}: AllReduce, \textbf{RS}: ReduceScatter, \textbf{A2A}: All2All.
}
\end{table}
T32 has 289 parameter tensors (9 for each block + embeddings). 
With batch parallelism, we expect one AllReduce (AR) for each parameter gradient tensor and one AR for the loss value. The resulting 290 ARs add up with the 4 ARs per layer that Megatron \cite{shoeybi2020megatronlm} requires when composing both strategies together.
Both Z2 and Z3 \cite{zero2019} partition parameter gradients and optimizer states along the batch axis (of embeddings and four-parameter tensors per layer in this experiment), resulting in 129 of the existing ARs becoming ReduceScatters (RS), and the introduction of one or two AllGather (AG) per parameter tensor in the case of Z2 or Z3, respectively. The embedding partition strategy (EMB) partitions the embedding tensor along the d\_model dimension, which has the
effect of partitioning activations.
The reasoning is similar for IT32 and UNet. Note that IT32 is an inference-only benchmark and does not require any AR for batch parallelism, the collectives in IT32 reflects the serving loop. Multi-Query sharding (MQ) \cite{inference_transformer} over the batch axis introduces an AR and two All2All (A2A) per layer, except for the final loop, which requires an extra AG.
Finally, GNS is made of nodes connected by edges; we partitioned it using \textit{Edge Sharding} (ES)~\cite{edge_sharding}, that partitions the GNS's edges to create edges subgraphs distributed to devices in the network while replicating the associated GNS's nodes.
Every message passing and propagation through the GNS introduces a collective to communicate updates from neighbors in the GNS's graph.
Thus, we expect 2 AR per messaging passing (24) through each MLP layer (5) of every node and one for the global feature aggregator, the molecular GNS has an additional 2 AR for the graph encoder and one final one at the decoder.

\paragraph{Takeaway.} The number of operations matches the analytically expected one that the designer of a partitioning strategy would expect to observe if partitioning was applied correctly. The simplest example is batch parallelism: we expect a number of AR that matches the parameters plus one for the loss function, because each device operates on an independent batch of data and the loss is additive.

\subsubsection{Partial or full automation in schedules}\label{eval:auto}
\begin{figure}[!t]
    \centering
    \includegraphics[width=0.4\textwidth]{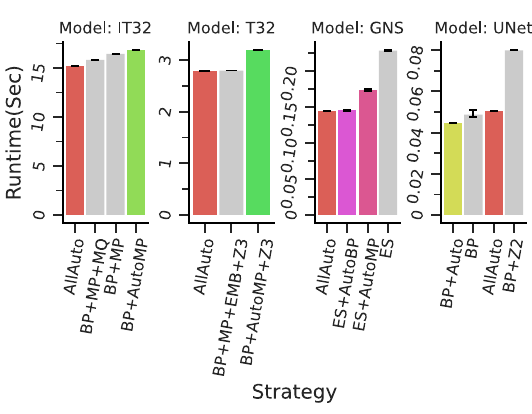}
    \caption{\small 
    One-step time (lower is better) on 8x4 TPUs. Grey-colored bars indicates schedule of manual tactics, while color-coded bars are schedules including automatic tactics.
    }
    \label{fig:auto_runtime}
\end{figure}

\partir{} users may not wish to partition their models manually, and not every model architecture has a well-studied set of partitioning strategies. 
Thus, users may explore different degrees of automation: fully manual, partially automatic, or fully automatic. 
\Cref{fig:auto_runtime} shows the actual runtime results of combining manual and automatic tactics and using them to partition the models for $32$ TPU devices. 
Using automatic partition can alleviate the burden of manually partitioning T32, where AllAuto results in a partition with comparable performance to a fully manual schedule. 
While combining manual with automatic partition gives us performance improvements for UNet and GNS, it can also come with a performance penalty: e.g., BP+Auto+Z3 in T32 results in slower runtimes than the fully manual schedule.

\subsection{Resolving conflicts with incrementality}\label{eval:conflict}
\begin{figure}[htbp]
\centering
\includegraphics[height=132pt]{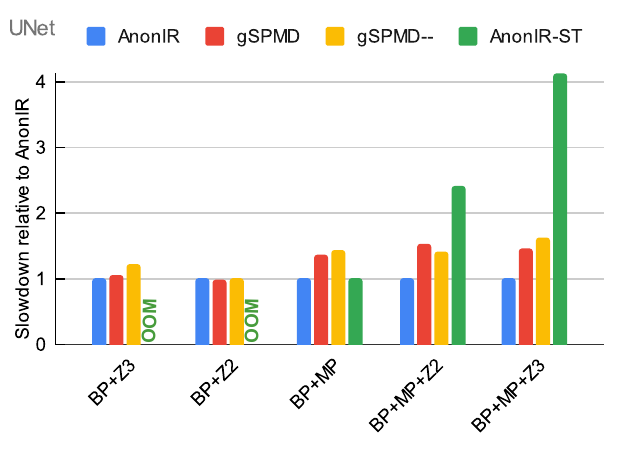}
\caption{\small Relative slowdown compared to \partir{} (higher is worse) for UNet partitioned on a
\inlc{\{8:batch, 2:model\}} TPU.
}
\label{fig:incremental}
\end{figure}
We show the importance of incrementality as a solution for compiler-internal conflicts in \Cref{fig:incremental}.
We compared \partir{} against \partir-st (\textbf{S}ingle \textbf{T}actic), which amalgamates all tactics of a schedule into a single tactic (i.e., no propagation in between tactics); 
GSPMD, with expert-defined sharding constraints baked into the model code to resolve conflicts; 
and GSPMD\--\--, which does not use internal sharding constraints for conflict resolution. 
We evaluated this on UNet with BP and Z2/Z3 -- two partitioning strategies that cause conflicts discussed in \Cref{sec:conflicts}. 
Furthermore, we show the addition of a Megatron-like~\cite{megatron2019} MP tactic to UNet along the model axis.
We performed a similar experiment with the transformer benchmarks used in~\Cref{eval:sota}, and (surprisingly) obtained comparable results of GSPMD and GSPMD\--\--, potentially because GSPMD heuristics to resolve conflicts have been highly tuned for Transformer models.

\partir{} achieves faster runtime compared to the baselines; 
(2) incrementality is fundamental for our design, \partir-st generated programs that exceeded the device memory limit without a way to resolve the partitioning conflicts; (3) in multi-axes settings, even when we do not expect conflicts (e.g., {\em BP+MP}), the lack of incrementality causes a performance regression in the GSPMD case; (4) in comparison, without sharding constraints, GSPMD\--\-- produces programs that fit but are noticeably slower  to \partir{}; and (5) Through a trial-and-error process we found optimal sharding constraints for GSPMD that matches \partir{} performance: UNet required 5 sharding constraints per layer (carefully placed after the down sampling layers); Transformer BP+Zero sharding required 2 sharding constraints per layer; and for GNS we could not even figure out where exactly to put sharding constraints to achieve Edge Sharding with reasonable effort. No internal such annotations were needed with \partir. 

\subsection{\partir partition time evaluation}\label{eval:overhead}
\begin{figure}[htbp]
\centering
    \includegraphics[width=0.39\textwidth]{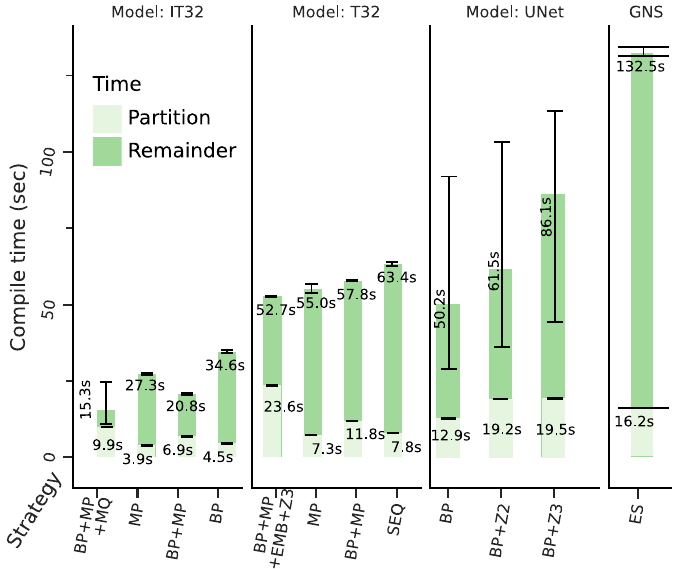}
    \caption{\small
    \partir partitioning vs. overall compilation time.
    }
    \label{fig:compile_time}
\end{figure}
\Cref{fig:compile_time} shows that \partir partitioning time is a small percentage (max of 14\%) of the overall XLAs' compilation time, which is important for an interactive user workflow. While non-negligible, 
these models usually train for several days or even months~\cite{gemini, chinchilla2022,gpt3_2020}; hence, even longer partitioning times compared to overall compilation is acceptable.


\section{Discussion and limitations}\label{sec:discussion}

\paragraph{Program rewriting versus annotations}
GSPMD~\cite{gspmd2021} follows the traditional view in HPC of distribution as a data layout problem. It separates the propagation of sharding annotations from code generation to deal with inserting and optimizing collectives; which we view as a brittle design, as their logic may go out of sync, leading to bugs that significantly degrade performance and go unnoticed. 
We discovered a bug where a model lowered using GSPMD partitioner had 3x slower training step-time compared to lowering it with \partir{}. This a consequence of  code generation relying on heuristics based on operand and result shardings, rather than on program rewrites during propagation; their partitioner introduced AllGathers on a matrix multiply when sharded on multiple axes on the same dimensions. To resolve this, the pull request \#13875~\cite{gspmdbug} adds another pattern to the partitioner. 
By contrast, the \partir{} rewrite system design is robust by compartmentalized dialects. 
It rewrites the program incrementally relying on pattern matching on the IR and introducing \partir:Core loop structures. 
These loops reflect the semantics and mesh axes in the IR allowing for temporal interpretation and testing without actual partitioning.
\partir:HLO consumes these loops to introduce the SPMD collectives, without specialized per-op code.

\paragraph{Reshape support}
Reshape ops pose a challenge for propagation: Consider a reshape from \inlc{tensor<16>} to \inlc{tensor<4x4>}. 
Assume that we are given a 1D mesh of 8 devices. This is too large to return chunks of the \inlc{tensor<4x4>}, and propagation will get blocked. Intuitively the solution is to logically reshape mesh \inlc{\{"A":8\}} to \inlc{\{"A1":4, "A2":2\}}, 
in which case propagation over the "A1'' axis will shard the output on dimension 0. \partir{} does not consider mesh transformations like that in the middle of propagation. By contrast, GSPMD~\cite{gspmd2021} manipulates the mapping of logical IDs to data, as it defines the sharding directly in terms of logical device IDs and that enables propagation through reshapes.
However, addressing logical IDs requires the propagation system and partitioner to pattern-match on these low-level representations, and
results in IR blow-up proportional to the number of devices.

\paragraph{Padding and spatial partitioning}
Value tiling and propagation via loops assume that the
number of devices in an axis must exactly divide a
partitioned dimension; otherwise, propagation gets blocked.
In addition, partitioning convolutional NNs on {\em spatial} dimensions requires communicating with neighboring devices~\cite{googleSpatial, spatialMedical}. 
\partir has limited support for these features due to lack of demand.

\paragraph{Explicitly replicating values} 
\partir propagation acts greedily (\Cref{ssec:propagation}) and may thus partition tensors the user wants to replicate.
For example, by partitioning the optimizer state, propagation also shards the parameters, but for Z2 the parameters must be replicated (\Cref{background:parallelism}).
Hence, \partir{} exposes an explicit \inlc{atomic<value, axis>} action, which creates a trivial loop whose sole purpose is to block propagation and keep a value replicated. 
For example \inlc{atomic<\%x, "M">} replaces \inlc{\%x} with this loop:
\begin{lstlisting}[language=mlir]
%xr = loop "M" [any] (%r:range<...>) { yield %x; }
\end{lstlisting}
The \inlc{[any]} ensures all devices compute the same value. 


\paragraph{Model-internal annotations.}\label{sec:tags}
\Cref{eval:conflict} argues that sequentialization of decisions lets us handle most conflicts. 
However, we are still left with corner cases. For example,
consider a component of matrix diagonalization, where a matrix is multiplied by its transpose:
\begin{lstlisting}[language=mlir]
func @main(%x: tensor<(256x256xf32>) {
  %tx = transpose %x {dims=[1,0]} : ...
  %y = matmul(%x, %tx) ...
\end{lstlisting}
If we shard \inlc{\%x} on dimension 0 along a given mesh axis "M", then \inlc{\%tx} (its transpose) is sharded on dimension 1. That is a propagation ``conflict'' and prevents sharding of the \texttt{matmul}.
To resolve this conflict, users must replicate the intermediate tensor \inlc{\%tx} before partitioning it, which is done by naming the tensor using a primitive called {\em tag}, then applying an atomic action on it:
\begin{lstlisting}[language=mlir]
func @main(%x: tensor<(256x256xf32>) {
  %tx = transpose %x {dims=[1,0]}
  %tag_tx = tag "transposed" %tx
  %atomic = loop "M" [any] (%r:range<...>) { 
    yield %tag_tx; 
  }
  %y = matmul(%x, %atomic) ...
\end{lstlisting}
By this forcing of replication of the intermediate \inlc{\%tx}/\inlc{\%tag_x} an \inlc{all\_gather} is inserted on the second operand of the \texttt{matmul} in the final partitioned function:
\begin{lstlisting}[language=mlir]
func @main(%x: tensor<(16x256xf32>) {
  %tx = transpose {dims=[1,0]} %x : tensor<256x16xf32>
  %gx = all_gather [{},{"M"}] %tx : tensor<256x256xf32>
  %y = matmul(%x, %gx) : tensor<16x256xf32>
  ...
\end{lstlisting}%





\section{Related work}\label{sec:actual-related-work}
The need for NN scaling motivated new partitioning tools~\cite{shazeer2018mesh,deepspeed_2021}, as frameworks like TensorFlow~\cite{tensorflow_osdi_2016} and PyTorch~\cite{pytorch2019} support only data and limited model parallelism.
JAX~\cite{jax2018github} provides two main partitioning APIs: \texttt{jax.jit}, which surfaces GSPMD~\cite{gspmd2021} (that builds on GShard~\cite{gshard}); and \texttt{shard\_map}, which disables GSPMD and lets users manually perform communication across devices.
\partir{} builds on top of some of JAX's partitioning infrastructure (e.g., meshes), but instead of users annotating their NN code with sharding annotations, or SPMD collectives in the case of \texttt{shard\_map}, our users define their partitioning strategies using schedules, which resemble ideas found in kernel-generating schedule-based systems~\cite{Taco2017,distal:pldi,Halide:PLDI:2013,tvm2018,Fireiron:PACT:2020}.
Although we took inspiration from Halide schedules~\cite{Halide:PLDI:2013}, the schedules in \partir{} differ in that they are applied to entire programs that span multiple devices rather than generating small program kernels for a single device.
Similar to Lift~\cite{Lift:ICFP:2015} and RISE~\cite{Elevate:ICFP:2020}, \partir{} is inspired by functional-style IRs~\cite{Lift:IR:CGO:2017} that transform programs by equality-based rewriting \cite{Visser:1998} and similar to the functional array representation in \cite{Dex}.
This is in contrast to DaCe \cite{dace} that applies graph transformations interactively by employing a representation based on data flow graphs.
DistIR~\cite{distir2021} exposes a Python API with explicit device placement and point-to-point communication to program the distribution directly. \partir:Core abstracts away distribution and deals with it during SPMD lowering.
Finally, \partir{} allows for an automatic partitioner to be used as tactic, making it orthogonal to Alpa \cite{alpa2022}, AutoMap \cite{automap, automap-reloaded} and others.

\paragraph{A newer generation of StableHLO partitioners} Addressing some \partir{} limitations (e.g. limited reshape support, relying on examining the loop structure for propagation), the newly introduced {\em Shardy} propagation system~\cite{shardy} employs rewriting-free sharding annotation propagation (like GSPMD), but based on mesh axes (like \partir{}). It extends the \partir{} TMR idea (\Cref{ssec:equivalences}) with sharding {\em factors} that allow for implicit mesh splitting. Currently, Shardy relies on a separate code generation/collectives insertion pass, the same one GSPMD uses. 
Which complicates the process of ensuring consistency between the two passes and to obtain a cost model at the MLIR level prior to entering the XLA compiler. 
However, there exists ongoing work to address these issues. 
Shardy supports incremental propagation based on {\em priority} annotations in sharding specifications, and features an elaborate hierarchy for conflict resolution. 
Finally the Mesh MLIR dialect\cite{mesh_dialect} is an attempt to introduce mesh-based operations at the StableHLO MLIR level that can be used to define a complete partitioner.


\section{Conclusion and Future Work}
\partir{} is an MLIR-based compiler that enables effortless partitioning of tensor programs by decoupling partitioning strategy from model code and instead expresses them using tactics. Each tactic desugars into a series of compiler rewrite actions.
\partir{}'s incremental design enables a powerful (yet simple) propagation system that automatically predictably partitions programs without relying on cost-based heuristics to handle conflicts.
In the future, we want to support training over heterogeneous device clusters using MPMD~\cite{gpipe, pipedream, pipedream_2bw}, which will require new partitioning tactics and IR extensions.



\clearpage

\bibliographystyle{plain}
\bibliography{main}

\newpage
\appendix 
\onecolumn 
\section{Experiments extended}\label{app:sec:benchmarks}
\subsection{MFU}
We evaluated the MFU to compare SOTA performance in \Cref{eval:sota}, here we document what is MFU.
MFU is defined as a ratio of actual FLOPs per second and theoretical FLOPs per second: 
\begin{equation*}\label{eqn:mfu}
100\times \frac{\text{model FLOPs} / \text{step time (s)}}{\#\text{devices} \times \text{peak FLOPs per second}}
\end{equation*}
An MFU of 100\% indicates that the model is utilizing the full computational capacity of the hardware.

\subsection{Composing automatic and manual tactic}
\Cref{table:extra_cost_auto} provides the the full metrics of using AutomaticPartition to find sharding strategies.

\begin{table}[ht]
\centering
\begin{tabular}{l|l|rrrrrr}
\toprule
     &            &       Mem &  Est.Runtime(ms) &   AG &     AR &   RS &    A2A \\
Model & Strategy &           &                  &      &        &      &        \\
\midrule
\multirow{4}{*}{\rotatebox[origin=c]{90}{\textbf{GNS}}} & ES &  10379.47 &           294.13 &    0 &    423 &    0 &      0 \\
     & ES+AutoMP &   8424.38 &           146.43 &  220 &    752 &   97 &      0 \\
     & ES+AutoBP &   8141.38 &           101.47 &   72 &    679 &   72 &      0 \\
     & AllAuto &   2508.92 &           118.12 &  476 &    854 &  272 &      0 \\
\cline{1-8}
\multirow{4}{*}{\rotatebox[origin=c]{90}{\textbf{IT32}}} & BP &  18302.16 &          1139.31 &    0 &      0 &    0 &      0 \\
     & BP+MP &   5607.73 &          1447.83 &    0 &  98304 &    0 &      0 \\
     & BP+MP+MQ &   5439.73 &          1498.92 &   64 &  98304 &    0 &  98240 \\
     & MP &   5151.44 &          4327.35 &    0 &  98304 &    0 &      0 \\
\cline{1-8}
\multirow{8}{*}{\rotatebox[origin=c]{90}{\textbf{T32}}} & BP & 100343.69 &          4803.34 &    0 &    290 &    0 &      0 \\
     & BP+AutoMP+Z3 &  40472.80 &          4902.41 &  547 &    161 &  353 &      0 \\
     & BP+MP &  59826.45 &          4856.25 &    0 &    418 &    0 &      0 \\
     & BP+MP+Z2 &  50124.45 &          4856.25 &  129 &    289 &  129 &      0 \\
     & BP+MP+Z3 &  45068.63 &          4960.32 &  259 &    289 &  129 &      0 \\
     & BP+MP+Z3+EMB &  47541.60 &          4946.35 &  515 &    354 &  257 &      0 \\
     & MP & 177148.23 &         10837.42 &    0 &    128 &    0 &      0 \\
     & EMB & 176974.51 &         10934.86 &  256 &    193 &  128 &      0 \\
\cline{1-8}
\multirow{5}{*}{\rotatebox[origin=c]{90}{\textbf{UNet}}} & BP &   2406.68 &            25.80 &    0 &    503 &    0 &      0 \\
     & BP+AutoMP &   1693.65 &            20.51 &  162 &    482 &   68 &      0 \\
     & BP+Z2 &    933.36 &            25.80 &  517 &      2 &  501 &      0 \\
     & BP+Z3 &    309.48 &            37.73 &  799 &      2 &  501 &      0 \\
     & AllAuto &   1126.94 &            15.74 &  415 &    565 &   88 &      2 \\
\bottomrule
\end{tabular}
\caption{
\label{table:extra_cost_auto}
Collective and simulator cost estimation of various tactics}
\end{table}

\subsection{Simulation results}\label{appendix:simulation}
Here, we show how close our simulator can approximate real performance on the TPUv3 32 device described in \Cref{eval:auto}.
While the details of the simulator we leave to another paper, it is indeed a very simple simulator due to most of the heavy lifting done by our \partir:HLO dialect.
\partir:HLO generates device-local communication ops that reference the mesh and device, and \partir keeps a registry of popular compilation devices (and it is easy to extend, requiring only high-level device specs). 
Combining both the \partir:HLO code, which has tensor shapes, and communication ops on each device, with the target specs, our simulator iterates over each SPMD context, tracks the live memory, and counts flops usage for the communication ops also tracks the byte transfers. Using the target device specs, it would estimate its statistics. 
This simple analytical cost model worked well for our AutomaticPartition tactic and user debugging. While the absolute values are not guaranteed to be correct, the relative improvements should still be sound.
For example, applying a {\em BP} tactic, you would expect the memory and flops usage to be reduced by a factor of the number of devices in the mesh if the BP was applied correctly. 
A search algorithm that will seek to improve the relative partitioning of the model will also benefit from this simple analytical function.

Hardware measurement follows the methodology described in \Cref{eval:setup}.
All experiments were run on a 32 TPUv3 as described in the evaluation section \Cref{eval:auto}.

\subsubsection{Runtime estimation}
\begin{figure}[htbp]
    \centering
\includegraphics[scale=0.75]{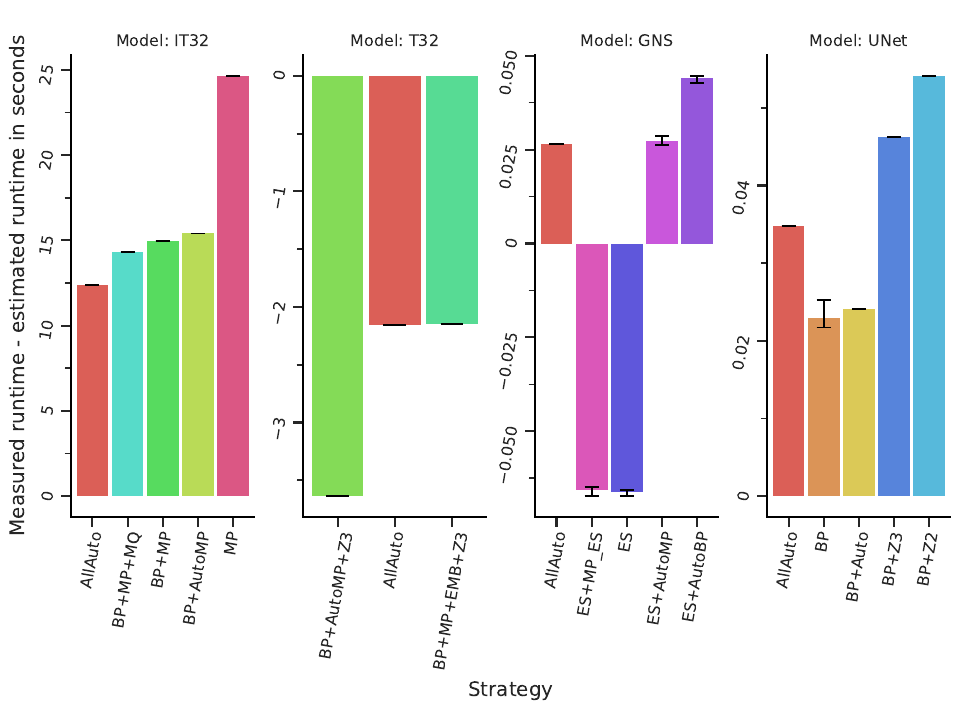}
    \caption{\small
    \partir's simulator runtime estimation compared to the actual measured runtime, in seconds, closer to the zero better.
    }
    \label{fig:simulation_vs_measured}
\end{figure}

First, we look at the difference between the estimated and actual runtime for the various strategies described in the paper.
\Cref{fig:simulation_vs_measured} shows the difference between the estimated and actual measured seconds on hardware. 
First, we notice that the error is within acceptable range (sub milliseconds) for specific configurations, especially for GNS and UNET. 
The simulator underestimated T32's most complicated strategy by 3 seconds at the worst case; these often can be attributed to the layout pass of XLA and XLA optimization.
IT32, on the other hand, the simulator overestimated its runtime (meaning we thought it would be slower than it is); this is due to the key-value caching optimization we implemented in the model; without caching, the numbers are much closer to the T32 error range.
Overall, both automatic tools and human users use the estimated runtime as a proxy for relative improvement of applying the sharding strategy rather than absolute value estimates.

Absolute value estimates require a learned cost model that learns what the backend compiler optimizer will do, different network topology profiles, and various specialized ops performances (e.g., caching). 
Improving the simulator is beyond the scope of our work, but \partir{} allows the end to plug in any simulator, and we are currently looking into a learned cost model for this exact reason.

\subsubsection{Memory estimation}
\begin{figure}[htbp]
\centering
    \includegraphics[scale=0.75]{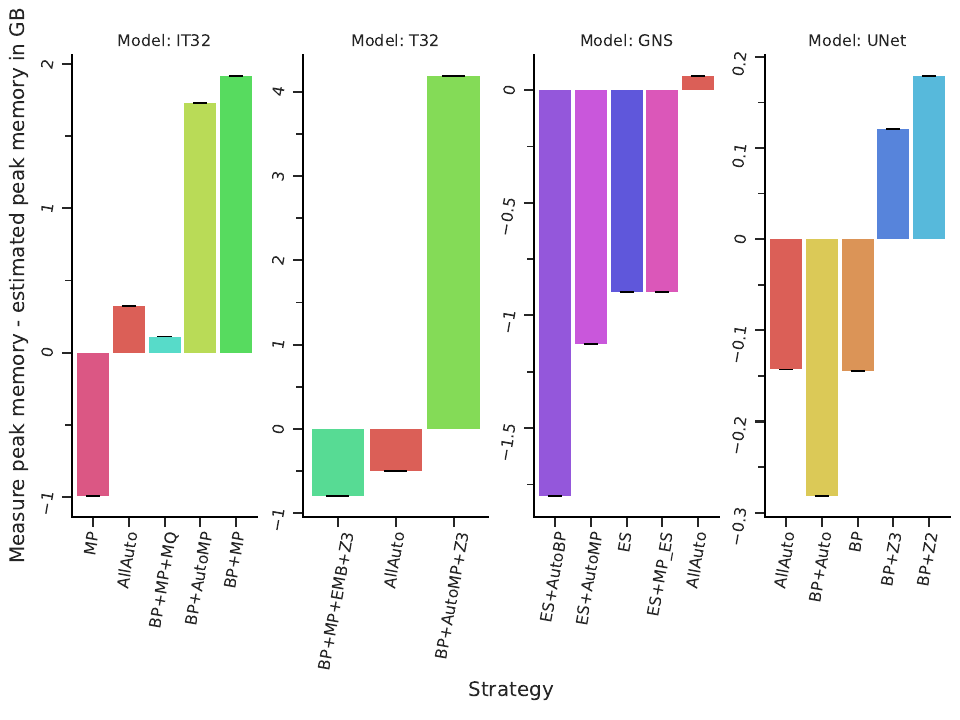}
    \caption{\small
    \partir's simulator estimates of memory compared to the actual measured memory, in GB, closer to the zero better.
    }
    \label{fig:sim_mem_vs_real}
\end{figure}

Related to runtime estimation is memory estimation; this is more important for automatic tools to learn if any proposed solution does not fit in the given device specification. 
We implement a live range analysis of a tensor usage in a given SPMD context at the \partir:HLO level, where we follow a tensor as long as it is being used; we also implement a simple fusion heuristic that will predict what the backend compiler will do to the tensor.
The results are shown in \Cref{fig:sim_mem_vs_real}.
Here, we can see the results are much closer to 0 (i.e., no error); often, we prefer to over-estimate the memory usage to discourage solutions close to the boundaries.
XLA and other backend compilers have specialized optimization when a model is on the boundaries - such as remating parts of the computation or reducing the fusion of ops. The simulator forces the automatic partitioners to avoid these regions and focus on models that fit comfortably, not to kick off aggressive memory optimization from the backend compiler. 

\subsubsection{Partitioning time}

\begin{figure}[htbp]
\centering
\includegraphics[scale=0.75]{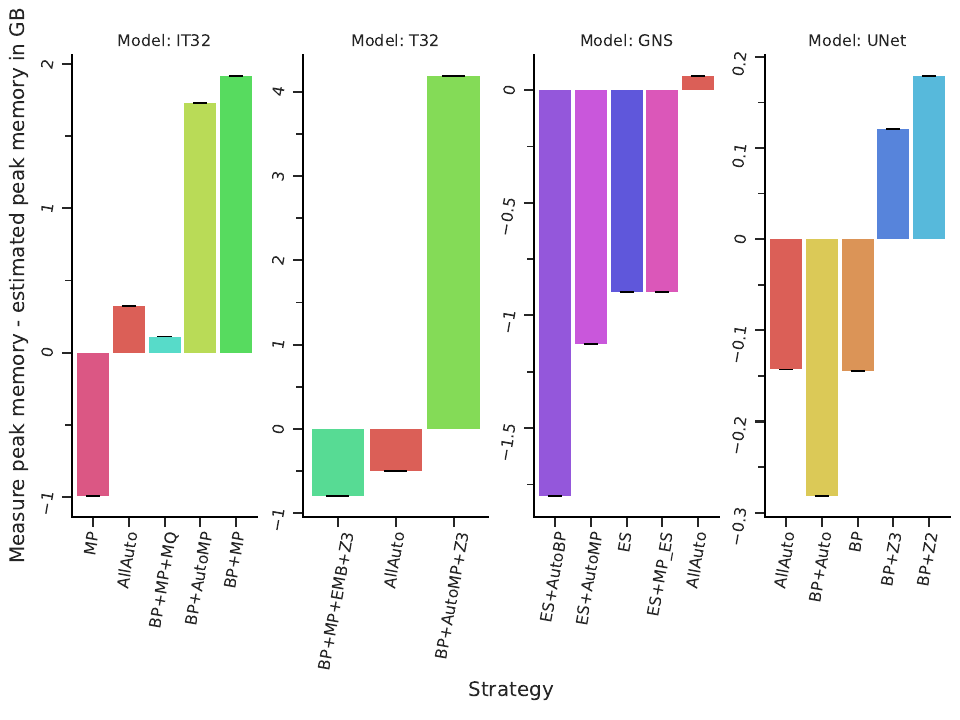}
    \caption{\small
    \partir's automatic partitioning tactic search time compared to manual partitioning. 
    }
    \label{fig:automatic_part_time}
\end{figure}

We showed the partitioning time of manual partitioning tactics in \Cref{eval:overhead}; here, we show the time it takes to run AutomaticPartition.
AutomaticPartition tactic depends on the algorithm implementation used for the search; we implemented an algorithm similar to the one described in~\cite{automap, automap-reloaded}.
As \partir{} is agnostic to the optimization algorithm, we are looking at different algorithms to implement, such as ILP solvers or RL-based solutions. 
Different algorithms have different search profiles, and naturally, by increasing the number of axes (and, as a result, the number of available decisions to make), \partir{} automatic partitioning time does increase, as can be seen in \Cref{fig:automatic_part_time}.
Even in the worst-case scenario, we see a search time of 1250 seconds, which might be a lot compared to the manual tactic (hence we allow composing manual and automatic tactics and benefit from both). However, we want to argue that: 1) the search time reduces the cognitive workload on the ML practitioner and saves them time from writing their sharding strategies and testing them; 2) these models train for weeks on end, a search time of 20 minutes is within the amortized acceptable range, especially when the shardings found can outperform known strategies; 3) we are actively developing the search mechanism and exploring different techniques, we expect this to only get faster with time.

\subsection{Schedules}\label{appendix:schedules}
Here, we provide the tactics we used for all the experiments; we note that the schedule is simply a list of these tactics as they appear in the order. For example, to replicate the schedule {\em BP+MP}, define the tactic as bellow, then pass to \partir\inlc{.jit(fun, schedule=[bp, mp])}. 

\paragraph{Batch parallelism and zero}
Regardless of the module, both tactics can easily be expressed as:
\begin{itemize}
    \item BP: shard the 0th dimension of the inputs, assuming the NN's update function takes X where its 0th dimension is the batch dimension:
\begin{lstlisting}[language=python, escapeinside={*@}{@*}]
BP = ManualPartition(inputs={"X": 0}, axis="batch")
\end{lstlisting}
    \item Z3: shard the 0th dimension of the gradients, and optimizer state, but replicate the parameters:
\begin{lstlisting}[language=python, escapeinside={*@}{@*}]
Z2 = ManualPartition(inputs={'params': REPLICATED,'opt_state': FIRST_DIVISIBLE_DIM}, axis="batch")
\end{lstlisting}
    \item Z3: shard the 0th dimension of the parameters, gradients, and optimizer state,
\begin{lstlisting}[language=python, escapeinside={*@}{@*}]
Z3 = ManualPartition(inputs={'params': FIRST_DIVISIBLE_DIM, 'opt_state': FIRST_DIVISIBLE_DIM}, axis="batch")
\end{lstlisting}

\end{itemize}

Our tactics take a callback that applies a sharding based on the parameter name in the NN, this is useful for Megatron and model parallelism tactics.

\paragraph{T32/T48/IT32 tactics}
There is no difference in tactics applied to any of the transformer models, showing the usefulness of separating the model code from the sharding strategies. 

MP: here we apply the Megatron~\cite{megatron2019} sharding strategy. 
\begin{lstlisting}[language=python]
    def _model_sharding(param_name): 
      if not 'w': # only shard the weights
        return UNKNOWN # Let the infer-propagate decides. 
      if multi_head_attention_regex.contains(param_name): 
        return 0 # shard the linear weights on 0th dimension
      if 'dense_up' in param_name:
        # dense layer on its cols
        return 2
      if 'qkv_einsum' in param_name:
        # the queries on first dimension
        return 1

MP = ManualPartition(inputs={'params': apply(_model_sharding)}, axis="model")
\end{lstlisting}

\paragraph{UNet tactics}
\begin{itemize}
    \item MP: here we try to mimic the Megatron sharding strategy. 
    \begin{lstlisting}[language=python]
    def _model_sharding(param_name): 
      if not 'w': # only shard the weights
        return UNKNOWN # Let the infer-propagate decides. 
      if multi_head_attention_regex.contains(param_name): 
        return 0 # shard the linear weights on 0th dimension
      if 'conv_residual_block' in param_name:
        # shard the convolutions on their weights not stride. 
        return 3 if is_2d_conv(param_name) else 2. 
MP = ManualPartition(inputs={'params': apply(_model_sharding)}, axis="model")
\end{lstlisting}
    
\end{itemize}
\paragraph{GNS tactics}

\begin{itemize}
    \item ES: Simply performs search on the batch axis.
\begin{lstlisting}[language=python]
# Shard both the sender and receivers of GNS that is using jraph library.
# predictions is a custom name inside the GNS implementation.
ES = ManualPartition(inputs={'edges': {"predictions": 0, "predictions_targets": 0}}, axis="batch")
\end{lstlisting}
\end{itemize}
\paragraph{Automatic tactics}
These tactics are the simplest. 

\begin{itemize}
    \item AutoBP: Simply performs search on the batch axis.
\begin{lstlisting}[language=python]
AutoBP = AutomaticPartition(axes=["batch"])
\end{lstlisting}
    \item AutoMP: Simply performs search on the model axis.
\begin{lstlisting}[language=python]
AutoMP = AutomaticPartition(axes=["model"])
\end{lstlisting}
\item AutoAll: Perform search on all axes:
\begin{lstlisting}[language=python]
AutoAll = AutomaticPartition(axes=["batch", "model"])
\end{lstlisting}
\end{itemize}

\section{\partir:Core extended}\label{appendix:core}
Here we discuss extended details around \partir:Core omitted from \Cref{sec:core}.
This section mostly focuses on the multi-axis case and how propagation and the loop construct handle nesting for multiple axes (in a 2D or higher-dimensional mesh).

\subsection{Multi-axis propagation and deep-tiling}\label{ssec:multi-axis}
\subsubsection{Multi-axis analysis for propagation}
\label{sec:multi-axis-propagation}
Propagation becomes involved in the presence of multiple axes.
Checking whether some operand is overly tiled or a result is overly sliced requires a deeper look into the definitions of
operands or uses of results. For example, consider the following: 
\begin{lstlisting}[
    language=mlir]
%x = loop "a" [#sum] (%ra: range<4>) {
   %t = loop "b" [#tile<0>] (%rb: range<2>) { ... }
   yield %t
}
%z = matmul(%x, %y)
\end{lstlisting}%
To realize that the left \opmatmul operand is tiled, we
have to look internally under the reduction loop.

The situation is even more complex when both tiling loops and slices are involved:
\begin{lstlisting}[
    language=mlir]
%x = loop "a" [#tile<0>] (%ra: range<4>) {
  %t = loop "b" [#tile<0>] (%rb: range<2>) { ... }
  yield %t
}

%w = loop "a" [#tile<0>] (%ra: range<4>) {
    %sx = slice 0 %x[%ra]
    %z = matmul(%sx, %y)
    ... 
}
\end{lstlisting}%
In this situation the left \opmatmul operand is not even the result of a 
\coreloop op, it is just a \inlc{slice}!
However, if we keep looking {\em backward}, we discover that the argument \inlc{\%x} of the \inlc{slice} instruction is produced by a tiling loop over \inlc{"a"} that can cancel out the \inlc{slice},
and internally there is {\em yet} another tiling loop, i.e. over \inlc{"b"}.
Consequently, we could rewrite the \opmatmul as a loop over \inlc{"b"}.

What the examples highlight is that when multiple axes are involved, it is necessary to traverse nests of \coreloop operations as well as chains of \coreslice operations to avoid missing relevant matches. A dual situation applies to \coreslice operations in backward propagation, where we need to apply a {\em forward} analysis starting at the uses of the result of an operation.
To deduce whether there is a match on the TMR entry one would have to look {\em inside} the
loop over axis \inlc{"a"} to determine that dimension 0 is indeed tiled across axis 
\inlc{"b"}. 
A dual situation arises with overly sliced results, where one may need to look
inside {\em chains} of slice operations. For this reason \partir employs a simple static analysis
to determine the per-dimension excess tiling (or slicing) for the operands (or results) of an
operation. 

\subsubsection{Deep tiling}
A big design goal of \partir is to support incremental partitioning (sometimes called ``recursive partitioning''~\cite{gspmd2021}) that never undoes previous actions.
Consider a value 
\inlc{\%x} that is already sliced (e.g. due to previous value tiling or propagation)
along axis \inlc{"a"}:
\begin{lstlisting}[
    language=mlir]
%xt = loop "a" [#tile<1>] (%ra: range<4>) { 
  %xs = slice 1 %x[%ra]) ; 
  ... 
}
\end{lstlisting}
Imagine a user or a tool needing to {\em further} tile the value \inlc{\%x} across axis \inlc{"b"} and dimension $1$.
It would be wrong to perform a flat value tiling --
instead we need to gather up any existing sliced uses of \inlc{\%x} and apply what we call a ``deep'' tiling action.
In code: 

\begin{lstlisting}[language=mlir]
// WRONG: would undo previous actions
//   \%xtt = loop "b" [\#tile<1>] (\%rb: range<2>) \{ yield (\%slice 1 \%x[\%rb]) \}
// CORRECT: deep tiling over both previous and new axes!
%(*@\codehl{yellow}{xtt}@*) = loop "a" [#tile<1>] (%ra: range<4>) { 
  %t = loop "b" [#tile<1>] (%rb: range<2>) {
    %xsa = slice 1 %x[%ra]
    %xsb = slice 1 %xsa[%rb]
    yield %xsb
  }
  yield %t
}    
%xt = loop "a" [#tile<1>] (%ra: range<4>) { 
  %tmp = slice 1 %(*@\codehl{yellow}{xtt}@*)[%ra]; 
  ... 
}
\end{lstlisting}
Using the multi-axis analysis from \Cref{sec:multi-axis-propagation}, \partir is able to deduce the full tiling context that needs to 
be inserted during value tiling, including previous slicing uses and the new tiling requested. This
applies to both user-initiated value tilings as well as inference-initiated value tilings.

\section{\partir type system and correctness of translation from \partir:Core to \partir:SPMD}\label{app:sec:translation-correctness}
In the paper, for the sake of being concise, we presented lowering from \partir:Core to \partir:HLO as a direct translation.
In reality our system actually lowers to  \partir:HLO via an intermediate dialect, \partir:SPMD,
which makes per-device computations and cross-device communication explicit.
Lowering via \partir:SPMD comes with three advantages:
(i) It structures the implementation of our system;
(ii) it eases testing of our compiler pipeline,
(iii) it facilitates the correctness proof of our lowering pipeline that we present in this appendix.
Our correctness result appears in \Cref{thm:correctness-of-translation}.

Note that lowering from \partir:SPMD to \partir:HLO (\Cref{sec:mhlo}) consists mostly of simple fusion passes, and hence is trivially correct.
This appendix therefore focuses on the correctness of translating \partir:Core (\Cref{sec:core}) to \partir:SPMD.

\paragraph{Propagation correctness}
We do not formally proof correctness of \partir{}'s propagation system since this would require formalized semantics for each StableHLO op, which we do not have.
But note that the trusted code base for the propagation system is small:
The tile-mapping registry (which is based on "obvious" algebraic properties of operations) and the written-once (not per-op) propagation code make it easier to empirically test \partir{}'s propagation system than in systems that define per-op propagation rules.

\subsection{Formal definition of \partir:Core}

\Cref{fig:partir:core} formally defines the syntax of \partir:Core programs and shows the interesting typing rules.
\Cref{fig:interpretation:partir:core} assigns semantics to \partir:Core programs in a denotational style.

\begin{figure}[ht]\footnotesize
\[\begin{array}{lll}
    \begin{array}{l}
        n, d, m, k \in \text{Integer constants} \\
        a, b, c \in \text{Axis identifiers}
    \end{array} &
    \begin{array}{l}
        x, y, z \in \text{Program variables} \\
        ~
    \end{array} &
    \begin{array}{lcll}
        \tau & ::= & \tensor{\ol{n}}  & \text{Tensor types} \\
        M & ::= & \{ a_1{:}n_1, \ldots, a_k{:}n_k \} & \text{Meshes}
    \end{array}
\end{array}\]
\\
\[\begin{array}{ll}
    \begin{array}{lcll}
        \multicolumn{3}{l}{\textbf{Value definitions}} \\
        v &  ::= & op(\xs)                    & \text{Tensor operations} \\
          & \mid & \ploop{a}{\sigma}{e}       & \text{Loop constructs} \\
          & \mid & \slice{d}{x}{r_{a}}        & \text{Slicing} \\
    \end{array} &
    \begin{array}{lcll}
        \multicolumn{3}{l}{\textbf{Loop actions}} \\
        \sigma & ::= & \tileaction{a}{d} \mid \sumaction{a} \\
        \multicolumn{3}{l}{\textbf{Expressions}} \\
        e & ::= & \plet x{:}\tau = v \pin e \;\mid\; \yield{x} \\
    \end{array}
\end{array}\]
\\
\begin{mathpar}
  \Infer{TTile}
        { M;\Gamma,r_a \tcore e : \tensor{...,n_d,...} \quad
          r_a \notin \Gamma \quad
          a{:}n \in M }
        { M;\Gamma \tcore \ploop{a}{(\tileaction{a}{d})}{e} : \tensor{...,n_d \cdot n,...} } \\ 
  \Infer{TSum}
        { M;\Gamma,r_a \tcore e : \tau \quad
          r_a \notin \Gamma \quad
          a{:}\_ \in M }
        { M;\Gamma \tcore \ploop{a}{(\sumaction{a})}{e} : \tau} \quad
  \Infer{TSlice}
        { x{:}\tensor{...,n_d\cdot n,...} \in \Gamma \quad
          r_a \in \Gamma \quad
          a{:}n \in M }
        { M;\Gamma \tcore \slice{d}{x}{r_a} : \tensor{...,n_d,...}}
\end{mathpar}
\caption{\partir:Core: abstract syntax and typing rules.}
\label{fig:partir:core}
\end{figure}

\begin{figure}\footnotesize
\begin{align*}
&
\begin{array}{llcl}
\multicolumn{4}{l}{\textbf{Interpretation of \partir:Core contexts}} \\
\llbracket \cdot \rrbracket := \cdot &
    \text{empty context} \\[2pt]
\llbracket \Gamma, x{:}\tau \rrbracket :=
    \llbracket \Gamma \rrbracket \times \llbracket \tau \rrbracket &
    \Gamma \text{ extended with tensor variable} & &
    \llbracket \tensor{\overline{n}} \rrbracket := \text{``space of arrays of shape $\overline{n}$''}
    \\[2pt]
\llbracket \Gamma, r_a \rrbracket :=
    \llbracket \Gamma \rrbracket \times \llbracket a \rrbracket &
    \Gamma \text{ extended with range variable} & &
    \llbracket a \rrbracket := \{0,\ldots,n-1\} \text{ for } a{:}n\in M
\end{array}
\\[4pt]
&
\begin{array}{l}
\textbf{Environment typing} \\
\gamma \in \llbracket \Gamma \rrbracket :\Leftrightarrow 
        dom(\gamma) = dom(\Gamma) \,\wedge
        \forall_{x\in dom(\gamma)}. \,\gamma(x) \in \llbracket\Gamma(x)\rrbracket
\end{array}
\\[4pt]
&
\begin{array}{l}
\textbf{Interpretation of values and expressions}
\\
\llbracket \Gamma \tcore e : \tau \rrbracket : \llbracket \Gamma \rrbracket \to \llbracket\tau\rrbracket
\\[2pt]
\llbracket \Gamma \tcore \ploop{a}{(\tileaction{a}{d})}{e} : \tau \rrbracket \,\gamma :=
    \bigoplus_{i\in\llbracket a \rrbracket}^{axis=d} \llbracket \Gamma,r_a \tcore e : \tau' \rrbracket
    \,\gamma\cup\{r_a\mapsto i\}
\\[3pt]
\llbracket \Gamma \tcore \ploop{a}{(\sumaction{a})}{e} : \tau \rrbracket \,\gamma :=
    \sum_{i\in\llbracket a \rrbracket} \llbracket \Gamma,r_a \tcore e : \tau \,\rrbracket
    \,\gamma\cup\{r_a\mapsto i\}
\\[3pt]
\llbracket \Gamma \tcore \slice{d}{x}{r_a} : \tensor{...,n_d,...} \rrbracket \,\gamma :=
    \gamma(x)\![..., \gamma(r_a)n_d : (\gamma(r_a)+1)n_d, ...]
\\[2pt]
\llbracket \Gamma \tcore op(\xs) : \tau \rrbracket \,\gamma :=
    op\left(\overline{\gamma(x)}\right)
\\[2pt]
\llbracket \Gamma \tcore \yield{x} : \tau \rrbracket \,\gamma := \gamma(x)
\\[2pt]
\llbracket \Gamma \tcore \plet x{:}\tau = v \pin e : \tau' \rrbracket \,\gamma :=
    \llbracket \Gamma, x{:}\tau \tcore e : \tau' \rrbracket \,\gamma \cup
    \left\{ x \mapsto \llbracket \Gamma \tcore v : \tau \rrbracket\,\gamma \right\}
\end{array}
\end{align*}
\caption{%
    Interpretation of \partir:Core,
    assuming a fixed mesh $M$ which is not explicitly spelled out in the typing judgements
    (unlike in \Cref{fig:partir:core}).
}
\label{fig:interpretation:partir:core}
\end{figure}

Recall from \Cref{sec:core} that \partir:Core extends StableHLO.
\partir:Core therefore also extends the semantics of StableHLO, which is seen in \Cref{fig:interpretation:partir:core} in two places. 
The first place is in the interpretation of a tensor type $\tensor{\overline{n}}$.
This interpretation is simply the space in which StableHLO interprets arrays of shape $\overline{n}$.

To complete the interpretation of \partir:Core types, note that the type $a$ of a range variable is interpreted as the finite set of the first $n$ natural numbers, assuming $a{:}n$ is in the mesh $M$.

We conclude the presentation of the top pane in \Cref{fig:interpretation:partir:core} by pointing out that a \partir:Core typing context $\Gamma$ is interpreted as the cartesian product of the interpretations of the types that appear in $\Gamma$, which is fairly standard.

An environment $\gamma$ is a mapping from the names of tensor and range variables to the (disjoint union of all) spaces that interpret types.
We will only be concerned with environments $\gamma$ that satisfy the typing judgement $\gamma\in\llbracket\Gamma\rrbracket$
(which presents a mild abuse of the set membership symbol $\in$).

The interpretations of \plet and \coreyield in \Cref{fig:interpretation:partir:core} are standard:
\yield{x} looks up the mapping of $x$ in the given environment $\gamma$;
and a \plet expression interprets its subexpression $e$ in an environment that extends $\gamma$ with a mapping for the \plet\!\!-bound variable $x$.
The extended environment is written as
$ \gamma \cup \left\{ x \mapsto \llbracket \Gamma \tcore v : \tau \rrbracket\,\gamma \right\}$
at the bottom of \Cref{fig:interpretation:partir:core}.

The remaining interpretations in \Cref{fig:interpretation:partir:core} are for the syntactic category of \partir:Core values.
They may warrant a little more explanation,
and we now go through them from bottom up.

For interpreting a StableHLO operation $op$, \partir:Core defers to StableHLO's semantics.

For the interpretation of \coreslice, \Cref{fig:interpretation:partir:core} relies on the Python/NumPy syntax for slicing arrays.
Note that for the Python/NumPy-style array slicing to make sense here, it is crucial that the type $a$ of a range variable $r_a$ is interpreted as the set $\{0,\ldots,n-1\}$, assuming $a{:}n \in M$.

A \partir:Core \coreloop with a \coresum action is straightforwardly interpreted as summation over the subexpression $e$, which may contain free occurrences of $r_a$.
In the interpretation of $e$, the range variable $r_a$ is then mapped to the summation index $i$.

The interpretation of a \coreloop with a \coretile action is analogous.
The only difference is that instead of reducing the interpretation of subexpression $e$ with a summation operator $\Sigma$, we now take the {\em direct sum} (in the terminology of linear algebra).
Operationally, the direct sum boils down to concatenating the tensor arguments of $\bigoplus^{axis=d}$ along the $d$-th dimension, as indicated by the superscript $axis=d$.
Note that in NumPy one would express $\bigoplus_{i=0,\ldots,n-1}^{axis=d} e_i$
as \texttt{concatenate([$e_0$, \ldots, $e_{n-1}$], axis=d)}.

\subsection{\partir:SPMD}
\label{sec:spmd_extended}
\partir:Core includes parallel loops, but it leaves implicit the actual distribution of tensors across the mesh of devices.
For lowering to SPMD computations, we introduce the \partir:SPMD dialect that features:
\begin{itemize}
    \item distributed tensor types to specify how data is laid out across the mesh,
    \item an \spmdredist operation between distributed tensors that operationally reshards the data to match the destination distributed type (acting like a coercive type-cast), and 
    \item an \spmdexecute operation which contains device-local computation and consumes and produces distributed tensors.
\end{itemize}
Syntax and relevant typing rules for \partir:SPMD are defined in \Cref{fig:partir:spmd}.

\begin{figure}\footnotesize
\[
\begin{array}{ll}
    \begin{array}{lcll}
        \multicolumn{3}{l}{\textbf{Value definitions}} \\
        v &  ::= & op(\xs)         & \text{Tensor operations} \\
          & \mid & \slice{d}{x}{r_{a}}        & \text{Slicing} \\
          & \mid & \execute{\as}{\xs}{\rs}{\ys}{e} & \text{SPMD execution} \\
          & \mid & \redist{x}{\mu} & \text{Redistribution} \\
          & \mid & \tst{\sigmas}{x} & \text{Tiling/red. actions}
    \end{array} &
    \begin{array}{lcll}
        \multicolumn{3}{l}{\textbf{Distributed tensor types}} \\
        \rho & ::= & \ddim{\as}{n} ~\mid~ n & \text{Distr'd dimensions} \\
        \mu & ::= & \disttensor{\as}{\rhos} \;|\; \tau & \text{Distr'd types} \\
        \multicolumn{3}{l}{\textbf{Expressions}} \\
        e & ::=  & \multicolumn{2}{l}{\plet x_1{:}\mu_1,\ldots,x_n{:}\mu_n = v \pin e} \\
          & \mid & \multicolumn{2}{l}{\yield{\xs}}
    \end{array}
\end{array}\]
\\
\begin{mathpar}
  \Infer{TExec}
        { \xs = x_1\cdots x_m \\
          \Gamma \tspmd x_i : \mu_i \\
          \tau_i = \localtype{\mu_i} \\
          \ys = y_1\cdots y_m \\
          \as = a_1\ldots a_k \\
          \rs = r_{a_1}\ldots r_{a_k} \\
          \rs,\ol{y{:}\tau} \tcore e : \tensor{\ol{n}} }
        { \Gamma \tspmd \execute{\as}{\xs}{\rs}{\ys}{e} : \disttensor{\as}{\ol{n}}} \\ 
  \Infer{TRedist}
        { \Gamma \tspmd x : \mu_1 \qquad \mu_1 \sim \mu_2}
        { \Gamma \tspmd \redist{x}{\mu_2} : \mu_2} \qquad 
  \Infer{TTileRed}
        { \Gamma \tspmd x : \mu_1 \qquad \tilered{\sigmas}{\mu_1} = \mu_2}
        { \Gamma \tspmd \tst{\sigmas}{x} : \mu_2}
\end{mathpar}
\\[4pt]
\textbf{Extraction of local ($\mathcal{L}$) and global ($\mathcal{G}$) tensor types from distributed types}
\[\begin{array}{cccc}
  \begin{array}{l}
    \localdim{n} = n \\
    \globaldim{n} = n
  \end{array} &
  \begin{array}{l}
    \localdim{\ddim{\as,a}{n}} = \localdim{\ddim{\as}{(n / m)}} \,, \,\, a{:}m\in M \\
    \globaldim{\ddim{\as}{n}} = n
  \end{array} &
  \begin{array}{l}
    \localtype{\tau} = \tau \\
    \globaltype{\tau} = \tau
  \end{array} & 
  \begin{array}{l}
    \localtype{\disttensor{\as}{\rhos}} = \tensor{\overline{\localdim{\rho}}} \\
    \globaltype{\disttensor{\cdot}{\rhos}} = \tensor{\overline{\globaldim{\rho}}}
  \end{array} 
\end{array}\]
\\[4pt]
\textbf{Data equivalence of distributed tensor types}
\begin{mathpar}
  \Infer{EqTM} 
    {\globaltype{\mu} = \tau}
    {\tau \sim \mu} \qquad 
  \Infer{EqMT} 
    {\globaltype{\mu} = \tau}
    {\mu \sim \tau} \qquad 
  \Infer{EqMM}
    { \globaldim{\rho_i} = \globaldim{\rho_i'}}
    {\disttensor{\as}{\rhos} \sim \disttensor{\as}{\rhos'}}
\end{mathpar}
\\[4pt]
\textbf{\coretile and \coresum actions on distributed tensor types}
\[\begin{array}{lcl}
  \tilered{\cdot}{\mu} & = & \mu \\
  \tilered{\tileaction{a}{d},\sigmas}{\disttensor{\as a}{[\ldots,\ddim{\ol{c}_d}{n_d},\ldots]}} & = & \disttensor{\as}{[\ldots,\ddim{\ol{c}_d,a}{(n_d\cdot n)},\ldots]}\,, \quad a{:}n\in M \\
  \tilered{\sumaction{a},\sigmas}{\disttensor{\as a}{\rhos}} & = & \disttensor{\as}{\rhos}
\end{array}\] 
\caption{\partir:SPMD abstract syntax and typing rules.}
\label{fig:partir:spmd}
\end{figure}

\subsubsection{Distributed types and redistribution}
Syntactically, distributed types $\mu$ (\Cref{fig:partir:spmd}) subsume tensor types $\tau$ (\Cref{fig:partir:core}) or have the form \disttensor{\as}{\rhos}, where
axes $\as$ are referred to as {\em stacked} axes, and $\rhos$ are a list of {\em distributed dimensions}.
We now explain the semantics of distributed types by considering ways to distribute a matrix $x : \tensor{256,8}$ across the mesh $M = {\tt \{a:4, b:2\}}$.

\begin{figure*}
    \centering
    \includegraphics[scale=0.42]{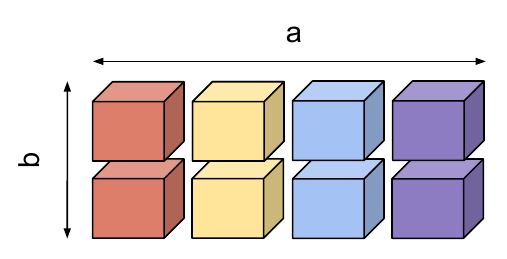}
    \hspace{10pt}
    \includegraphics[scale=0.42]{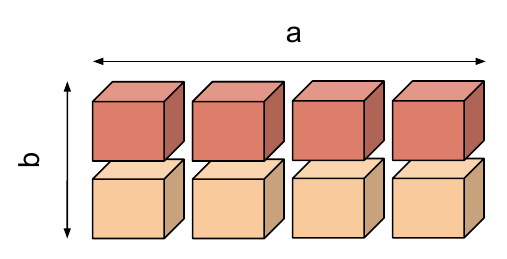}
    \hspace{10pt}
    \includegraphics[scale=0.42]{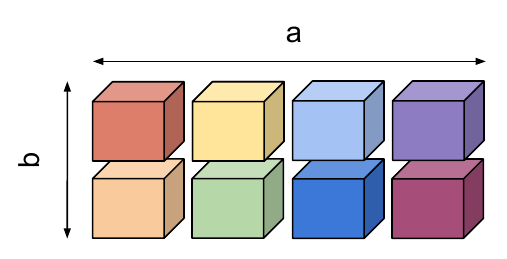}
    \caption{%
        Distributed types over mesh $M = \{a\colon 4, b\colon 2\}$.
        Left:   $\disttensor{\{\}}{[\{a\}256, 8]}$ (device-local type is
        $\tensor{64,8}$).
        Middle: $\disttensor{\{\}}{[256, \{b\}8]}$ (device-local type is
        $\tensor{256,4}$).
        Right:  $\disttensor{\{\}}{[\{a\}256, \{b\}8]}$ (device-local
        type is $\tensor{64,4}$).
        Different boxes correspond to different devices in the mesh;
        different colours indicate different data. 
    }
    \label{fig:spmd-distributed-types}
\end{figure*}

\newcommand\muacomma{\xspace\disttensor{\emptystacked}{[\{a\}256, 8]}\xspace}
\newcommand\mucommab{\xspace\disttensor{\emptystacked}{[256, \{b\}8]}\xspace}
\newcommand\muacommab{\xspace\disttensor{\emptystacked}{[\{a\}256, \{b\}8]}\xspace}

\paragraph{Distributed types without stacked axes}
\Cref{fig:spmd-distributed-types} shows different ways of distributing $x$.
For example, the type $\muacomma$ (left) specifies that each device along axis $a$ holds a different shard
of $256 / 4 = 64$ rows of $x$, and every device along \inlc{"b"} holds the same data.
Hence, the device-local type is $\localtype{\muacomma} = \tensor{64, 8}$.

\newcommand\mucommaba{\disttensor{\emptystacked}{[256, \{b,a\}8]}}
\newcommand\mucommaab{\disttensor{\emptystacked}{[256, \{a,b\}8]}}

There are more ways to distribute tensor $x$ beyond the ones in Figure~\ref{fig:spmd-distributed-types}.
Full replication of $x$'s data is expressed by the type $\disttensor{\emptystacked}{[256, 8]}$.
Furthermore, the same dimension of $x$ may be distributed along multiple axes.
For example, $\mucommaba$ and $\mucommaab$ both specify distributions where each device holds $8 / 4 / 2 = 1$ column of $x$ -- but the assignment to mesh
devices is transposed. Finally, we remark that distributed tensor types cannot
mention the {\em same} axis more than once in the distributed dimensions
$\rhos$.

\begin{figure}
    \centering
    \includegraphics[scale=0.54]{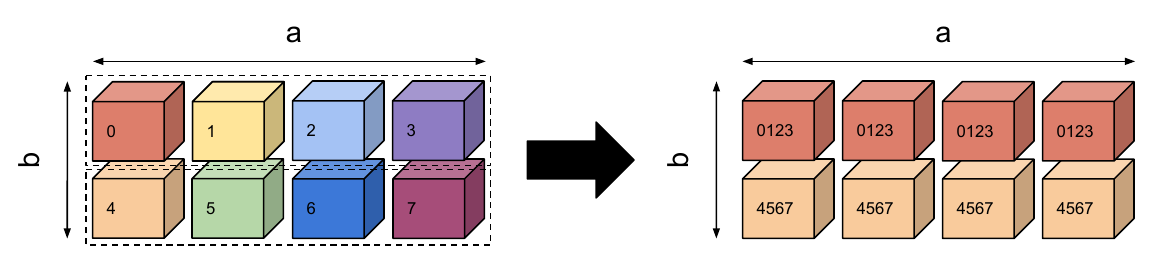}
    \caption{%
        Redistribution from
        $\disttensor{\emptystacked}{[256, \{a,b\}8]}$
        to
        $\disttensor{\emptystacked}{[256, \{b\}8]}$.
        Boxes correspond to devices;
        colours indicate data.
        On the left, each box holds one column of the global tensor, and column indices are inscribed inside the boxes.
        On the right, each box holds a set of four adjacent columns, as indicated by the inscribed numbers.
        This redistribution is a form of collective {\tt all\_gather}.
    }
    \label{fig:spmd-redistribution-allgather}    
\end{figure}

\paragraph{Redistribution}
All of the distributed types discussed above specify distributions of the
same {\em global} \tensor{256,8}.
Whenever two distributed types $\mu_1$ and $\mu_2$ have the same global type,
i.e.~$\globaltype{\mu_1} = \globaltype{\mu_2}$ in \Cref{fig:partir:spmd}, 
we can always convert a value of $\mu_1$ to a value of $\mu_2$ via {\em redistribution}.
To express such redistributions, \partir:SPMD features a \spmdredist operation, see rule \textsc{TRedist} in \Cref{fig:partir:spmd}.
Operationally, \spmdredist may introduce communication, as \Cref{fig:spmd-redistribution-allgather} shows.
For a detailed discussion of redistribution see~\cite{redist2021}.

\paragraph{Stacked axes}
Consider, for example, that each device in the mesh $M$ has performed a local computation yielding a local tensor of type \tensor{64, 4}.
A priori, there is no relationship between the data held in the local tensors on different devices -- it is 
just an unstructured collection of $64x4$-sized chunks of data.
We express this collection using the type $\disttensor{\{a,b\}}{[64, 4]}$,
where we record the axes as stacked.
Note that while $\localtype{\disttensor{\{a,b\}}{[64, 4]}} = \tensor{64, 4}$, the global type of a distributed type with stacked axes is undefined (see Figure~\ref{fig:partir:spmd}).
In order to define a global view, we apply loop actions to the collection of local tensors using the \spmdtst instruction. 

Acting on the aforementioned stacked type with $[\tileaction{a}{0}, \tileaction{b}{1}]$ results in local tensors being {\em viewed} as tiles of a global matrix of shape $256\times 8$ and type $\disttensor{\emptystacked}{[\{a\}256, \{b\}8]}$. 
Alternatively, acting with $[\tileaction{a}{1}, \tileaction{b}{0}]$ leads to \disttensor{\emptystacked}{[\{b\}128, \{a\}16]}, where the local tensors are viewed as tiles in a global matrix of shape $128\times 16$.
Hence, \coretile actions transform stacked axes in the input type into distributed axes in the output type.

When applying \coresum actions, on the other hand, the axes arguments disappear from the result type, denoting replication in the result.
For example, $[\sumaction{a}, \sumaction{b}]$ produces a global tensor of type \disttensor{\emptystacked}{[64, 4]}, which is the result of summing up all device-local tensors.

Note that it is generally possible to mix \coretile and \coresum actions in the same \spmdtst instruction;
and the actions in a \spmdtst instruction are not required to eliminate {\em all} stacked axes from the tensor argument's type.

\paragraph{Redistribution with stacked axes}
By virtue of rules \textsc{TRedist} and \textsc{EqMM} in \Cref{fig:partir:spmd}, redistribution is only allowed between distributed types with identical stacked axes.
In other words, the \spmdredist \emph{preserves} stacked axes. 
Hence, communication only takes place within groups of devices that have the same
mesh coordinates along the stacked axes.

\subsubsection{The \spmdexecute instruction}
\label{sec:spmd:spmd-instruction}

The \spmdexecute instruction expresses device-local computation and therefore returns distributed tensors with stacked axes.
It is analogous to the \partir:Core \coreloop instruction but more restrictive,
making \spmdexecute a useful target for lowering \partir:Core's \coreloop.
Specifically:
\begin{itemize}
    \item
    The computation in the body of an \spmdexecute instruction may not capture variables from outside the \spmdexecute instruction.
    All free variables of the body (i.e. the $\ys$ in rule \textsc{TExec}) must be explicit arguments to the \spmdexecute instruction (i.e. as the $\xs$ in rule \textsc{TExec}).
    \item
    The arguments of an \spmdexecute instruction may have arbitrary distributed types, but the body parameters have their corresponding local types.
    \item
    The result of an \spmdexecute instruction is a distributed tensor in which axes appear only as stacked axes (cf. the type annotation \disttensor{\as}{\ol{n}} in the conclusion of rule \textsc{TExec}).
    \item
    The stacked axes in the result type of a \spmdexecute instruction agree with the axes that the \spmdexecute instruction operates on (i.e. the $\as$ in rule \textsc{TExec}).
    \item
    Nesting of \spmdexecute instructions is disallowed (as enforced in rule \textsc{TExec} by requiring that $e$ must be a \partir:Core expression).
\end{itemize}
The one way in which \spmdexecute is less restrictive than \partir:Core's \coreloop instruction is in being able to span multiple mesh axes $\as$, which is necessary to represent nested \coreloop instructions without nesting \spmdexecute.
This is convenient since we are interested in targeting flat SPMD parallelism and do not have to concern ourselves with nested parallelism.

As an illustration of \spmdexecute, consider the following example that adds the device-local tiles of two $256\!\times\!8$-matrices given in the distributed tensors \inlinecode{\%x1}, \inlinecode{\%x2}:
\begin{lstlisting}[language=mlir]
    %r = spmd.execute "a" "b"
            (%x1: dtensor<{}, [{"a"}256, {"b"}8]>, %x2: dtensor<{}, [{"a"}256, {"b"}8]>)
            (%ra: range<4>, %rb: range<2>, %y1: tensor<64x4xf32>, %y2: tensor<64x4xf32>) {
      yield (add(%y1, %y2)) : tensor<64x4xf32>
    } : dtensor<{"a", "b"}, [64, 4]>
\end{lstlisting}
While the \spmdexecute instruction has arguments \inlinecode{\%x1}, \inlinecode{\%x2}, its body operates only on the local portions of these arguments, i.e. on \inlinecode{\%y1}, \inlinecode{\%y2} of type $\localtype{\disttensor{\emptystacked}{[\{a\}256, \{b\}8]}} = \tensor{64,4}$ each.
The result is a collection of local tensors of shape \inlinecode{64x4} that are stacked across both axes $a$ and $b$, as signified by the stacked axes in the result type of the \spmdexecute instruction.
It is worth pointing out that the exact same \spmdexecute instruction would also be valid if applied to operands with stacked axes, i.e. of type $\disttensor{\{a, b\}}{[64, 4]}$.

\subsubsection{Formal semantics}
\Cref{fig:interpretation:partir:spmd} formally defines the semantics of \partir:SPMD programs in a denotational style.

\begin{figure}\footnotesize
\begin{align*}
&
\begin{array}{l}
\multicolumn{1}{l}{\textbf{Interpretation of distributed types}} \\
\llbracket \disttensor{\cdot }{\overline{n}} \rrbracket := \llbracket \tensor{\overline{n}} \rrbracket
\\[2pt]
\llbracket \disttensor{a_1\cdots a_k}{\overline{n}} \rrbracket :=
    \llbracket a_1 \rrbracket \to \cdots \to \llbracket a_k \rrbracket \to \llbracket \tensor{\overline{n}} \rrbracket
\\[2pt]
\llbracket \disttensor{a_1\cdots a_k}{[n_1,\ldots, n_{d-1}, \{a\}n_d, n_{d+1}, \ldots, n_r]} \rrbracket :=
    \llbracket a_1 \rrbracket \to \cdots \to \llbracket a_k \rrbracket \to \llbracket a \rrbracket \to^{d}
        \llbracket \tensor{\overline{n}} \rrbracket
\end{array}
\\[4pt]
&
\begin{array}{llcll}
\multicolumn{5}{l}{\textbf{Interpretation of \partir:SPMD contexts}\hspace{30mm}} \\
\llbracket \cdot \rrbracket := \cdot &
    \text{empty context}
& \hspace{10mm} &
\llbracket \Gamma, x{:}\mu \rrbracket :=
    \llbracket \Gamma \rrbracket \times \llbracket \mu \rrbracket &
    \Gamma \text{ extended with tensor variable}
\end{array}
\\[4pt]
&
\begin{array}{l}
\textbf{Environment typing (same as in \Cref{fig:interpretation:partir:core})} \\
\gamma \in \llbracket \Gamma \rrbracket :\Leftrightarrow 
        dom(\gamma) = dom(\Gamma) \,\wedge
        \forall_{x\in dom(\gamma)}. \,\gamma(x) \in \llbracket\Gamma(x)\rrbracket
\end{array}
\\[4pt]
&
\begin{array}{l}
\textbf{Interpretation of values and expressions}
\\[2pt]
\llbracket \Gamma \tspmd e : \mu \rrbracket : \llbracket \Gamma \rrbracket \to \llbracket\mu\rrbracket
\\[2pt]
\llbracket \Gamma \tspmd \execute{\as}{\xs}{\rs}{\ys}{e} : \disttensor{\as}{\ol{n}} \rrbracket
        \, \gamma :=
        \lambda\overline{i}. \,
            \llbracket \rs,\ol{y{:}\tau} \tcore e : \tensor{\ol{n}} \rrbracket
            \, \left\{\overline{r\mapsto i}\right\} \cup \left\{\overline{y\mapsto\gamma(x)\overline{i}}\right\}
            \,, \\
\qquad\text{where $\gamma(x)\overline{i}$ means applying $\gamma(x)$ to as many indices of $\overline{i}$
as there are stacked axes in the distributed type of $x$}
\\[3pt]
\llbracket \Gamma \tspmd \tst{\sumaction{a}}{x} : \disttensor{\as}{\overline{n}} \rrbracket
        \, \gamma :=
        \lambda\overline{i}. \!
            \sum_{j\in\llbracket a \rrbracket}\gamma(x) \,\overline{i} \,j
        \,, \\[2pt]
\quad\text{where $\Gamma \tspmd x : \disttensor{\as a}{\overline{n}}$}
\\[4pt]
\llbracket \Gamma \tspmd \tst{\tileaction{a}{d}}{x} : \disttensor{\as}{\overline{\rho}} \rrbracket
        \, \gamma :=
        \lambda\overline{i}. \,
            \lambda^d j. \, \gamma(x) \,\overline{i} \,j
        \,, \\[2pt]
\qquad\text{%
    where $\Gamma \tspmd x : \disttensor{\as a}{\overline{n}}$
    and $\overline{\rho} = [n_1,\ldots, n_{d-1}, \{a\}n_d, n_{d+1}, \ldots, n_r]$
}
\\[4pt]
\llbracket \Gamma \tspmd \redist{x}{\disttensor{\as}{\overline{n}}} : \disttensor{\as}{\overline{n}} \rrbracket
        \, \gamma :=
        \lambda\overline{i}. \,
            \bigoplus_{j\in\llbracket a \rrbracket}^{axis=d} \gamma(x) \,\overline{i} \,j
        \,, \\[2pt]
\qquad\text{%
    where $\Gamma \tspmd x : \disttensor{\as}{\overline{\rho}}$
    and $\overline{\rho} = [n_1,\ldots, n_{d-1}, \{a\}n_d, n_{d+1}, \ldots, n_r]$
}
\\[4pt]
\llbracket \Gamma \tspmd op(\overline{x}) : \disttensor{\cdot}{\overline{n}} \rrbracket
        \, \gamma :=
        op\left(\overline{\gamma(x)}\right)
\\[3pt]
\llbracket \Gamma \tspmd \yield{x_1, \ldots, x_k} : \mu_1,\ldots,\mu_k \rrbracket
        \, \gamma :=
        \gamma(x_1), \ldots, \gamma(x_k)
\\[3pt]
\llbracket \Gamma \tspmd \plet \overline{x{:}\mu} = v \pin e : \mu \rrbracket
        \, \gamma :=
        \llbracket \Gamma, \overline{x{:}\mu} \tspmd e : \mu \rrbracket
        \, \gamma \cup \left\{\overline{x\mapsto \llbracket \Gamma \tspmd v : \mu \rrbracket\,\gamma}\right\}
\end{array}
\end{align*}
\caption{%
    Interpretation of \partir:SPMD,
    with at most a single axis appearing in distributed dimensions.
}
\label{fig:interpretation:partir:spmd}
\end{figure}

The top pane of \Cref{fig:interpretation:partir:spmd} specifies that distributed tensor types with {\em stacked} axes are interpreted as (curried) functions that map into the space $\llbracket \tensor{\overline{n}} \rrbracket$.

The complete semantics of \partir:SPMD needs to interpret general distributed tensor types, in which axes may appear in distributed dimensions (and not just as stacked axes).
However, for our correctness proof of the translation from \partir:Core to \partir:SMPD
it suffices to consider only distributed tensor types in which at most one distributed dimension contains a non-empty list of axes, and the non-empty list consists of a single axis only.
In the top pane of \Cref{fig:interpretation:partir:spmd} we therefore restrict ourselves to interpreting only distributed types where at most one axis appears in the distributed dimensions, and all other axes are stacked.
Types with an axis in a distributed dimension are in fact interpreted as the same function spaces as distributed types that contain only stacked axes;
we only annotate the final arrow with a superscript $d$ to indicate the dimension in which the single non-stacked axis occurs.

The interpretations of \partir:SPMD typing contexts $\Gamma$, and the typing of environments $\gamma$ are unsurprising.
We therefore move on to discussing the bottom pane of \Cref{fig:interpretation:partir:spmd}, i.e. the interpretations of values and expressions.

As in \Cref{fig:interpretation:partir:core}, the interpretations of \coreyield and \plet are standard.
Unlike in \partir:Core, however, \coreyield and \plet expressions in \partir:SPMD may involve tuples.
Specifically, \coreyield may return a tuple of values, and \plet may bind a tuple of values.
Hence, the subexpression $e$ in a \plet expression is interpreted in an extension of the environment $\gamma$ that includes a mapping for each of the variable names in $\xs$, as indicated by 
$
    \left\{\overline{x\mapsto \llbracket \Gamma \tspmd v : \mu \rrbracket\,\gamma}\right\}
$
at the bottom of \Cref{fig:interpretation:partir:spmd}.

We discuss the remaining interpretations in the bottom pane of \Cref{fig:interpretation:partir:spmd},
i.e. the interpretations of \partir:SPMD values, from top down.

The \spmdexecute instruction introduces stacked axes $\as$ into its result type.
Based on our interpretation of distributed types, an \spmdexecute instruction must therefore be interpreted as a function.
The body of this function is the interpretation of the subexpression $e$;
but note that $e$ is interpreted as a \partir:Core expression, in a suitable typing context $\overline{r},\overline{y{:}\tau}$.
This interpretation of $e$ is guaranteed to be meaningful by the premises of the typing rule \textsc{TExec} from \Cref{fig:partir:spmd}.

The interpretation of an \spmdtst instruction with a \sumaction{a} action is straightforward.
We sum over an index $j$ that takes values in $\llbracket a \rrbracket$,
where $a$ is the last of the stacked axes.

Applying a \tileaction{a}{d} action is even simpler,
thanks to our choice of assigning the same function space interpretation to both a type with only stacked axes and a type that has a single axis occurring in a distributed dimension:
we only use slightly different notation for these function spaces, by annotating the final arrow with a superscript $d$ if an axis occurs in a distributed dimension.
Accordingly, we write the result of interpreting a \tileaction{a}{d} action as a function with a final $\lambda^{d}$,
where the superscript $d$ is merely a notational reminder that the corresponding arrow also carries $d$ as an annotation.

The interpretation of \spmdredist in \Cref{fig:interpretation:partir:spmd} is only given for the situation where \spmdredist acts as an \texttt{all\_gather} operation along axis $a$.
This is in fact the only kind of \spmdredist instruction that will be needed for our correctness proof of the translations from \partir:Core to \partir:SPMD.
When \spmdredist acts as an \texttt{all\_gather} operation, it performs a concatenation of tiles;
hence the use of $\bigoplus^{axis=d}$ in the interpretation of \spmdredist in \Cref{fig:interpretation:partir:spmd}.
Note that because of
\begin{align*}
    &\Gamma \tspmd x : \disttensor{\as}{\overline{\rho}} \:\:\text{and} \\
    &\overline{\rho} = [n_1,\ldots, n_{d-1}, \{a\}n_d, n_{d+1}, \ldots, n_r] \,,    
\end{align*}
the index $j$ is necessarily passed as an argument to a $\lambda^{d}$.

Lastly, we discuss the interpretation in \partir:SPMD of operations $op$ that are inherited from StableHLO.
In a \partir:SPMD program, an operation $op$ can either appear at top level or nested under an \spmdexecute instruction.
When $op$ is nested under \spmdexecute, its interpretation in \partir:SPMD is, by definition, identical to its interpretation in \partir:Core
(see the equation for \spmdexecute in \Cref{fig:interpretation:partir:spmd}).
When an $op$ appears at top level, it neither consumes nor produces values that have stacked axes in their distributed types.
This is why, in \Cref{fig:interpretation:partir:spmd}, we restrict interpretations of typing judgements for $op$ to cases where the resulting type is of the form $\disttensor{\cdot}{\overline{n}}$,
i.e. has no stacked axes.

\paragraph{A note on \coreslice}
The \coreslice instruction from \partir:Core is also valid in \partir:SPMD
(cf. the definitions of values in \partir:SPMD in \Cref{fig:partir:spmd}).
Note that \coreslice instructions require an argument $r_a$ that is a range variable.
This is why, in a well-typed \partir:SPMD program, a \coreslice instruction can only appear in the body of an \spmdexecute.%
\footnote{%
    Note that \partir:SPMD typing contexts $\Gamma$ do not include range variables.
}
Since the body of an \spmdexecute instruction is interpreted with the \partir:Core semantics, \Cref{fig:interpretation:partir:spmd} does not include an explicit interpretation for \coreslice instructions.

\begin{figure}\footnotesize
\textbf{(Functional) Translation relation $\boxed{\mathcal{M} ; \sigmas \vdash \langle \pctx{C}; e \rangle \rightsquigarrow e'}$}
\\
\begin{mathpar}
    \infer[\textsc{SOp}]
        { \mathcal{M}; \sigmas \vdash \langle \pctx{C} ; \plet x{:}\tau = op(\ys) \pin e \rangle \rightsquigarrow e'}
        { \mathcal{M}; \sigmas \vdash \langle \pctx{C}[\plet x{:}\tau = op(\ys) \pin -]; e \rangle
                \rightsquigarrow e'} \qquad
    \infer[\textsc{SSlice}]
        { \mathcal{M}; \sigmas \vdash \langle \pctx{C}; \plet x{:}\tau = \slice{d}{y}{r_{a}} \pin e \rangle \rightsquigarrow e'}
        { \mathcal{M} ; \sigmas \vdash \langle \pctx{C}[\plet x{:}\tau = \slice{d}{y}{r_{a}} \pin -]; e \rangle
                \rightsquigarrow e'} \\
    \infer[\textsc{SYldTop}]
        { \mathcal{M} ; \cdot \vdash \langle \pctx{C}; \yield{x} \rangle
                \rightsquigarrow \pctx{C}[\yield{y}][\ol{\mathcal{M}(z)}/\zs] } 
        { \begin{array}{c}
            \mathcal{M}(x) = y \quad 
            \zs = \FV{\pctx{C}}
          \end{array} } \\
    \infer[\textsc{SYldL}]
        { \mathcal{M} ; \sigmas\sigma \vdash \langle \pctx{C} ; \yield{x} \rangle \rightsquigarrow 
          \plet x' = \execute{\as a}{\,\ol{\mathcal{M}(z)}}{\rs}{\ys}{\pctx{C}[\yield{x}][\ys / \zs]} \pin \yield{x'} }
        { \begin{array}{c}
            x{:}\tau\in\defs{\pctx{C}} \quad
            \as a = \axes{\sigmas\sigma} \quad
            \zs = \FV{\pctx{C}}
          \end{array} } \\
    \infer[\textsc{SYldP}]
        { \begin{array}{l} 
            \mathcal{M} ; \sigmas\sigma \vdash \langle \pctx{C}; \yield{x}\rangle \rightsquigarrow
            \plet x' = \execute{\as a}{z}{\rs}{y}{\yield{y}} \pin \yield{x'} 
          \end{array}}
        { \begin{array}{c}
            x{:}\_\notin\defs{\pctx{C}} \quad \as a = \axes{\sigmas\sigma} \quad
            \mapping{x{:}\tensor{\ol{n}}}{z{:}\disttensor{\ol{c}}{\ol{n}}}{} \in \mathcal{M} \quad 
            \ol{c} \subseteq \as
          \end{array} } \\
    \infer[\textsc{SYldC}]
        { \mathcal{M}; \sigmas\sigma \vdash \langle \pctx{C}; \yield{x}\rangle \rightsquigarrow \yield{z} }
        { \begin{array}{c}
            x{:}\_\notin\defs{\pctx{C}} \quad \as a = \axes{\sigmas\sigma} \quad
            \mapping{x{:}\tensor{\ol{n}}}{z{:}\disttensor{\as a}{\ol{n}}}{} \in \mathcal{M}
          \end{array} } \\
    \infer[\textsc{SLoop}]
        { \mathcal{M}; \sigmas \vdash \langle \pctx{C}; \plet x{:}\tau = \ploop{a}{\sigma}{e_1} \pin e_2\rangle 
                \rightsquigarrow \psctx{C}[\psctx{C}_1[\psctx{C}_*[e']]] } 
        { \begin{array}{c}
            \ol{x_\live{:}\tau_\live} = \{x_\live{:}\tau_\live \in \defs{\pctx{C}} \mid x_\live\in \FV{e_1,e_2}\} \quad 
            \as = \axes{\sigmas} \quad
            \zs = \FV{\pctx{C}} \\[2pt]
            \psctx{C} \defeq \left\{\begin{array}{l}
                -
                    \,, \text{ if $\ol{x_\live{:}\tau_\live}$ is empty} \\[1pt]
                \plet \ol{x'_\live} = \execute{\as}{\,\ol{\mathcal{M}(z)}}{\rs}{\ys}{\pctx{C}[\yield{\xs_\live}][\ys / \zs]} \pin -
                    \,, \text{ otherwise}
                \end{array}\right. \\[8pt]
            \tau_{\live i} = \tensor{\ol{n}_i} \qquad
            \mu_i = \disttensor{\as}{\ol{n}_i} \qquad
            \mathcal{M}_1 = \mathcal{M},\mapping{x_{\live i}{:}\tau_{\live i}}{x_{\live i}'{:}\mu_i}{} \\[2pt]
            \mathcal{M}_1 ; \sigmas\sigma \vdash \langle - ; e_1 \rangle \rightsquigarrow \psctx{C}_1[\yield{z}] \\[1pt]
            \tau = \tensor{\ol{n}} \qquad
            \mu = \disttensor{\ol{a}}{\ol{n}} \qquad
            \sigma \sfixup z \Downarrow \mu \rightsquigarrow \psctx{C}_*[\yield{z'}] \\[1pt]
            \mathcal{M}_1,\mapping{x{:}\tau}{z'{:}\mu}{\;} ; \sigmas \vdash \langle -; e_2 \rangle \rightsquigarrow e'
          \end{array} } \\
\text{{\bf \coretile/\coresum action with type coercion} $\boxed{\sigma \sfixup z \Downarrow \mu \rightsquigarrow e'}$}
\\
\begin{array}{l|l}
\begin{array}{l}
  \sumaction{a} \sfixup z \Downarrow \mu \rightsquigarrow \\
    \;\; \plet z'{:} \mu =  \tst{\sumaction{a}}{z} \\
    \;\; \pin \yield{z'} \\
    \phantom{\pin}
\end{array} & 
\begin{array}{l}
  \tileaction{a}{d} \sfixup z \Downarrow \mu \rightsquigarrow \\
  \;\; \begin{array}{l}
      \;\plet z'  = \tst{\tileaction{a}{d}}{z} \pin \\
      \;\plet z'' = \redist{z'}{\mu} \\
      \;\pin \yield{z''}
    \end{array}
\end{array}
\end{array}
\end{mathpar}
\\[4pt]
\textbf{Other auxiliary definitions}
\\[-8pt]
\[\begin{array}{lcll}
\pctx{C} & ::= & \plet x{:}\tau = op(\xs) \pin \pctx{C} \;\mid\; \plet x{:}\tau = \slice{d}{y}{r_{a}} \pin \pctx{C} \;\mid\; - & \text{Simple contexts} \\ 
\psctx{C} & ::= & \plet \ol{x{:}\mu} = v \pin \psctx{C} \;\mid\; - & \text{SPMD contexts} \\
\mathcal{M} & ::= & \cdot \;\mid\; \mathcal{M},\mapping{x{:}\tau}{y{:}\mu}{} & \text{variable and type maps} \\ \;\\
\end{array}\]
\\[-12pt]
\[\begin{array}{c|c|c}
\begin{array}{l}
\defs{-} = \emptyset \\
\defs{\plet x{:}\tau = \ldots \pin \pctx{C}} = \\
 \;\;\{x{:}\tau\} \cup \defs{\pctx{C}} 
\end{array} &
\begin{array}{l}
\mathcal{M}(x) = \left\{\begin{array}{ll}
  x \text{ if } x{:}\_ \notin dom(\mathcal{M}) \\
  y \text{ if } \mapping{x{:}\tau}{y{:}\mu}{} \in \mathcal{M}
  \end{array}\right.
\end{array} & 
\begin{array}{l}
  \axes{\cdot} = \cdot \\
  \axes{\sigmas,\tileaction{a}{d}} = \axes{\sigmas}a \\
  \axes{\sigmas,\sumaction{a}} = \axes{\sigmas}a
\end{array}
\end{array}\]
\caption{Translation from \partir:Core to \partir:SPMD.}
\label{fig:Core-to-SPMD-translation}
\end{figure}

\subsection{Lowering \partir:Core to \partir:SPMD}
\label{sec:spmd:lowering}

Lowering \partir:Core to \partir:SPMD is essentially a matter of translating \coreloop instructions to \spmdexecute instructions, with the main complication being the need to flatten nested \coreloop structures.
Our translation $\mathcal{M} ; \sigmas \vdash \langle \pctx{C}; e \rangle \rightsquigarrow e'$, defined in \Cref{fig:Core-to-SPMD-translation}, deals with this complication by keeping track of (a) the nesting level $\sigmas$ of \coreloop instructions and (b) local definitions $\pctx{C}$ in the \partir:Core source program.

\subsubsection{Overview of the translation relation}

The relation $\mathcal{M} ; \sigmas \vdash \langle \pctx{C}; e \rangle \rightsquigarrow e'$ specifies when a \partir:Core expression $\pctx{C}[e]$ lowers to the \partir:SPMD expression $e'$, in the presence of $\mathcal{M}$ and $\sigmas$.
\pctx{C} is a {\em simple context}, defined towards the bottom of \Cref{fig:Core-to-SPMD-translation}, that records the definitions of local variables that are in scope for $e$.
Here, {\em local} means that the path from the definition of a variable in \pctx{C} to its use in $e$ does not cross any \coreloop instructions.
The sequence $\sigmas$ indicates the nesting level at which $\pctx{C}[e]$ appears in the full source program.
Note that the $\sigmas$ keep track not only of the axes spanned by a \coreloop nest enclosing $\pctx{C}[e]$, but also of the corresponding loop actions.
Lastly, the map $\mathcal{M}$, defined towards the bottom of \Cref{fig:Core-to-SPMD-translation}, records the names and types $x{:}\tau$ of variables that are in scope for $\pctx{C}[e]$ in the full source program.
The names in the domain of $\mathcal{M}$ are mapped to the variable names and types $y{:}\mu$ they have been translated to while lowering those parts of the full source program that lexically precede $\pctx{C}[e]$.

The translation relation deals with nesting levels as follows.
While the translation keeps seeing local definitions by $op$ or \coreslice instructions, at fixed nesting level $\sigmas$, it pushes these instructions into the local context $\pctx{C}$ (rules \textsc{Sop} and \textsc{SSlice}).
When a \coreyield instruction is encountered, the current nesting level is exited.
Generally (rule \textsc{SYldL}) this means that the current local definitions in $\pctx{C}$ must be emitted into an \spmdexecute instruction that spans axes $\as a$, corresponding to the current nesting level $\sigmas\sigma$.
No \spmdexecute instruction is required when the \coreyield instruction appears at top level, i.e. at the end of the program (rule \textsc{SYldTop}), or when the yielded variable was defined outside $\pctx{C}$ but at the current nesting level $\sigmas\sigma$ (rule \textsc{STldC}).

When the translation encounters an instruction of the form \ploop{a}{\sigma}{e_1}, it must pass from nesting level $\sigmas$ to the deeper nesting $\sigmas\sigma$ (rule \textsc{SLoop}).
However, before the translation can process $e_1$ at nesting level $\sigmas\sigma$, it must emit the current context $\pctx{C}$.
This is because the variables that are defined in $\pctx{C}$ are defined at nesting level $\sigmas$ and therefore must be emitted into an \spmdexecute instruction that spans axes $\as=\axes{\sigmas}$.
Note that rule \textsc{SLoop} emits definitions only for those variables $\ol{x_\live{:}\tau_\live}$ (subscript $\live$ for {\em live}) that are defined in $\pctx{C}$ and also used in either $e_1$ or in the remainder of the program, i.e. in $e_2$.
These variables are translated into the $\ol{x_\live'}$ defined by the \partir:SPMD program fragment $\psctx{C}$.
The map $\mathcal{M}$ is extended with mappings from $\ol{x_\live}$ to $\ol{x_\live'}$ to give a new map $\mathcal{M}_1$, which is then used in the translation of $e_1$, at nesting level $\sigmas\sigma$.

The result of translating $e_1$ is matched against $\psctx{C}_1[\yield{z}]$ because the variable $z$ is needed in the remaining hypotheses of rule \textsc{SLoop}.
Specifically, $z$ is an input to the helper function $\sigma \sfixup z \Downarrow \mu \rightsquigarrow \ldots$ that (i) implements the loop action $\sigma$ with an \spmdtst instruction and (ii) coerces the result of the \spmdtst instruction to type $\mu$, which amounts to inserting a \spmdredist instruction when $\sigma$ is a \coretile action.
The \partir:SPMD code fragment returned from this helper function is matched against $\psctx{C}_{*}[\yield{z'}]$, again because the variable $z'$ plays a role in the remaining hypothesis of \textsc{SLoop}:
$\mathcal{M}_1$ is extended with a mapping from $x$ to $z'$, and the remainder $e_2$ of the input \plet expression is translated under this extended map and, importantly, at the original nesting level $\sigmas$.
This gives a \partir:SPMD expression $e'$, and rule \textsc{SLoop} concludes, finally, by stacking up the collected SPMD contexts $\psctx{C}$, $\psctx{C}_1$, $\psctx{C}_*$ and substituting $e'$ for the whole $-$ in the resulting context.

We conclude our overview of \Cref{fig:Core-to-SPMD-translation} by noting that the rules defining $\mathcal{M} ; \sigmas \vdash \langle \pctx{C}; e \rangle \rightsquigarrow e'$ are syntax-directed.
It is in fact straightforward to check that they define a function that takes a tuple $(\mathcal{M}, \sigmas, \pctx{C}, e)$ to a unique $e'$.
Our implementation is a direct transcription of the rules in \Cref{fig:Core-to-SPMD-translation} into a recursive function that executes in a single pass over the input \partir:Core program.

\subsubsection{Example translation}

We illustrate the translation function defined in \Cref{fig:Core-to-SPMD-translation} by discussing the lowering to \partir:SPMD of the code in \Cref{lst:Core-tiled-chain-matmul}.
The result of this lowering is shown in \Cref{lst:SPMD-tiled-chain-matmul};
and we now justify this, making reference to the definition of  $\mathcal{M} ; \sigmas \vdash \langle \pctx{C}; e \rangle \rightsquigarrow e'$.

\begin{lstlisting}[
    language=mlir,
    caption={Chained matrix multiplication using loops with {\coretile} and {\coresum} actions.},
    label={lst:Core-tiled-chain-matmul}
]
func @main(%x: tensor<256x8xf32>, %w1: tensor<8x16xf32>, %w2: tensor<16x8xf32>) 
  -> tensor<256x8xf32> attributes {mesh = {"a":4, "b":2}} {
    %r = loop "a" [#tile<"a", 0>] (%ra: range<4>) {
      %xs  = slice 0 %x[%ra]  : tensor<64x8xf32>
      %x1s = matmul(%xs, %w1) : tensor<64x16xf32>
      %x2s = loop "b" [#sum<"b">] (%rb: range<2>) {
        %x1ss = slice 1 %x1s[%rb]   : tensor<64x8xf32>
        %w2s  = slice 0 %w2[%rb]    : tensor<8x8xf32>
        %x2ss = matmul(%x1ss, %w2s) : tensor<64x8xf32>
        yield %x2ss : tensor<64x8xf32>
      }
      yield %x2s : tensor<64x8xf32>
    }
    return %r : tensor<256x8xf32>
}
\end{lstlisting}

\begin{lstlisting}[
    language=mlir,
    caption={\partir:SPMD code for the \inlc{@main} function from \Cref{lst:Core-tiled-chain-matmul}.},
    label={lst:SPMD-tiled-chain-matmul}
]
func @main(%x: dtensor<{}, [256,8]>, %w1: dtensor<{}, [8,16]>, %w2: dtensor<{}, [16,8]>)
  -> dtensor<{}, [256,8]> attributes {mesh = {"a":4, "b":2}} {
    %x1s0 = spmd.execute "a" 
                (%x: dtensor<{}, [256,8]>, %w1: dtensor<{}, [8,16]>)
                (%ra: range<4>, %yx: tensor<256x8xf32>, %yw1: tensor<8x16xf32>) {
        %xs  = slice 0 %yx[%ra]   : tensor<64x8xf32>
        %x1s = matmul(%xs,  %yw1) : tensor<64x16xf32>
        yield %x1s : tensor<64x16xf32>
    } : dtensor<"a", [64,16]>
    %x2s0 = spmd.execute "a" "b"
                (%x1s0: dtensor<"a", [64, 16]>, %w2: dtensor<{}, [16,8]>)
                (%ra: range<4>, %rb: range<2>, %yx1s0: tensor<64x16xf32>,
                 %yw2: tensor<16x8xf32>) {
        %x1ss = slice 1 %yx1s0[%rb] : tensor<64x8xf32>
        %w2s  = slice 0 %yw2[%rb]   : tensor<8x8xf32>
        %x2ss = matmul(%x1ss, %w2s) : tensor<64x8xf32>
        yield %x2ss : tensor<64x8xf32>
    } : dtensor<{"a","b"}, [64,8]>
    %x2s1 = spmd.tile_reduce [#sum<"b">] %x2s0     : dtensor<{"a"}, [64,8]>
    %x2s2 = spmd.tile_reduce [#tile<"a", 0>] %x2s1 : dtensor<{}, [{"a"}256,8]>
    %x2s3 = spmd.redistribute %x2s2 -> dtensor<{}, [256,8]>
    yield %x2s3 : dtensor<{}, [256,8]>
}
\end{lstlisting}

The translation starts with an empty map $\mathcal{M} = \cdot$, at empty nesting level $\sigmas = \cdot$ and with an empty local context $\pctx{C} = -$.
The first instruction that is encountered is the \ploop{a}{[\tileaction{a}{0}]}{e_{1}} that defines \inlc{\%r}.
Rule \textsc{SLoop} is triggered, but in the empty context $\pctx{C} = -$ an empty $\psctx{C} = -$ is generated and the body $e_{1}$ is translated under $\mathcal{M}_1 = \mathcal{M} = \cdot$.

Translation of $e_{1}$ uses rules \textsc{SSlice} and \textsc{SOp} to build up a nontrivial local context $\pctx{C}$ before the nested \ploop{b}{[\sumaction{b}]}{e_{11}} instruction that defines \inlc{\%x2s} is encountered.
This triggers \textsc{SLoop} again, this time with a nontrivial local context that defines variable \inlc{\%x1s}, which is used in the remainder of the program.
The SPMD context $\psctx{C}$ generated by this instance of \textsc{SLoop} produces the \spmdexecute instruction that defines \inlc{\%x1s0} in \Cref{lst:SPMD-tiled-chain-matmul}.

Translation then proceeds with the loop body $e_{11}$ and under the map $\mapping{\it x1s{:}\_}{{\it x1s0}{:}\_}{}$.
When the instruction \inlc{yield \%x2ss} inside $e_{11}$ is reached, rule \textsc{SYldL} triggers, producing the \spmdexecute instruction that defines \inlc{\%x2s0} in \Cref{lst:SPMD-tiled-chain-matmul}.
In this instance of \textsc{SYldL}, the free variables $\zs$ are \inlc{\%x1s} and \inlc{\%w2}.
Looking these up in the current map $\mapping{\it x1s{:}\_}{{\it x1s0}{:}\_}{}$ yields \inlc{\%x1s0} and \inlc{\%w2}, respectively, and these are indeed the arguments of the second \spmdexecute instruction in \Cref{lst:SPMD-tiled-chain-matmul}.

To finish the translation of the nested \ploop{b}{[\sumaction{b}]}{e_{11}} from \Cref{lst:Core-tiled-chain-matmul}, rule \textsc{SLoop} requires that the \sumaction{b} is applied to \inlc{\%x2s0}.
The result of this is assigned to \inlc{\%x2s1} in \Cref{lst:SPMD-tiled-chain-matmul}, implying that the remainder of the outer loop body $e_1$ is translated under the map $\mapping{\it x2s{:}\_}{\it x2s1{:}\_}{}$.

The remainder of $e_1$ consists of only the \inlc{yield \%x2s} instruction.
Since the local context $\pctx{C}$ is empty and the nesting level is $\sigma = \tileaction{a}{0}$, rule \textsc{SYldC} is triggered.
In the presence of the map $\mapping{\it x2s{:}\_}{\it x2s1{:}\_}{}$, this results in the translated instruction \inlc{yield \%x2s1}.
This finishes the loop body $e_1$;
and by virtue of the first invocation of \textsc{SLoop}, the \tileaction{a}{0} must be applied to \inlc{\%x2s1}, followed by the \spmdredist instruction that defines \inlc{\%x2s3} in \Cref{lst:SPMD-tiled-chain-matmul}.

The remainder of the full program is then translated under the map $\mapping{\it r{:}\_}{\it x2s3{:}\_}{}$.
Since the remainder consists of only the \inlc{return \%r} instruction, i.e. a top-level \coreyield, rule \textsc{SYldTop} applies.
The local context $\pctx{C}$ is empty, and looking up \inlc{\%r} in the current mapping gives \inlc{\%x2s3}.
Hence, the final instruction in the translated program in \Cref{lst:SPMD-tiled-chain-matmul} is \inlc{yield \%x2s3}.

The single rule from \Cref{fig:Core-to-SPMD-translation} that has not featured in our discussion so far is \textsc{SYldP}.
This rule is needed to extend the stacked axes $\cs$ in the type of the translation $z$ of a non-locally defined variable $x$ to the current nesting level $\sigmas\sigma$, where $\axes{\sigmas\sigma} = \as a$.
Note that due to $x{:}\_\notin\defs{\pctx{C}}$, the entire local context $\pctx{C}$ is dead code.

\subsection{Correctness of the translation from \partir:Core to \partir:SPMD}

To state the correctness theorem for the translation from \Cref{fig:Core-to-SPMD-translation}, we need a few auxiliary definitions.
First, we introduce the judgement $\Gamma\vdash\mathcal{M}$ to express that a mapping $\mathcal{M}$ is compatible with a (\partir:Core) typing context $\Gamma$.

\begin{definition}[Judgement $\Gamma\vdash\mathcal{M}$, Associated SPMD Typing Context]
Let $\Gamma$ be a typing context containing range variables and tensor variables with \partir:Core tensor types
(i.e. no distributed tensor types).
Let $r_{a_1}, \ldots, r_{a_k}$ be the range variables that appear in $\Gamma$
(in that order, from left to right).
We write $\Gamma\vdash\mathcal{M}$ precisely if $\mathcal{M}$ is a mapping such that
\begin{align*}
    \mapping{x{:}\tensor{\ol{m}}}{y{:}\disttensor{\bs}{\ol{m}}}{} \in \mathcal{M}
        \quad \Leftrightarrow \quad
    x{:}\tensor{\ol{m}}\in\Gamma
    \,\land\, 
    \ol{r_b} \text{ precede } x \text{ in }\Gamma
    \,.
\end{align*}

If $\Gamma \vdash \mathcal{M}$, we define the \emph{associated SPMD typing context} $\Gamma_{\mathcal{M}}$ to consist of all $y{:}\mu$ in the image of $\mathcal{M}$, ordered the same way as the pre-images of the $y{:}\mu$ are ordered in $\Gamma$.
\end{definition}

With this definition, we can already state (and prove) and interesting result.

\begin{theorem}[Typing simulation]
\label{thm:top-level-translation}
Let $\Gamma \tcore \pctx{C}[e] : \tensor{\ol{n}}$,
let $\Gamma\vdash\mathcal{M}$,
and let $\sigmas$ be such that $\axes{\sigmas} = a_1\cdots a_k$, where $r_{a_1},\ldots,r_{a_k}$ are the range variables in $\Gamma$.
There exists a (necessarily unique) $e'$ such that
\begin{itemize}
    \item $\mathcal{M}, \sigmas \vdash \langle \pctx{C}; e\rangle \rightsquigarrow e'$ and
    \item $\Gamma_{\mathcal{M}} \tspmd e' : \disttensor{a_1\cdots a_k}{\ol{n}}$.
\end{itemize}
\begin{proof}
Structural induction on $e$.
\end{proof}
\end{theorem}

Note that when studying correctness of program transformations, one is often interested in type {\em preservation} results.
The theorem essentially captures the interplay of range variables in \partir:Core and stacked axes in distributed tensor types in \partir:SPMD.
Because of this interplay, where stacked axes may appear in the distributed type of the translated expression $e'$, one cannot expect types to be preserved when lowering from \partir:Core to \partir:SPMD.
The typing {\em simulation} captured by \Cref{thm:top-level-translation} is as close as one can get to preserving types.

The key ingredient in our correctness proof for the translation from \Cref{fig:Core-to-SPMD-translation} is the following relation.
It relates an environment $\gamma$ for interpreting \partir:Core expressions to an environment $\gamma'$ for interpreting \partir:SPMD expressions.

\begin{definition}[Environment relation]
Let $\Gamma \vdash \mathcal{M}$.
For environments $\gamma\in\llbracket \Gamma\rrbracket$
and $\gamma'\in\llbracket \Gamma_{\mathcal{M}}\rrbracket$
define a relation $\sim_{\mathcal{M}}$ as follows:
\begin{align}
    \gamma \sim_{\mathcal{M}} \gamma' :\Leftrightarrow
        \forall_{\mapping{x{:}\tensor{\ol{m}}}{y{:}\disttensor{a_1\cdots a_k}{\ol{m}}}{} \in \mathcal{M}}.\,
            \gamma(x) = \gamma'(y) \, \gamma(r_{a_1}) \cdots \gamma(r_{a_k})
    \,.
\end{align}
\end{definition}

It is worth pointing out that this relation is very natural.
Since the domain of $\gamma'$ consists of tensor variables whose types generally include stacked axes, $\gamma'$ takes values in function spaces.
Therefore, only after evaluating $\gamma'(y)$ at a suitable number of arguments does one obtain a tensor that can be compared to $\gamma(x)$.
If given only $\gamma$ and $\gamma'$, then the only values that have the correct types to be passed as arguments to  $\gamma'(y)$ are the values that $\gamma$ takes on range variables.
(Recall that $\Gamma_{\mathcal{M}}$ does not contain any range variables, and hence $\gamma'$ cannot be evaluated on range variables.)

Using the relation $\sim_{\mathcal{M}}$, we now collect a number of lemmas that will help us streamline the proof of the correctness theorem for the translation form \partir:Core to \partir:SPMD.

\begin{lemma}
\label{thm:yield-op-in-context}
Let \pctx{C} be a simple context of the form
\begin{align*}
    \pctx{C} = \plet x_1{:}\tau_1 = op(\overline{y}_1) \pin \ldots \plet x_n{:}\tau_n = op(\overline{y}_n) \pin -
    \,.
\end{align*}
Let $\Gamma$ be a typing context such that
\begin{itemize}
    \item $\Gamma$ contains only \partir:Core tensor variables
    (i.e. no range variables and no distributed types),
    \item $\Gamma \tcore \pctx{C}[\yield{x}] : \tensor{\overline{n}}$.
    (Note that this necessitates $\FV{\pctx{C}[\yield{x}]}\subset\Gamma$.)
\end{itemize}
If $\Gamma\vdash\mathcal{M}$ and $\gamma\sim_{\mathcal{M}}\gamma'$, then
\begin{align*}
    \llbracket \Gamma_{\mathcal{M}} \tspmd \pctx{C}[\yield{y}][\overline{\mathcal{M}(z)}/\zs]
        : \disttensor{\cdot}{\overline{n}} \rrbracket
        \,\gamma'
    = \llbracket \Gamma \tcore \pctx{C}[\yield{x}] : \tensor{\overline{n}} \rrbracket
        \,\gamma
    \,,
\end{align*}
where $y = \mathcal{M}(x)$ and $\zs = \FV{\pctx{C}}$.
\end{lemma}
\begin{proof}
First note that the left-hand side of the claimed equation is well-defined by virtue of \Cref{thm:top-level-translation}.
Indeed, by rule \textsc{SYldTop} from \Cref{fig:Core-to-SPMD-translation}, we have
\begin{align*}
    \mathcal{M};\cdot \vdash \langle\pctx{C};\yield{x}\rangle \rightsquigarrow
        \pctx{C}[\yield{y}][\overline{\mathcal{M}(z)}/\zs]
    \,,
\end{align*}
and hence
\begin{align*}
    \Gamma_{\mathcal{M}} \tspmd \pctx{C}[\yield{y}][\overline{\mathcal{M}(z)}/\zs]
        : \disttensor{\cdot}{\overline{n}}
    \,.
\end{align*}

We now proceed by induction on the structure of $\pctx{C}$.
The base case for the induction is $\pctx{C} = -$, which implies $\zs = \FV{\pctx{C}} = \cdot\,$.
The left-hand side of the claimed equation then reduces to
\begin{align*}
    \llbracket \Gamma_{\mathcal{M}} \tspmd \yield{y} : \disttensor{\cdot}{\overline{n}} \rrbracket
        \, \gamma'
        &= \gamma'(y) \\
        &= \gamma(x) \\
        &= \llbracket \Gamma \tcore \yield{x} : \tensor{\overline{n}} \rrbracket \, \gamma
    \,,
\end{align*}
where we have made use of $\mapping{x{:}\tensor{\overline{n}}}{y{:}\disttensor{\cdot}{\overline{n}}}{} \in\mathcal{M}$ and $\gamma\sim_{\mathcal{M}}\gamma'$.

For the induction step, consider
\begin{align*}
    \pctx{C} = \plet x_1{:}\tensor{\overline{m}} = op(\overline{y}_1) \pin \pctx{C'}
    \,,
\end{align*}
and define
\begin{align*}
    &\zs_1 = \FV{\pctx{C'}} \\
    &\Gamma_1 = \Gamma, x_1{:}\tensor{\overline{m}} \\
    &\mathcal{M}_1 = \mathcal{M}, \mapping{x_1{:}\tensor{\overline{m}}}{x_1{:}\disttensor{\cdot}{\overline{m}}}{} \\
    &\gamma_1 = \gamma \cup \{x_1 \mapsto \llbracket \Gamma \tcore op(\overline{y}_1) : \tensor{\overline{m}} \rrbracket \gamma\} \\
    &\gamma'_1 = \gamma' \cup \{x_1 \mapsto
        \llbracket \Gamma_{\mathcal{M}} \tspmd op\!\left(\overline{\mathcal{M}(y_1)}\right)
            : \disttensor{\cdot}{\overline{m}} \rrbracket \gamma'\} \\
    &\hat{x} = \llbracket \Gamma_1 \tcore \pctx{C'}[\yield{x}] : \tensor{\overline{n}} \rrbracket \,\gamma_1 \\
    &\hat{x}' = \llbracket \Gamma_{1\mathcal{M}_1} \tspmd \pctx{C'}[\yield{y}][\overline{\mathcal{M}_1(z_1)}/\zs_1] 
                    : \disttensor{\cdot}{\overline{n}} \rrbracket \,\gamma'_1
    \,.
\end{align*}
It is straightforward to check that $\gamma_1\sim_{\mathcal{M}_1}\gamma'_1$, which follows from $\gamma\sim_{\mathcal{M}}\gamma'$ and the interpretations of $op$ in \partir:Core and \partir:SPMD.
Hence $\hat{x}' = \hat{x}$, by the induction hypothesis.
Now, starting from the right-hand side of the claimed equation,
\begin{align*}
    &\llbracket \Gamma \tcore \plet x_1{:}\tensor{\overline{m}} = op(\overline{y}_1) \pin \pctx{C'}[\yield{x}]
        : \tensor{\overline{n}} \rrbracket \,\gamma \\
    &= \llbracket \Gamma_1 \tcore \pctx{C'}[\yield{x}] : \tensor{\overline{n}} \rrbracket \,\gamma_1 \\
    &= \llbracket \Gamma_{1\mathcal{M}_1} \tspmd \pctx{C'}[\yield{y}][\overline{\mathcal{M}_1(z_1)}/\zs_1] 
                    : \disttensor{\cdot}{\overline{n}} \rrbracket \,\gamma'_1 \\
    &= \llbracket \Gamma_{\mathcal{M}} \tspmd \left(
                \plet x_1{:}\disttensor{\cdot}{\overline{m}} = op(\overline{y}_1)
                \pin \pctx{C'}[\yield{y}]
            \right)\![\overline{\mathcal{M}(z)}/\zs] 
                : \disttensor{\cdot}{\overline{n}} \rrbracket \,\gamma'
    \,,
\end{align*}
establishing the claim.
\end{proof}

The next lemma states that evaluating a \plet expression is equivalent to extending an environment $\gamma$ with suitable mappings for the \plet\!\!-bound variables.
This lemma is not particularly specific to \partir since versions of this lemma hold for interpretations of \plet expressions in any language.
We therefore state the next lemma without proof, which is a straightforward induction on $n$.

\begin{lemma}[and Definition of Extension of Environments]
\label{thm:environment-extension}
In both cases below, let $\gamma\in\llbracket\Gamma\rrbracket$.
\begin{enumerate}
    \item (\partir:Core).
    Let
    \begin{align*}
        \pctx{C} = \plet x_1{:}\tau_1 = v_1 \pin \ldots \plet x_n{:}\tau_n = v_n \pin -
        \,.
    \end{align*}
    If $\,\Gamma \tcore \pctx{C}[e] : \tau$, then
    \begin{align*}
        \llbracket \Gamma \tcore \pctx{C}[e] : \tau \rrbracket \, \gamma =
            \llbracket \Gamma \cup \defs{\pctx{C}} \tcore e : \tau \rrbracket \, \gamma_n
        \,,
    \end{align*}
    where
    \begin{align*}
        &\gamma_0 := \gamma \,, \\
        &\gamma_{k+1} := \gamma_k \cup
        \left\{x_{k+1} \mapsto \llbracket \Gamma, x_1{:}\tau_1,\ldots, x_k{:}\tau_k \tcore v_{k+1} : \tau_{k+1} \rrbracket \, \gamma_k \right\}
        \:\text{for}\: k=0,\ldots,n-1.
    \end{align*}
    We set $\gamma_{\pctx{C}} := \gamma_n$ and refer to $\gamma_{\pctx{C}}$ as the \emph{extension of $\gamma$ with the definitions from $\pctx{C}$}.
    ~\\[-6pt]
    \item (\partir:SPMD).
    Let
    \begin{align*}
        \psctx{C} = \plet x_1{:}\mu_1, \ldots, x_n{:}\mu_n = v \pin -
        \,.
    \end{align*}
    If $\,\Gamma \tspmd \psctx{C}[e] : \mu$, then
    \begin{align*}
        \llbracket \Gamma \tspmd \psctx{C}[e] : \mu \rrbracket \, \gamma =
            \llbracket \Gamma, x_1{:}\mu_1, \ldots, x_n{:}\mu_n \tspmd e : \mu \rrbracket \, \gamma_{\psctx{C}}
        \,,        
    \end{align*}
    where
    \begin{align*}
        \gamma_{\psctx{C}}
            & := \gamma \cup
             \left\{  x_1\mapsto \pi_1\!\left(\llbracket \Gamma \tspmd v : \mu_1,\ldots,\mu_n \rrbracket\,\gamma \right),
                      \ldots, \right. \\
            &\hspace{11.2mm}
             \left.   x_n\mapsto \pi_n\!\left(\llbracket \Gamma \tspmd v : \mu_1,\ldots,\mu_n \rrbracket\,\gamma \right)
            \right\} 
        \,,
    \end{align*}
    where $\pi_1,\ldots,\pi_n$ are the projections onto the components of an $n$-tuple.
    We refer to $\gamma_{\psctx{C}}$ as the \emph{extension of $\gamma$ with the definitions from $\psctx{C}$}.
\end{enumerate}
\end{lemma}

In our final lemma, we show that related environments $\gamma$ and $\gamma'$ remain related when extending with definitions from contexts $\pctx{C}$ and $\psctx{C}$, respectively.
The key to making this work is to choose a suitable context $\psctx{C}$ that simulates in \partir:SPMD the definitions from the \partir:Core context $\pctx{C}$.

\begin{lemma}
\label{thm:relation-for-extended-environments}
Let $\pctx{C}$ be a simple context with $\zs = \FV{\pctx{C}}$.
Let $\Gamma$ be a typing context that contains range variables $r_{a_1},\ldots,r_{a_l}$
and such that $\zs\subset\Gamma$,
and let $S = \{\overline{x{:}\tau}\} \subset\defs{\pctx{C}}$.
Define
\begin{align*}
    &\psctx{C} = \left\{\begin{array}{l}
            - \,, \text{ if } S = \emptyset \\[1pt]
            \plet \ol{x'} = \execute{\as}{\,\ol{\mathcal{M}(z)}}{\rs}{\ys}{\pctx{C}[\yield{\xs}][\ys / \zs]} \pin -
                \,, \text{ otherwise}
        \end{array}\right.
    \\
    &\Gamma_S = \Gamma \cup S \\
    &\mathcal{M}_S = \mathcal{M} \cup \{
            \mapping{x_i{:}\tensor{\overline{n}_i}}{x'_i{:}\disttensor{\as}{\overline{n}_i}}{} ~\mid~
            x_i{:}\tensor{\overline{n}_i}\in S
        \}
    \,.
\end{align*}
If $\,\Gamma\vdash\mathcal{M}$ and $\gamma\sim_{\mathcal{M}}\gamma'$, then also $\Gamma_S\vdash\mathcal{M}_S$ and
\begin{align*}
    \gamma_{\pctx{C}\vert S} \sim_{\mathcal{M}_S} \gamma'_{\psctx{C}}
    \,,
\end{align*}
where $\gamma_{\pctx{C}\vert S}$ is the restriction of $\gamma_{\pctx{C}}$ to $dom(\mathcal{M}_S)$.
\end{lemma}
\begin{proof}
We proceed by induction on $\pctx{C}$.
In the base case for the induction, $\pctx{C}=-$.
Hence $S=\emptyset$ and thus $\psctx{C}=-$.
It follows that $\gamma_{\pctx{C}\vert S} = \gamma$
and $\gamma'_{\psctx{C}} =\gamma'$.
Hence there is nothing left to show.

For the induction step, let 
\begin{align*}
    \pctx{C} = \pctx{C'}[\plet x_{k+1}{:}\tensor{\overline{n}_{k+1}} = v \pin -]
    \,.
\end{align*}
Consider two cases:
\begin{enumerate}
    \item $x_{k+1}{:}\tensor{\overline{n}_{k+1}} \notin S$.
    Then $\gamma_{\pctx{C}\vert S} = \gamma_{\pctx{C'}\vert S}$,
    and $\gamma_{\pctx{C'}\vert S} \sim_{\mathcal{M}_S}\gamma'_{\psctx{C}}$ holds by the induction hypothesis.
    
    \item $x_{k+1}{:}\tensor{\overline{n}_{k+1}} \in S$.
    For $x'_i$ with $i\le k$, we have
    \begin{align*}
        \gamma'_{\psctx{C}}(x'_i) \,\overline{\gamma_{\pctx{C}\vert S}(r)}
        = \gamma'_{\psctx{C'}}(x'_i) \,\overline{\gamma_{\pctx{C}\vert S}(r)}
        = \gamma'_{\psctx{C'}}(x'_i) \,\overline{\gamma_{\pctx{C'}\vert S}(r)}
        = \gamma_{\pctx{C'}\vert S}(x_i)
        = \gamma_{\pctx{C}\vert S}(x_i)
        \,,
    \end{align*}
    where the induction hypothesis was used in the second-to-last equation.
    
    For $x'_{k+1}$,
    \begin{align*}
        &\gamma'_{\psctx{C}}(x'_{k+1}) \,\overline{\gamma_{\pctx{C}\vert S}(r)} \\
        &= \llbracket \Gamma_{\mathcal{M}} \tspmd
            \execute{\as}{\overline{\mathcal{M}(z)}}{\rs}{\ys}{\pctx{C}[\yield{x_{k+1}}][\ys / \zs]}
                : \_
            \rrbracket \,\gamma'\,\overline{\gamma_{\pctx{C}\vert S}(r)} \\
        &= \llbracket \overline{r}, \overline{y{:}\_} \tcore \pctx{C}[\yield{x_{k+1}}][\ys / \zs]
                : \_
            \rrbracket 
            \left\{\overline{r \mapsto \gamma_{\pctx{C}\vert S}(r)}\right\} \cup
                \left\{
                    \overline{y \mapsto \gamma'(\mathcal{M}(z))
                        \,\overline{\gamma_{\pctx{C}\vert S}(r)}}
                \right\} \\
        &= \llbracket \overline{r}, \overline{y{:}\_} \tcore \pctx{C}[\yield{x_{k+1}}][\ys / \zs]
                : \_
            \rrbracket 
            \left\{\overline{r \mapsto \gamma(r)}\right\} \cup
                \left\{
                    \overline{y \mapsto \gamma'(\mathcal{M}(z))
                        \,\overline{\gamma(r)}}
                \right\} \\
        &= \llbracket \overline{r}, \overline{y{:}\_} \tcore \pctx{C}[\yield{x_{k+1}}][\ys / \zs]
                : \tensor{\overline{n}_{k+1}} \rrbracket 
            \left\{\overline{r \mapsto \gamma(r)}\right\} \cup
                \left\{\overline{y \mapsto \gamma(z)}\right\} \\
        &= \llbracket \Gamma \tcore \pctx{C}[\yield{x_{k+1}}]
                : \tensor{\overline{n}_{k+1}} \rrbracket 
            \,\gamma
            \\
        &= \llbracket \Gamma\cup\defs{\pctx{C}} \tcore \yield{x_{k+1}}
                : \tensor{\overline{n}_{k+1}} \rrbracket 
            \,\gamma_{\pctx{C}} \\
        &= \gamma_{\pctx{C}}(x_{k+1}) \\
        &= \gamma_{\pctx{C}\vert S}(x_{k+1})
        \,,
    \end{align*}
    where the third-to-last equality holds by the definition of $\gamma_{\pctx{C}}$,
    and the last equality holds because $x_{k+1}{:}\tensor{\overline{n}_{k+1}} \in S$.
    Note that we also used $\gamma_{\pctx{C}\vert S}(r_{a_j}) = \gamma(r_{a_j})$, for $j=0,\ldots, l$,
    which holds since no mappings for range variables are added in constructing $\gamma_{\pctx{C}}$ from $\gamma$.
    Finally, note that we used the assumption $\gamma\sim_{\mathcal{M}}\gamma'$ to rewrite with
    \begin{align*}
        \gamma'(\mathcal{M}(z))\,\overline{\gamma(r)} = \gamma(z)
    \end{align*}
    in going from the third to fourth equality above.
\end{enumerate}
~\\[-24pt]
\end{proof}

\subsubsection{The correctness theorem}
We finally have all the pieces in place to state and proof the following theorem
which expresses that the translation relation from \Cref{fig:Core-to-SPMD-translation} preserves semantics when lowering a \partir:Core program to \partir:SPMD.
In other words, the translation defined by \Cref{fig:Core-to-SPMD-translation} is correct,
relative to the \partir:Core semantics from \Cref{fig:interpretation:partir:core} and the \partir:SPMD semantics from \Cref{fig:interpretation:partir:spmd}.

\begin{theorem}[Correctness of translation]
\label{thm:correctness-of-translation}
Let $\Gamma \tcore \pctx{C}[e] : \tensor{\ol{n}}$ and let $\sigmas$ be such that $\axes{\sigmas} = a_1\cdots a_k$, where $r_{a_1},\ldots,r_{a_k}$ are the range variables in $\Gamma$.
Let $\mathcal{M}$, $e'$, $\gamma$, $\gamma'$ such that
\begin{itemize}
    \item $\Gamma\vdash\mathcal{M}$,
    \item $\mathcal{M}, \sigmas \vdash \langle \pctx{C}; e\rangle \rightsquigarrow e'$ and
    \item $\gamma\sim_{\mathcal{M}}\gamma'$.
\end{itemize}
Then,
\begin{align*}
    \llbracket \Gamma_{\mathcal{M}} \tspmd e' : \disttensor{a_1\cdots a_k}{\ol{n}} \rrbracket \, \gamma'\, \gamma(r_{a_1}) \cdots \gamma(r_{a_k}) =
        \llbracket \Gamma \tcore \pctx{C}[e] : \tensor{\ol{n}} \rrbracket
        \, \gamma
    \,.
\end{align*}
Note that the left-hand side is well-defined by virtue of \Cref{thm:top-level-translation}.
\end{theorem}

\begin{proof}
We proceed by induction on the height of the derivation tree of $\mathcal{M}, \sigmas \vdash \langle \pctx{C}; e\rangle \rightsquigarrow e'$.
The base case for the induction is when the derivation tree has height one, i.e. when a single rule from \Cref{fig:Core-to-SPMD-translation} establishes $\mathcal{M}, \sigmas \vdash \langle \pctx{C}; e\rangle \rightsquigarrow e'$.
This single rule must then be either \textsc{SYldTop}, \textsc{SYldL}, \textsc{SYldP} or \textsc{SYldC}.
\begin{enumerate}
    \item \textsc{SYldTop}.
    Since $\sigmas = \cdot$, the typing context $\Gamma$ contains no range variables.
    Hence, the simple context \pctx{C} must be of the form
    \begin{align*}
        \pctx{C} = \plet x_1{:}\tau_1 = op(\overline{y}_1) \pin \ldots \plet x_n{:}\tau_n = op(\overline{y}_n) \pin -
        \,.
    \end{align*}
    (No \coreslice operations can occur in \pctx{C} since no range variables are in scope.)
    The claim then follows from \Cref{thm:yield-op-in-context}.
    ~\\
    \item \textsc{SYldL}.
    \begin{align*}
        &\llbracket \Gamma_{\mathcal{M}} \tspmd \plet x' = \ldots \pin \yield{x'}
            : \disttensor{\as a}{\overline{n}} \rrbracket
            \,\gamma'\,\overline{\gamma(r)} \\
        & = \llbracket \Gamma_{\mathcal{M}} \tspmd \execute{\as a}{\overline{\mathcal{M}(z)}}{\rs}{\ys}{\ldots}
            : \disttensor{\as a}{\overline{n}} \rrbracket
            \,\gamma'\,\overline{\gamma(r)} \\
        & = \llbracket \rs, \overline{y{:}\tau} \tcore \pctx{C}[\yield{x}][\ys/\zs]
            : \tensor{\overline{n}} \rrbracket
            \,\left\{\overline{r\mapsto \gamma(r)}\right\} \cup
              \left\{
                \overline{y \mapsto \gamma'(\mathcal{M}(z))\,\overline{\gamma(r)}}
              \right\} \\
        & = \llbracket \rs, \overline{z{:}\tau} \tcore \pctx{C}[\yield{x}]
            : \tensor{\overline{n}} \rrbracket
            \,\left\{\overline{r\mapsto \gamma(r)}\right\} \cup
              \left\{
                \overline{z \mapsto \gamma'(\mathcal{M}(z))\,\overline{\gamma(r)}}
              \right\} \\
        & = \llbracket \rs, \overline{z{:}\tau} \tcore \pctx{C}[\yield{x}]
            : \tensor{\overline{n}} \rrbracket
            \,\left\{\overline{r\mapsto \gamma(r)}\right\} \cup
              \left\{
                \overline{z \mapsto \gamma(z)}
              \right\} \\
        & = \llbracket \Gamma \tcore \pctx{C}[\yield{x}]
            : \tensor{\overline{n}} \rrbracket \,
                \gamma
        \,,
    \end{align*}
    where, in going to the second-to-last line, we used $\gamma\sim_{\mathcal{M}}\gamma'$;
    and the last equality holds because we necessarily have $\{\rs, \overline{z{:}\tau}\}\subset\Gamma$ since $\zs=\FV{\pctx{C}}$.
    ~\\
    \item \textsc{SYldP}.
    \begin{align*}
        &\llbracket \Gamma_{\mathcal{M}} \tspmd \plet x' = \ldots \pin \yield{x'}
            : \disttensor{\as a}{\overline{n}} \rrbracket 
            \,\gamma'\,\overline{\gamma(r)} \\
        & = \llbracket \Gamma_{\mathcal{M}} \tspmd \execute{\as a}{z}{\rs}{y}{\yield{y}}
            : \disttensor{\as a}{\overline{n}} \rrbracket
            \,\gamma'\,\overline{\gamma(r)} \\
        & = \llbracket \rs, y{:}\tensor{\overline{n}} \tcore \yield{y}
            : \tensor{\overline{n}} \rrbracket
            \,\left\{\overline{r\mapsto \gamma(r)} \right\} \cup
              \left\{y \mapsto \gamma'(z)\,\overline{\gamma(r)} \right\} \\
        & = \llbracket \rs, x{:}\tensor{\overline{n}} \tcore \yield{x}
            : \tensor{\overline{n}} \rrbracket
            \,\left\{\overline{r\mapsto \gamma(r)} \right\} \cup
              \left\{x \mapsto \gamma'(z)\,\overline{\gamma(r)} \right\} \\
        & = \llbracket \rs, x{:}\tensor{\overline{n}} \tcore \yield{x}
            : \tensor{\overline{n}} \rrbracket
            \,\left\{\overline{r\mapsto \gamma(r)} \right\} \cup
              \{x \mapsto \gamma(x) \} \\
        & = \gamma(x) \\
        & = \llbracket \Gamma \tcore \pctx{C}[\yield{x}]
            : \tensor{\overline{n}} \rrbracket \,
                \gamma
        \,,
    \end{align*}
    where, in going to the third-to-last line,
    we used $\mapping{x{:}\tensor{\overline{n}}}{z{:}\disttensor{\cs}{\overline{n}}}{} \in \mathcal{M}$
    from the premises of \textsc{SYldP}, in combination with $\gamma\sim_{\mathcal{M}}\gamma'$;
    and the last equality holds because of the premise $x\notin\defs{\pctx{C}}$ of \textsc{SYldP}.
    ~\\
    \item \textsc{SYldC}.
    \begin{align*}
        &\llbracket \Gamma_{\mathcal{M}} \tspmd \yield{z}
            : \disttensor{\as a}{\overline{n}} \rrbracket
            \,\gamma'\,\overline{\gamma(r)} \\
        & = \gamma'(z)\,\overline{\gamma(r)} \\
        & = \gamma(x) \\
        & = \llbracket \Gamma \tcore \pctx{C}[\yield{x}] : \tensor{\overline{n}} \rrbracket
            \,\gamma
        \,,
    \end{align*}
    where, in going to the second-to-last line,
    we used $\mapping{x{:}\tensor{\overline{n}}}{z{:}\disttensor{\as a}{\overline{n}}}{} \in \mathcal{M}$
    from the premises of \textsc{SYldC}, in combination with $\gamma\sim_{\mathcal{M}}\gamma'$;
    and the last equality holds because of the premise $x\notin\defs{\pctx{C}}$ of \textsc{SYldC}.
\end{enumerate}

For the induction step, assume that the claim holds whenever the judgement $\mathcal{M}, \sigmas \vdash \langle \pctx{C}; e\rangle \rightsquigarrow e'$ has a derivation tree of height $h$.
We then need to establish the claim for derivation trees of height $h+1$.
In this case, the final rule from \Cref{fig:Core-to-SPMD-translation} that is used in the derivation of $\mathcal{M}, \sigmas \vdash \langle \pctx{C}; e\rangle \rightsquigarrow e'$ must be either \textsc{SOp}, \textsc{SSlice} or \textsc{SLoop}.  
\begin{enumerate}
    \item \textsc{SOp}.
    By the induction hypothesis, we can assume that the conclusion holds for context $\pctx{C}[\plet x{:}\tau = op(\ys) \pin -]$, \partir:Core expression $e$ and translation result $e'$.
    Hence, the conclusion also holds for context $\pctx{C}$, \partir:Core expression
    \begin{align*}
        \plet x{:}\tau = op(\ys) \pin e
    \end{align*}
    and translation result $e'$.
    But this is precisely the desired claim since
    \begin{align*}
        \pctx{C}[\plet x{:}\tau = op(\ys) \pin -][e] = \pctx{C}[\plet x{:}\tau = op(\ys) \pin e]
        \,.
    \end{align*}
    \\[-16pt]
    \item \textsc{SSlice}.
    The same reasoning as in the previous case (for rule \textsc{SOp}) applies.
    ~\\
    \item \textsc{SLoop}.
    We show the details of the proof for the case $\sigma =\sumaction{a}$.
    The case $\sigma =\tileaction{a}{d}$ is handled analogously.
    
    Let $\mathcal{M}_1$ be as in the premise of \textsc{SLoop} and define
    \begin{align*}
        &\Gamma_1 = dom(\mathcal{M}_1) \cup \{r_a\}
        \,.
    \end{align*}
    Then, $\Gamma_1\vdash \mathcal{M}_1$,
    and for $\gamma_1\sim_{\mathcal{M}_1}\gamma'_1$ we have 
    \begin{align}
        \llbracket \Gamma_{1\mathcal{M}_1} \tspmd \psctx{C}_1[\yield{z}]
        : \disttensor{\as a}{\overline{n}} \rrbracket
        \,\gamma'_1\,\overline{\gamma_1(r)}\,\gamma_1(r_a) \nonumber \\
        = \llbracket \Gamma_1 \tcore e_1 : \tensor{\overline{n}} \rrbracket \,\gamma_1
        \label{eq:correctness:sloop:ih1}
    \end{align}
    by the induction hypothesis.
    Analogously, define
    \begin{align*}
        &\Gamma_2 = dom(\mathcal{M}_1) \cup \{x{:}\tau\} \\
        &\mathcal{M}_2 = \mathcal{M}_1, \mapping{x{:}\tau}{z'{:}\mu}{}
        \,.
    \end{align*}
    Then, $\Gamma_2\vdash \mathcal{M}_2$,
    and for $\gamma_2\sim_{\mathcal{M}_2}\gamma'_2$ we have 
    \begin{align}
        \llbracket \Gamma_{2\mathcal{M}_2} \tspmd e'
        : \disttensor{\as}{\overline{m}} \rrbracket
        \,\gamma'_2\,\overline{\gamma_2(r)}
        = \llbracket \Gamma_2 \tcore e_2 : \tensor{\overline{m}} \rrbracket \,\gamma_2
        \,,
        \label{eq:correctness:sloop:ih2}
    \end{align}
    again by the induction hypothesis.
    
    Now, let $S = \{x_\live{:}\tau_\live \in \defs{\pctx{C}} \mid x_\live\in \FV{e_1,e_2}\}$
    (from the premise of \textsc{SLoop}),
    and note that $\Gamma\cup S = dom(\mathcal{M}_1)$.
    Starting from the right-hand side of the equality claimed in the statement of the theorem,
    \begin{align*}
        &\llbracket \Gamma \tcore \pctx{C}[\plet x{:}\tau = \ploop{a}{\sigma}{e_1} \pin e_2]
            : \tensor{\overline{m}} \rrbracket \,\gamma \\
        &= \llbracket \Gamma\cup\defs{\pctx{C}} \tcore \plet x{:}\tau = \ploop{a}{\sigma}{e_1} \pin e_2
            : \tensor{\overline{m}} \rrbracket \,\gamma_{\pctx{C}} \\
        &= \llbracket \Gamma\cup S \tcore \plet x{:}\tau = \ploop{a}{\sigma}{e_1} \pin e_2
            : \tensor{\overline{m}} \rrbracket \,\gamma_{\pctx{C}\vert S} \\
        &= \llbracket \Gamma_2 \tcore e_2
            : \tensor{\overline{m}} \rrbracket
            \,\gamma_{\pctx{C}\vert S} \cup
                \left\{ x\mapsto
                    \Sigma_j \llbracket \Gamma_1 \tcore e_1 : \tau \rrbracket
                        \,\gamma_{\pctx{C}\vert S}\cup\{r_a\mapsto j\}
                \right\}
        \,,
    \end{align*}
    By \Cref{thm:relation-for-extended-environments} we have $\gamma_{\pctx{C}\vert S} \sim_{\mathcal{M}_1} \gamma'_{\psctx{C}}$, and hence also
    \begin{align*}
        \left(\gamma_{\pctx{C}\vert S} \cup \{r_a\mapsto j\}\right) \sim_{\mathcal{M}_1} \gamma'_{\psctx{C}}
        \,.
    \end{align*}
    (Adding mappings for range variables to the end of an environment $\gamma$ does not affect the relation $\gamma\sim_{\mathcal{M}}\gamma'$.)
    Setting
    \begin{align*}
        \hat{x} = 
            \llbracket \Gamma_{1\mathcal{M}_1} \tspmd \psctx{C}_1[\yield{z}]
                : \disttensor{\as a}{\overline{n}} \rrbracket
                    \,\gamma'_{\psctx{C}}
        \,,
    \end{align*}
    we can therefore apply \Cref{eq:correctness:sloop:ih1} to arrive at
    \begin{align*}
        &\llbracket \Gamma \tcore \pctx{C}[\plet x{:}\tau = \ploop{a}{\sigma}{e_1} \pin e_2]
            : \tensor{\overline{m}} \rrbracket \,\gamma \\
        &= \llbracket \Gamma_2 \tcore e_2
            : \tensor{\overline{m}} \rrbracket
            \,\gamma_{\pctx{C}\vert S} \cup 
                \left\{ x\mapsto \Sigma_j\hat{x}\,\overline{\gamma(r)} j \right\}
    \end{align*}
    It is straightforward to check that
    \begin{align*}
        \left(
            \gamma_{\pctx{C}\vert S} \cup 
                \left\{ x\mapsto \Sigma_j\hat{x}\,\overline{\gamma(r)} j \right\}
        \right)
        \sim_{\mathcal{M}_2}
        \left(
            \gamma'_{\psctx{C}} \cup 
                \left\{ z' \mapsto \lambda\overline{i}.\,\Sigma_j \hat{x}\,\overline{i} j \right\}
        \right)
        \,,
    \end{align*}
    which allows us to apply \Cref{eq:correctness:sloop:ih2} to obtain
    \begin{align*}
        &\llbracket \Gamma \tcore \pctx{C}[\plet x{:}\tau = \ploop{a}{\sigma}{e_1} \pin e_2]
            : \tensor{\overline{m}} \rrbracket \,\gamma \\
        &= \llbracket \Gamma_{2\mathcal{M}_2} \tspmd e' : \disttensor{\as}{\overline{m}} \rrbracket
                \left(
                    \gamma'_{\psctx{C}} \cup 
                        \left\{ z' \mapsto \lambda\overline{i}.\,\Sigma_j \hat{x}\,\overline{i} j \right\}
                \right) \overline{\gamma(r)} \\
        &= \llbracket \Gamma_{1\mathcal{M}_1}, z'{:}\mu \tspmd e' : \disttensor{\as}{\overline{m}} \rrbracket
                \left(
                    \gamma'_{\psctx{C}} \cup 
                        \left\{ z' \mapsto \lambda\overline{i}.\,\Sigma_j \hat{x}\,\overline{i} j \right\}
                \right) \overline{\gamma(r)} \\
        &= \llbracket \Gamma_{1\mathcal{M}_1}, z{:}\_ \tspmd \psctx{C}_{*}[e'] : \disttensor{\as}{\overline{m}} \rrbracket
                \left(
                    \gamma'_{\psctx{C}} \cup 
                        \left\{ z \mapsto \lambda\overline{i}.\,\lambda j.\, \hat{x}\,\overline{i} j \right\}
                \right) \overline{\gamma(r)} \\
        &= \llbracket \Gamma_{1\mathcal{M}_1}, z{:}\_ \tspmd \psctx{C}_{*}[e'] : \disttensor{\as}{\overline{m}} \rrbracket
                \left(
                    \gamma'_{\psctx{C}} \cup 
                        \{ z \mapsto \hat{x} \}
                \right) \overline{\gamma(r)} \\
        &= \llbracket \Gamma_{1\mathcal{M}_1} \tspmd \psctx{C}_1[\psctx{C}_{*}[e']] : \disttensor{\as}{\overline{m}} \rrbracket
                \,\gamma'_{\psctx{C}} \,\overline{\gamma(r)} \\
        &= \llbracket \Gamma_{\mathcal{M}} \tspmd \psctx{C}[\psctx{C}_1[\psctx{C}_{*}[e']]] 
                : \disttensor{\as}{\overline{m}} \rrbracket
                \,\gamma'\,\overline{\gamma(r)}
        \,,
    \end{align*}
    where the last equality holds by the definition of $\gamma'_{\psctx{C}}$.
\end{enumerate}
~\\[-24pt]
\end{proof}

\subsection{Lowering redistribution and \spmdtst instructions}

Redistribution generally requires communication.
Hence, the \spmdredist instruction is ultimately expanded into the \partir:HLO collectives from \Cref{lst:mhlo-collectives}.
The axis attributes on these collectives specify communication patterns in the context of a fixed device mesh.
This aligns well with our distributed types by admitting typing rules such as the following for \spmdallgather:
\begin{mathpar}\footnotesize
  \Infer{TAllGather}
        { \Gamma \tspmd x : \disttensor{\cs}{[\{\as_1\bs_1\}n_1,\ldots,\{\as_k\bs_k\}n_k]} }
        { \Gamma \tspmd \allgather{[\as_1, \ldots, \as_k]}{x} : \disttensor{\cs}{[\{\bs_1\}n_1,\ldots,\{\bs_k\}n_k]} }
\end{mathpar}

While \cite{redist2021} gives a general method for efficiently implementing redistribution, in \partir we have found the following to work sufficiently well in practice.
Given types
{\small\begin{gather*}
    \mu_1 = \disttensor{\cs_{\textit stacked}}{[\{\as_1\cs_1\}n_1,\ldots,\{\as_k\cs_k\}n_k]}
    \,, \\
    \mu_2 = \disttensor{\cs_{\textit stacked}}{[\{\bs_1\cs_1\}n_1,\ldots,\{\bs_k\cs_k\}n_k]}
    \,,
\end{gather*}}%
we implement \redist{x}{\mu_2} for $x$ of type $\mu_1$ as
{\small\begin{gather*}
    \plet x''{:}\mu'' = \allgather{[\as_1,\ldots,\as_k]}{x} \pin
    \allslice{[\bs_1,\ldots,\bs_k]}{x'} \,,
\end{gather*}}%
where $\mu'' = \disttensor{\cs_{\textit stacked}}{[\{\cs_1\}n_1,\ldots,\{\cs_k\}n_k]}$,
and \spmdallslice has a typing rule dual to \textsc{TAllGather}.
Syntactically, one regards $\mu''$ as a common suffix type of $\mu_1$ and $\mu_2$.
Semantically, this means that the intermediate tensor $x''$ is only as replicated as it needs to be to enable an implementation of \redist{x}{\mu_2}
with a single pair of \spmdallgather and \spmdallslice.

A \tst{[\sumaction{a}]}{x} also requires communication, and it lowers to \spmdallsum:
\begin{mathpar}\footnotesize
  \Infer{TAllSum}
        { \Gamma \tspmd x : \disttensor{\cs a}{[\{\bs_1\}n_1,\ldots,\{\bs_k\}n_k]} }
        { \Gamma \tspmd \allsum{a}{x} : \disttensor{\cs}{[\{\bs_1\}n_1,\ldots,\{\bs_k\}n_k]} }
\end{mathpar}
A \tst{[\tileaction{a}{d}]}{x}, on the other hand, is a trivial operation:
it only changes the type of $x$ to specify how the local tensors stacked along axis $a$ are to be viewed as a global tensor.

\end{document}